\documentclass[11pt]{article}
\usepackage{fullpage}
\usepackage[hyphens]{url}
\usepackage{amsmath,amssymb,amsthm,mathtools}
\usepackage{enumitem}
\usepackage{microtype}
\usepackage{xcolor}
\usepackage{colortbl}
\usepackage{booktabs}
\usepackage{needspace}
\usepackage[algo2e,ruled]{algorithm2e}
\usepackage{letltxmacro}
\usepackage[round,authoryear]{natbib}
\usepackage[pdfusetitle,hypertexnames=false,pagebackref,colorlinks,
  linkcolor=blue!45!black,citecolor=blue!45!black,urlcolor=blue!55!black]{hyperref}
\usepackage[nameinlink,capitalize,noabbrev]{cleveref}
\usepackage{thm-restate}
\usepackage{autonum}

\renewcommand*{\backref}[1]{}
\renewcommand*{\backrefalt}[4]{%
  \ifcase#1\relax
  \or (cited on page~#2)%
  \else (cited on pages~#2)%
  \fi}

\newtheorem{theorem}{Theorem}
\newtheorem{lemma}[theorem]{Lemma}
\newtheorem{proposition}[theorem]{Proposition}
\newtheorem{corollary}[theorem]{Corollary}

\NewCommandCopy{\arxivproof}{\proof}
\NewCommandCopy{\endarxivproof}{\endproof}
\RenewDocumentEnvironment{proof}{o}{%
  \arxivproof
  \IfValueT{#1}{[#1]\enspace}%
  \ignorespaces
}{\endarxivproof}

\newcommand{\restate}[1]{%
  \begingroup
  \LetLtxMacro\savedrestatelabel\label
  #1*%
  \GlobalLetLtxMacro\label\savedrestatelabel
  \endgroup}

\DeclarePairedDelimiter{\prn}{(}{)}
\let\set\relax
\DeclarePairedDelimiter{\set}{\{}{\}}
\DeclarePairedDelimiterX{\Set}[2]{\{}{\}}{\,{#1}\,:\,{#2}\,}
\DeclarePairedDelimiter{\abs}{|}{|}
\DeclarePairedDelimiter{\norm}{\|}{\|}
\DeclarePairedDelimiter{\inpr}{\langle}{\rangle}
\DeclarePairedDelimiter{\brc}{[}{]}
\DeclarePairedDelimiter{\ceil}{\lceil}{\rceil}
\DeclarePairedDelimiter{\floor}{\lfloor}{\rfloor}
\newcommand{\R}{\mathbb{R}}

\newcommand{\E}{\mathbb{E}}
\newcommand{\B}{\mathbb{B}}
\newcommand{\eps}{\varepsilon}
\newcommand{\conv}{\operatorname{conv}}
\newcommand{\poly}{\operatorname{poly}}
\DeclareMathOperator*{\argmax}{arg\,max}
\DeclareMathOperator*{\argmin}{arg\,min}
\DeclareMathOperator{\diag}{diag}
\DeclareMathOperator{\tr}{tr}
\DeclareMathOperator{\dist}{dist}
\DeclareMathOperator{\sgn}{sgn}
\DeclareMathOperator{\KL}{KL}
\DeclareMathOperator{\TV}{TV}

\newcommand{\decompterm}[3]{%
  \underbrace{\raisebox{2\normalbaselineskip}[0pt][0pt]{\hypertarget{decomp.#1.#2}{}}#3}_{\mathrm{(#2)}}}
\newcommand{\decompref}[2]{\hyperlink{decomp.#1.#2}{\ifmmode\mathrm{(#2)}\else(#2)\fi}}
\newcommand{\width}{\omega_{\psi_1}}

\crefname{section}{Section}{Sections}
\Crefname{section}{Section}{Sections}
\crefname{subsection}{Section}{Sections}
\Crefname{subsection}{Section}{Sections}
\crefname{subsubsection}{Section}{Sections}
\Crefname{subsubsection}{Section}{Sections}
\crefname{appendix}{Appendix}{Appendices}
\Crefname{appendix}{Appendix}{Appendices}
\crefname{theorem}{Theorem}{Theorems}
\Crefname{theorem}{Theorem}{Theorems}
\crefname{lemma}{Lemma}{Lemmas}
\Crefname{lemma}{Lemma}{Lemmas}
\crefname{proposition}{Proposition}{Propositions}
\Crefname{proposition}{Proposition}{Propositions}
\crefname{corollary}{Corollary}{Corollaries}
\Crefname{corollary}{Corollary}{Corollaries}
\crefname{definition}{Definition}{Definitions}
\Crefname{definition}{Definition}{Definitions}
\crefname{example}{Example}{Examples}
\Crefname{example}{Example}{Examples}
\crefname{remark}{Remark}{Remarks}
\Crefname{remark}{Remark}{Remarks}
\crefname{conjecture}{Conjecture}{Conjectures}
\Crefname{conjecture}{Conjecture}{Conjectures}
\crefname{axiom}{Axiom}{Axioms}
\Crefname{axiom}{Axiom}{Axioms}
\crefname{table}{Table}{Tables}
\Crefname{table}{Table}{Tables}
\crefname{figure}{Figure}{Figures}
\Crefname{figure}{Figure}{Figures}
\crefname{algorithm}{Algorithm}{Algorithms}
\Crefname{algorithm}{Algorithm}{Algorithms}
\crefname{algocf}{Algorithm}{Algorithms}
\Crefname{algocf}{Algorithm}{Algorithms}

\crefname{moveproperty}{Lemma}{Lemmas}
\Crefname{moveproperty}{Lemma}{Lemmas}
\crefformat{moveproperty}{Lemma~#2\getrefnumber{lem:move-revoke}#1#3}
\Crefformat{moveproperty}{Lemma~#2\getrefnumber{lem:move-revoke}#1#3}
\crefrangeformat{moveproperty}{Lemma~#3\getrefnumber{lem:move-revoke}#1#4--#5#2#6}
\Crefrangeformat{moveproperty}{Lemma~#3\getrefnumber{lem:move-revoke}#1#4--#5#2#6}
\crefname{widthproperty}{Lemma}{Lemmas}
\Crefname{widthproperty}{Lemma}{Lemmas}
\crefformat{widthproperty}{Lemma~#2\getrefnumber{lem:width-properties}#1#3}
\Crefformat{widthproperty}{Lemma~#2\getrefnumber{lem:width-properties}#1#3}

\makeatletter
\renewcommand{\fps@figure}{tb}
\renewcommand{\fps@table}{tb}
\renewcommand{\fps@algocf}{tb}
\expandafter\LetLtxMacro\expandafter\originalalgorithm\csname algorithm2e\endcsname
\expandafter\renewcommand\csname algorithm2e\endcsname[1][tb]{\originalalgorithm[#1]}
\expandafter\LetLtxMacro\expandafter\originalalgorithmstar\csname algorithm2e*\endcsname
\expandafter\renewcommand\csname algorithm2e*\endcsname[1][tb]{\originalalgorithmstar[#1]}
\makeatother

\title{Minimax-Optimal Small-Loss Regret in Online Convex Optimization and~Applications to Contextual Search and Online Inverse Optimization}
\author{Shinsaku Sakaue%
  \thanks{CyberAgent, Tokyo, Japan. National Institute of Informatics, Tokyo, Japan.
    Center for Advanced Intelligence Project, RIKEN, Tokyo, Japan.
    \texttt{shinsaku.sakaue@gmail.com}.}}
\date{}

\makeatletter
\patchcmd{\@maketitle}{\@title}{%
  \makebox[\linewidth][c]{%
    \parbox{0.83\paperwidth}{\centering\@title\par}}%
}{}{\PackageError{arxiv-title}{Could not set the title width}{}}
\makeatother

\begin{document}
\SetAlgoNoEnd
\maketitle

\begin{abstract}
We give \emph{small-loss regret bounds} for online convex
optimization (OCO), using one subgradient per round.
Let~$\Delta_T^\star$ denote the cumulative loss of the best fixed decision
in hindsight, measured relative to the per-round minima.
Our algorithm achieves the regret bound~%
$O\prn{1+\sqrt{\Delta_T^\star}}$ without knowing~$\Delta_T^\star$, up to domain and gradient-scale factors.
The bound is minimax optimal, including its domain dependence,
on~$\ell_p$ balls for every~$1\le p\le\infty$.
No smoothness or condition bounding gradient norms by loss values is assumed,
and neither loss values nor the time horizon are needed.
Algorithmically, we extend the \emph{metric-update revocation} of Gupta et al.\ (2026), originally developed for \emph{online inverse optimization}, to general OCO using the follow-the-regularized-leader framework and a mixture of multiple base learners.
We also apply our algorithm to \emph{contextual search} and online inverse optimization.
In~$d$-dimensional contextual search with comparison flips of unknown bias~%
$\kappa>0$, we obtain minimax expected cumulative
absolute error~$\Theta\prn{\sqrt d/\kappa^2}$ under Euclidean normalization
using polynomially many arithmetic operations, resolving an open question of
Liu et al.\ (2021).
For online inverse optimization with the Euclidean unit-ball utility domain, our algorithm bounds the regret relative to observed actions by $O\prn{\sqrt d+d^{1/4}\sqrt{D_T}}$, where~$D_T$ is their
cumulative suboptimality; this matches the regret rate of Gupta et al.\ (2026) when~$D_T=0$. 
Our guarantee extends to other utility domains;
for example, on the probability simplex, the bound improves to
$O\prn{\log\prn{d+1}+\sqrt{D_T\log\prn{d+1}}}$.

\end{abstract}

\section{Introduction}\label{sec:intro}
Online convex optimization (OCO) is a framework for sequential decision-making
with convex losses. In each round~$t$, a learner chooses a point~%
$w_t$ in a fixed compact convex set~$\mathcal K\subset\R^d$ and incurs the loss~%
$f_t(w_t)$ for a continuous convex loss function~$f_t$. We consider subgradient
feedback: after choosing~$w_t$, the learner observes~$g_t\in\partial f_t(w_t)$, with a known bound~$G>0$ on~$\norm{g_t}_2$,
but need not observe the loss value or the function itself.
The goal is to compete with every fixed decision~$u\in\mathcal K$.
After~$T$ rounds, where~$T$ is the time horizon, the \emph{regret} against~$u$ is
\begin{equation}\label{eq:regret}
R_T(u)\coloneqq\sum_{t=1}^T\prn{f_t(w_t)-f_t(u)}.
\end{equation}
By convexity, $R_T(u)$ is at most the \emph{linearized regret}
$R_T^{\mathrm{lin}}(u)\coloneqq\sum_{t=1}^T\inpr{g_t,w_t-u}$.
In this paper, we assume that~$\mathcal K$ contains at least two points, as both~$R_T(u)$ and~$R_T^{\mathrm{lin}}(u)$ vanish on a singleton domain.

For general convex losses, the worst-case regret grows at the rate of~$\sqrt T$. For nonnegative losses, \emph{small-loss bounds} express this guarantee in terms of the minimum cumulative loss of a fixed decision, $L_T^\star\coloneqq \min_{w\in\mathcal K}\sum_{t=1}^T f_t(w)$, and can remain bounded when~$L_T^\star$ is bounded.
In OCO, small-loss bounds often rely on assumptions that relate gradient norms to loss values.
For example, \citet{srebro2010smoothness} obtain small-loss bounds growing as~$\sqrt{L_T^\star}$ for nonnegative losses on~$\R^d$ with smoothness constant~$\beta_{\mathrm{sm}}>0$, using the \emph{self-bounding} inequality~$\norm{\nabla f_t(w_t)}_2^2\le4\beta_{\mathrm{sm}} f_t(w_t)$.
For general nonnegative convex losses, \citet{leme2022density} show that the learner's cumulative loss can exceed~$\prn{1+O\prn{\eps}}L_T^\star$ by at most a term logarithmic in~$T$, for fixed sufficiently small~$\eps>0$.

We seek small-loss bounds without the horizon dependence, including for
general (nonsmooth) convex losses. Since adding a constant to a loss changes neither the regret nor its
subgradients, we measure the comparator's loss relative to the minimum
in each round, thereby allowing losses to be negative. Specifically, we define the \emph{shifted comparator loss} and its minimum by
\begin{equation}\label{eq:intro-gap}
\Delta_T(u)\coloneqq\sum_{t=1}^T\prn{f_t(u)-\min_{w\in \mathcal K}f_t(w)}
\qquad
\text{and}
\qquad
\Delta_T^\star\coloneqq\min_{u\in \mathcal K}\Delta_T(u).
\end{equation}
When every per-round minimum is zero, we have $\Delta_T(u)=\sum_{t=1}^T f_t(u)$ and $\Delta_T^\star=L_T^\star$.
We aim to achieve a regret bound in terms of~$\Delta_T(u)$, with no explicit dependence on~$T$.

\subsection{Contributions}\label{sec:contributions}
We present a deterministic OCO algorithm with a small-loss regret bound on any compact convex domain, using one observed subgradient per round.
This small-loss guarantee follows from a stronger bound on the linearized regret in terms of
$N_T(u)\coloneqq\sum_{t=1}^T\max\set{\inpr{g_t,u-w_t},0}$,
the sum of the negative parts of the linearized excess losses, which satisfies~$N_T(u)\le\Delta_T(u)$.
The bounds depend on the domain through the \emph{maximal subexponential mean width}~%
$\width(\mathcal K)$ (see \cref{sec:width}).
\begin{restatable}{theorem}{adaptivewidthbound}\label{thm:adaptive-known-width}
For the standard OCO setting (see \cref{sec:protocol}), given known bounds~$G,W>0$ with $G\ge\norm{g_t}_2$ for all $t\ge1$ and $W\ge\width(\mathcal K)$, \cref{alg:adaptive} in \cref{app:master} satisfies,
simultaneously for all~$u\in\mathcal K$ and~$T\ge1$,
\begin{equation}\label{eq:intro-adaptive}
R_T(u)\le R_T^{\mathrm{lin}}(u)\le c\prn*{GW+\sqrt{GW N_T(u)}}\le c\prn*{GW+\sqrt{GW\Delta_T(u)}},
\end{equation}
where~$c>0$ is a universal constant. The bound on~$R_T^{\mathrm{lin}}(u)$ in terms of~%
$N_T(u)$ also holds for arbitrary inputs with~$\norm{g_t}_2\le G$.
\end{restatable}
This also gives a regret bound for online linear optimization (OLO)
in terms of~$N_T(u)$, the sum of the negative parts of the per-round excess losses.
\cref{app:lower} shows that on~$\ell_p$ balls with~$1\le p\le\infty$, ellipsoids, boxes, and the probability simplex, choosing an upper bound~$W\asymp\width(\mathcal K)$ makes the regret bound for OCO minimax optimal for the expected regret against the best fixed comparator; the worst-case instance ranges over all horizons and loss sequences with~$\Delta_T^\star\le b$,
for any~$b\ge0$, and the algorithm needs no prior knowledge of~$b$ or~$\Delta_T(u)$.

\paragraph{Method.}
We extend the \emph{metric-update revocation} of \citet{gupta2026optimal}, originally developed for \emph{online inverse linear optimization} (OILO), to general
OCO using the follow-the-regularized-leader (FTRL) framework.
We add quadratic regularization terms in directions obtained from the feedback. Roughly speaking, we revoke a term's curvature when the iterate has moved far enough in that direction from the point where the term was added, thereby limiting the influence of past regularization on later updates.
For ease of presentation, we first focus on the Euclidean unit-ball domain and obtain a
horizon-independent bound on $2R_T^{\mathrm{lin}}(u)-\varepsilon R_T^{\mathrm{abs}}(u)$, where $R_T^{\mathrm{abs}}(u)\coloneqq\sum_{t=1}^T\abs{\inpr{g_t,u-w_t}}$, for any fixed $\varepsilon\in(0,1]$ (\cref{thm:ball-oco}).
For a general domain~$\mathcal K$, we choose the coordinates and the initial regularizer so that the regret bound depends on~$\mathcal K$ through its width~$\width(\mathcal K)$ (\cref{thm:main} in \cref{sec:width-extension}).
To obtain the small-loss regret bound without knowing~$\Delta_T(u)$, we mix the predictions of~%
$O\prn{\log\prn{t+2}}$ base learners with different fixed $\varepsilon$ values.
In doing so, we rescale subgradients sent to base learners, which lets us cancel the mixing error in the proof of \cref{thm:adaptive-known-width} (see \cref{sec:adaptation}).\looseness=-1

\paragraph{Applications.}
We apply the algorithm that achieves the bound in \cref{thm:adaptive-known-width} to \emph{contextual search}
\citep{liu2021optimal,leme2022density}, where a learner observes a context and predicts its inner product with a fixed unknown vector. The learner then receives a bit indicating whether its prediction is too high or
too low. For independent flips with probability~$p<1/2$ and bias~%
$\kappa\coloneqq1-2p$, \citet[Section~5.2]{liu2021optimal}
bound the expected cumulative absolute error by~$O_\kappa\prn{d^{25}}$ and ask for a
polynomial-time algorithm with a horizon-independent~$\poly\prn{d}$ error bound.
When the unknown vector and the contexts lie in the Euclidean unit ball, our polynomial-time algorithm attains the minimax bound~$\Theta\prn{\sqrt d/\kappa^2}$ without knowing~$\kappa$, thereby resolving their open question. See \cref{sec:search}.

We also apply \cref{thm:adaptive-known-width} to OILO, where a learner recommends
an action and observes a user's choice under an unknown linear utility.
We allow the user's choices to be suboptimal and let~$D_T$ be their cumulative utility shortfall relative to optimal actions. In \cref{sec:oilo}, we give a regret bound of
$O\prn{GW+\sqrt{GW D_T}}$ relative to observed actions, without knowing~$D_T$.
Here~$G$ bounds action-set diameters and~$W$ bounds the width of the utility
domain.
For~$\mathcal K=\B_2^d$, $G=O\prn{1}$ and $D_T=0$, our bound recovers the~$O\prn{\sqrt d}$ bound of \citet{gupta2026optimal}.
Our result is more general: on the probability simplex, for example, our bound is
$O\prn{G\log\prn{d+1}+\sqrt{GD_T\log\prn{d+1}}}$.

\subsection{Related Work}\label{sec:related}
We focus here on the work most closely related to our work; see \cref{app:additional-related} for more discussions.

Small-loss bounds have their origins in prediction with expert advice \citep{littlestone1994weighted,freund1997decision}, with early results also for online linear regression under square loss \citep{cesabianchi1996quadratic}.
In OCO, \citet{srebro2010smoothness} obtain small-loss bounds for losses smooth and nonnegative on~$\R^d$ using projected gradient descent with a step size tuned to~$L_T^\star$; \citet[Theorem~2]{zhang2019adaptive} remove the need for this prior knowledge.
For general nonnegative convex losses, \citet{leme2022density} bound the learner's cumulative loss by~$\prn{1+O\prn{\eps}}L_T^\star$ plus a term logarithmic in~$T$, for fixed small~$\eps>0$.
Recently, \citet{mayo2026minimax} establishes minimax small-loss bounds for OLO on~$\ell_p$ balls for fixed~$2\le p<\infty$ and loss vectors of dual norm at most one, including a dimension-free guarantee on the Euclidean ball.
Note that the guarantee for linear losses does not directly extend to convex losses, as linearization can increase the shifted comparator loss (\cref{app:cone-small-loss}).
\Cref{tab:small-loss} summarizes these OCO and OLO guarantees together with ours on the Euclidean unit ball.
Our bound for this setting is minimax optimal in its dependence on~$d$ and~$\Delta_T^\star$ (\cref{thm:minimax} in \cref{app:lower}).\looseness=-1

\begin{table}[tb]
\centering
\caption{Small-loss bounds against the best fixed decision on~$\mathcal K=\B_2^d$, with gradient or subgradient norms bounded by~$G=1$.
All dependence on~$d$ is shown in the bounds; the smoothness parameter is omitted as a fixed constant.
The notation~$O_\eps$ hides constants depending on~$\eps$.
Feedback~$g_t$ and~$\nabla f_t(w_t)$ denote a subgradient and a gradient at~$w_t$, respectively.}\label{tab:small-loss}
{\footnotesize
\begin{tabular*}{\linewidth}{m{.369\linewidth}@{\extracolsep{\fill}}m{.243\linewidth}@{\hspace{.016\linewidth}}m{.098\linewidth}@{\hspace{.016\linewidth}}m{.23\linewidth}}
\toprule
\raggedright Loss assumptions & \raggedright Regret bound & \raggedright Feedback & \raggedright Computation per round \tabularnewline
\midrule
\raggedright Smooth and nonnegative on~$\R^d$\newline \citep{srebro2010smoothness,zhang2019adaptive} &
\raggedright $O\prn{1+\sqrt{L_T^\star}}$ &
$\nabla f_t(w_t)$ &
\raggedright (Scale-free) projected gradient descent \tabularnewline
\addlinespace
\raggedright Nonnegative convex\newline \citep{leme2022density} &
\raggedright \mbox{$O\prn{\eps}L_T^\star+O_\eps\prn{d\log\prn{1+T}}$}\newline for fixed small~$\eps>0$ &
$g_t$ &
\raggedright Density updates and centroid computation \tabularnewline
\addlinespace
\raggedright Linear \citep{mayo2026minimax} &
\raggedright $O\prn{1+\sqrt{\Delta_T^\star}}$ &
\mbox{Loss vector} &
\raggedright FTRL tuned by shifted loss \tabularnewline
\addlinespace
\rowcolor[gray]{0.95}
\raggedright General convex\newline (This work, \cref{thm:adaptive-known-width}) &
\raggedright $O\prn{\sqrt d+d^{1/4}\sqrt{\Delta_T^\star}}$ &
$g_t$ &
\raggedright Mixing of ONS-style base learners with revocation \tabularnewline
\bottomrule
\end{tabular*}}
\end{table}

Small-loss bounds are also studied under other structural assumptions and with access to offline optimization.
\citet{abernethy2019privacy} obtain expected small-loss bounds for oblivious sequences of bounded nonnegative convex losses with bounded Hessians of rank at most one, using objective perturbation in offline optimization.
\citet{wang2022adaptive} obtain expected small-loss bounds for a finite action set using an offline oracle and a perturbation matrix satisfying certain approximability and implementability conditions; see \cref{app:ftpl-comparison} for further discussions.

Our algorithm builds on the metric-update revocation of \citet{gupta2026optimal} for OILO, which refines prior metric-update methods without revocation \citep{sakaue2025online,cai2026efficient}.
In our OCO formulation, optimal feedback actions in OILO give~$N_T(u)=0$ for the true utility vector~$u$.
We extend their revocation mechanism to general OCO through an FTRL analysis and adapt to unknown~$N_T(u)$ by mixing multiple base learners.

\section{Preliminaries}\label{sec:results}
\textbf{Notation.}
We use~$\inpr{\cdot,\cdot}$ and~$\norm{\cdot}_2$ for the Euclidean inner product
and norm, and~$\brc{a}_+\coloneqq\max\set{a,0}$ for~$a\in\R$.
For symmetric matrices, $\succeq$ and~$\succ$ denote the positive semidefinite and positive definite orders.
For~$H\succ0$, write~$\norm{v}_H^2\coloneqq v^\top Hv$ for compatible vectors~$v$.
We use~$I_d$ and~$\tr$ to denote the identity matrix and trace.
For a nonempty closed convex set~$C$, $\Pi_C$ denotes the Euclidean projection onto~$C$.
For a nonempty bounded set~$C\subset\R^d$, write $\operatorname{diam}_2(C)\coloneqq\sup_{x,y\in C}\norm{x-y}_2$.
For~$1\le p\le\infty$ and an integer~$n\ge1$, let $\B_p^n\coloneqq\Set{v\in\R^n}{\norm{v}_p\le1}$ be the unit~$\ell_p$ ball.
The probability simplex is denoted by $\triangle_d\coloneqq\Set{w\in\R_+^d}{\sum_{i=1}^d w_i=1}$. 
Logarithms are natural unless a base is specified.

\subsection{Online Learning Protocol}\label{sec:protocol}
The learner is given the compact convex domain~$\mathcal K\subset\R^d$ and a bound~$G>0$, but need not know
the time horizon~$T$. 
In each round~$t=1,2,\ldots$,
the learner first chooses~$w_t\in\mathcal K$
using the previously observed feedback. The environment then chooses a
continuous convex loss~$f_t\colon\mathcal K\to\R$
and a subgradient~$g_t\in\partial f_t(w_t)$ with~$\norm{g_t}_2\le G$.
The learner incurs~$f_t(w_t)$ and observes only~$g_t$.
Both the loss and the choice of subgradient may depend on the history
and the current prediction. We bound the regret~$R_T(u)$
for all fixed comparators~$u\in\mathcal K$ and finite horizons~$T$.

We write $r_t(u)\coloneqq\inpr{g_t,w_t-u}$ for the linearized excess loss and
have $R_T^{\mathrm{lin}}(u)=\sum_{t=1}^T r_t(u)$. The sum of the negative parts of the linearized excess losses
$N_T(u)=\sum_{t=1}^T\brc{-r_t(u)}_+$ satisfies
\begin{equation}\label{eq:negative-gap}
N_T(u)\le\Delta_T(u),
\end{equation}
since $-r_t(u)\le f_t(u)-f_t(w_t)\le f_t(u)-\min_{\mathcal K}f_t$.\footnote{
For fixed pairs~$(w_t,g_t)$, the minimum of~$\Delta_T(u)$ over continuous convex
losses consistent with these subgradients equals~$N_T(u)$.
Indeed, the losses $f_t(v)\coloneqq\brc{\inpr{g_t,v-w_t}}_+$ for~$v\in\mathcal K$
satisfy~$g_t\in\partial f_t(w_t)$ and $\min_{\mathcal K}f_t=f_t(w_t)=0$,
attaining equality in~\eqref{eq:negative-gap} simultaneously for all~$u\in\mathcal K$.
}
From $|r_t(u)|=r_t(u)+2\brc{-r_t(u)}_+$, we have
\begin{equation}\label{eq:convex-transfer}
R_T^{\mathrm{abs}}(u)=R_T^{\mathrm{lin}}(u)+2N_T(u).
\end{equation}

\subsection{Maximal Subexponential Mean Width}\label{sec:width}
We express the domain dependence of our bounds using a width of~$\mathcal K$.
For a real random variable~$X$, we define the \emph{subexponential norm} by $\norm{X}_{\psi_1}\coloneqq\inf\Set{s>0}{\E\exp\prn{\abs{X}/s}\le2}$
\citep[Definition 2.8.4]{vershynin2026hdp}.
Using this norm, we define the \emph{maximal subexponential mean width} by
\begin{equation}\label{eq:width}
\width(\mathcal K)\coloneqq\sup_Z\Set*{
  \E\brc*{\sup_{w\in\mathcal K}\inpr{Z,w}}
}{
  \E Z=0,\norm{\inpr{v,Z}}_{\psi_1}\le1\text{ for all }v\in\B_2^d
}.
\end{equation}
Here the outer supremum is over~$\R^d$-valued random vectors~$Z$.
For~$d\ge2$, we have~%
$\width(\B_2^d)\asymp\sqrt d$ and $\width(\triangle_d)\asymp\log\prn{d+1}$,
with universal implicit constants (see \cref{app:explicit-initializations,app:simplex-extension}).

\Needspace{10\baselineskip}
\section{Curvature Revocation on the Ball}\label{sec:algorithm}\label{sec:analysis}
We focus on the Euclidean unit-ball domain and bound the linearized regret minus a fraction of the sum of absolute linearized excess losses.
\begin{theorem}\label{thm:ball-oco}
For any fixed~$\eps\in(0,1]$ and inputs with~$\norm{g_t}_2\le1$, the predictions of \cref{alg:ball-oco} on~$\mathcal K=\B_2^d$ satisfy
\begin{equation}\label{eq:ball-played-signed}
2R_T^{\mathrm{lin}}(u)-\eps R_T^{\mathrm{abs}}(u)
\le\frac{20}{\eps}\ceil*{\sqrt d}
\end{equation}
for all~$u\in\B_2^d$ and~$T\ge1$.
\end{theorem}
Under the OCO protocol, combining \cref{thm:ball-oco} with~\eqref{eq:convex-transfer}, $N_T(u)\le\Delta_T(u)$, and the convexity of each loss~$f_t$ gives $R_T(u)\le R_T^{\mathrm{lin}}(u)\le20\ceil{\sqrt d}/\eps+2\eps\Delta_T(u)$.
The term~$-\eps R_T^{\mathrm{abs}}(u)$ in~\eqref{eq:ball-played-signed} also appears in the bound for general domains. In \cref{sec:adaptation}, we use the fixed-parameter guarantee to combine learners with different values of~$\eps$.
We prove \cref{thm:ball-oco} in \crefrange{sec:discrete-step}{sec:movement-proof}.

\subsection{Outer Update and Curvature Revocation}\label{sec:discrete-step}
For convenience, we consider a learner who maintains an unconstrained iterate~$z_t\in\R^d$,
plays its projection~$w_t\coloneqq\Pi_{\mathcal K}(z_t)\in\mathcal K$,
and receives~$g_t$ there. Under the OCO protocol, $g_t$ is a subgradient at~$w_t$.
Following \citet[Theorem~2]{cutkosky2020parameterfree}, set~%
$n_t\coloneqq z_t-w_t$ and define the corrected input by
\begin{equation}\label{eq:ball-correction}
\widetilde g_t\coloneqq
\begin{cases}
g_t,&n_t=0,\\
g_t-\dfrac{\min\set{0,\inpr{g_t,n_t}}}{\norm{n_t}_2^2}n_t,&n_t\ne0.
\end{cases}
\end{equation}
The correction gives
$\inpr{g_t,w_t-u}\le\inpr{\widetilde g_t,z_t-u}$ for every~$u\in\mathcal K$
and~$\norm{\widetilde g_t}_2\le\norm{g_t}_2$; see \cref{app:projection} for the proof.
We analyze the update on linear losses~$z\mapsto\inpr{\widetilde g_t,z}$
over~$\R^d$. Allowing arbitrary inputs~$\widetilde g_t$ lets us use the same update for the transformed gradients on general domains in \cref{sec:width-extension}.

The learner updates the iterate based on the FTRL framework.
The update starts at~$z_1\coloneqq0$ with an initial regularizer $\Psi_1(z)\coloneqq\norm{z}_B^2/2$, where $B\coloneqq\prn{10\ceil{\sqrt d}/\eps}I_d$. For each round~$t$, let~$\Psi_t$ be the cumulative regularizer at its start and define the objective
\begin{equation}\label{eq:ball-ftrl}
F_t(z)\coloneqq\Psi_t(z)+\sum_{i=1}^{t-1}\inpr{\widetilde g_i,z}.
\end{equation}
The iterates~$z_t$ generated in \cref{alg:ball-oco} satisfy
$z_t=\argmin_{z\in\R^d}F_t(z)$, as verified in \cref{sec:movement-properties}.
A term added in round~$i$ has direction~$q_i\coloneqq\widetilde g_i$ and coefficient~$\gamma_i>0$.
Writing~$m_i\coloneqq\inpr{q_i,z_{i+1}}$ for the value of~$\inpr{q_i,z}$ at the point where the term is added, we define
\begin{equation}\label{eq:ball-quadratic}
Q_i(z)\coloneqq\frac{\gamma_i}{2}\prn{\inpr{q_i,z}-m_i}^2.
\end{equation}
It is \emph{active} until revoked and has a corresponding threshold~$h_i>0$ specified later.
We call~$\inpr{q_i,z}-m_i$ the \emph{residual} of~$Q_i$ at~$z\in\R^d$.
We store the active terms in the indexed family
$\mathcal Q\coloneqq\prn{q_i,m_i,\gamma_i,h_i}_{i\in\mathcal A}$,
where~$\mathcal A$ is its current index set. Initially, set~$\mathcal Q_1\coloneqq\varnothing$.
For a differentiable convex function~$Q$, define its \emph{Bregman divergence} by
\begin{equation}\label{eq:ball-bregman}
D_Q(z,a)\coloneqq Q(z)-Q(a)-\inpr{\nabla Q(a),z-a}.
\end{equation}
To \emph{revoke}~$Q_i$ at~$a\in\R^d$, we delete entry~$i$ from~$\mathcal Q$
and subtract~$D_{Q_i}(\cdot,a)$ from the current cumulative regularizer~$\Psi$:
\begin{equation}\label{eq:ball-cancellation}
\Psi(z)\leftarrow\Psi(z)-D_{Q_i}(z,a),\quad\text{where}\;\; D_{Q_i}(z,a)=\frac{\gamma_i}{2}\inpr{q_i,z-a}^2.
\end{equation}
The point~$a$ remains fixed. In~$\Psi$, $Q_i(z)$ is replaced by its tangent function $Q_i(a)+\inpr{\nabla Q_i(a),z-a}$.
Thus revocation preserves the regularizer's value and gradient at~$a$ and removes curvature~%
$\gamma_iq_iq_i^\top$ from the Hessian. 
This is the counterpart of
\emph{stretch revocation} in \citet{gupta2026optimal}.

Each round of \cref{alg:ball-oco} calls \cref{alg:move-revoke} to update the iterate
and revoke terms before any addition.
At the start of round~$t$, we have the regularizer~$\Psi_t$, active family~%
$\mathcal Q_t$ indexed by~$\mathcal A_t$, and Hessian
$H_t\coloneqq\nabla^2\Psi_t=B+\sum_{i\in\mathcal A_t}\gamma_iq_iq_i^\top$.
Within \cref{alg:move-revoke}, $\Psi,H,\mathcal Q$ denote their current values.
For a nonzero input~$\widetilde g_t$, start at~$z\coloneqq z_t$ with remaining input weight~$\alpha\coloneqq1$.
Each step uses the direction~$\widetilde g_{t,H}\coloneqq H^{-1}\widetilde g_t$.
For the current active index set~$\mathcal A$, define $L\coloneqq\max_{i\in\mathcal A}\abs{\inpr{q_i,H^{-1}\widetilde g_t}}/h_i$, taking~$L=0$ when~$\mathcal A=\varnothing$.
The weight used in the step is~$\alpha_{\mathrm{step}}\coloneqq\alpha$ if~$L=0$, and $\alpha_{\mathrm{step}}\coloneqq\min\set{\alpha,2/L}$ if~$L>0$.
We update
\begin{equation}\label{eq:movement-step-update}
z\leftarrow z-\alpha_{\mathrm{step}}\widetilde g_{t,H}
\quad\text{and}\quad\alpha\leftarrow\alpha-\alpha_{\mathrm{step}}.
\end{equation}
At the new point~$z$, revoke each active term~$i$ with~$\abs{\inpr{q_i,z}-m_i}\ge h_i$ using~\eqref{eq:ball-cancellation}, and update~$H\leftarrow H-\gamma_iq_iq_i^\top$ for each revoked term.
Repeat while~$\alpha>0$, performing the revocations even on the final step, and return~$z_{t+1}\coloneqq z$.
For~$\widetilde g_t=0$, return~$z_{t+1}\coloneqq z_t$ with the regularizer, Hessian, and active family unchanged.
Write~$\Psi_t^-,H_t^-,\mathcal Q_t^-$ for the returned values before addition;
after the addition step, these three objects have index~$t+1$.
The returned scalar and, for~$s_t^2>0$, the new parameters are\looseness=-1
\begin{equation}\label{eq:parameters}
s_t^2\coloneqq\widetilde g_t^\top\prn{H_t^-}^{-1}\widetilde g_t,\qquad
\gamma_t\coloneqq\frac{\eps^2}{48s_t^2}\qquad\text{and}\qquad h_t\coloneqq\frac{8s_t^2}{\eps}.
\end{equation}
A new term is added only if~$s_t^2>\zeta_t$, where~$\zeta_t\ge0$.
In that case, set~$\Psi_{t+1}\coloneqq\Psi_t^-+Q_t$ and form~$\mathcal Q_{t+1}$ by appending~$\prn{q_t,m_t,\gamma_t,h_t}$ as entry~$t$ to~$\mathcal Q_t^-$.
Otherwise, set~$\Psi_{t+1}\coloneqq\Psi_t^-$ and~$\mathcal Q_{t+1}\coloneqq\mathcal Q_t^-$.
The Hessian is~$H_{t+1}=\nabla^2\Psi_{t+1}$ in both cases.
We take~$\zeta_t=0$ here for simplicity; \cref{app:ball-implementation} uses positive thresholds to limit
the number of active terms and reduce computation.

\begin{figure}
\noindent\begin{minipage}[t]{0.48\textwidth}
\begin{algorithm2e}[H]
\small
\caption{Main Procedure on the Ball}\label{alg:ball-oco}
\KwIn{$\eps\in(0,1]$.}
\KwOut{Predictions~$w_1,w_2,\ldots\in\B_2^d$.}
Set $B\leftarrow\prn{10\ceil{\sqrt d}/\eps}I_d$, $z_1\leftarrow0$, $\mathcal Q_1\leftarrow\varnothing$\;
Set~$\Psi_1(z)\leftarrow\norm{z}_B^2/2$ and~$H_1\leftarrow B$\;
\For{$t=1,2,\ldots$}{
Set~$w_t\leftarrow\Pi_{\mathcal K}(z_t)$ and play~$w_t$\;
Observe~$g_t$ and compute~$\widetilde g_t$ by~\eqref{eq:ball-correction}\;
Obtain $z_{t+1},s_t^2,\Psi_t^-,H_t^-,\mathcal Q_t^-$ by calling\newline
\textsc{MoveAndRevoke}$\prn{z_t,\widetilde g_t,\Psi_t,H_t,\mathcal Q_t}$\;
$\prn{\Psi_{t+1},H_{t+1},\mathcal Q_{t+1}}\leftarrow\prn{\Psi_t^-,H_t^-,\mathcal Q_t^-}$\;
\If{$s_t^2>0$}{
Set~$q_t\leftarrow\widetilde g_t$, $m_t\leftarrow\inpr{q_t,z_{t+1}}$\;
Set~$\gamma_t\leftarrow\eps^2/\prn{48s_t^2}$, $h_t\leftarrow8s_t^2/\eps$\;
Set $\Psi_{t+1}\leftarrow\Psi_{t+1}+Q_t$\;
Add~$\prn{q_t,m_t,\gamma_t,h_t}$ to~$\mathcal Q_{t+1}$ as entry~$t$\;
Set $H_{t+1}\!\leftarrow\! H_{t+1}\!+\!\gamma_tq_tq_t^\top$\;}}
\end{algorithm2e}
\end{minipage}\hfill
\begin{minipage}[t]{0.50\textwidth}
\begin{algorithm2e}[H]
\small
\caption{\textsc{MoveAndRevoke}}\label{alg:move-revoke}
\KwIn{$z_t,\widetilde g_t,\Psi,H,\mathcal Q=(q_i,m_i,\gamma_i,h_i)_{i\in\mathcal A}$.}
\KwOut{$z_{t+1},s_t^2,\Psi,H,\mathcal Q$.}
\If{$\widetilde g_t=0$}{\Return $\prn{z_t,0,\Psi,H,\mathcal Q}$\;}
Set~$z\leftarrow z_t$, $\alpha\leftarrow1$\;
\While{$\alpha>0$}{
$\widetilde g_{t,H}\leftarrow H^{-1}\widetilde g_t$\;
$L\leftarrow\max_{i\in\mathcal A}\abs{\inpr{q_i,\widetilde g_{t,H}}}/h_i$\;
{\renewcommand{\arraystretch}{0.9}$\alpha_{\mathrm{step}}\leftarrow\Bigl\{\begin{array}{@{}l@{\quad}l@{}}\alpha & \text{if }L=0\\ \min\set{\alpha,2/L} & \text{if }L>0\end{array}$}\;
$z\leftarrow z-\alpha_{\mathrm{step}}\widetilde g_{t,H}$,\, $\alpha\leftarrow\alpha-\alpha_{\mathrm{step}}$\;
\ForEach{$i\in\mathcal A$ with~$\abs{\inpr{q_i,z}-m_i}\ge h_i$}{
Set $\Psi\leftarrow\Psi\!-\!D_{Q_i}(\cdot,z)$, $H\leftarrow H\!-\!\gamma_iq_iq_i^\top$\; Delete entry~$i$ from~$\mathcal Q$ and index $i$ from~$\mathcal A$\;}}
Set~$z_{t+1}\leftarrow z$, $s_t^2\leftarrow\widetilde g_t^\top H^{-1}\widetilde g_t$\;
\Return $\prn{z_{t+1},s_t^2,\Psi,H,\mathcal Q}$\;
\end{algorithm2e}
\end{minipage}
\end{figure}

The above update adds rank-one curvature as in Online Newton Step (ONS)
\citep[see also \citealp{orabona2026online}, Section~7.10]{hazan2007logarithmic}, but also removes curvature by revocation.
The standard ONS analysis uses an elliptical potential argument, and the resulting regret bound retains a $\log T$ term.
We use revocation to avoid this term in our regret bound, as detailed in \cref{sec:lifetime}.

\subsection{Unconstrained Iterates and Inner Procedure}\label{sec:movement-properties}
We first analyze the unconstrained iterates $z_t$, taking $\widetilde g_t\in\R^d$ as an arbitrary input to the movement and addition steps  of \cref{alg:ball-oco} in each round. 
We return to the projected iterates $w_t$ in \cref{sec:lifetime}. 
For any~$B\succ0$, start with~$z_1\coloneqq0$, $\Psi_1(z)\coloneqq\norm{z}_B^2/2$, $H_1\coloneqq B$, and~$\mathcal Q_1\coloneqq\varnothing$.
In each round, we set
\begin{equation}\label{eq:unconstrained-update}
\prn{z_{t+1},s_t^2,\Psi_t^-,H_t^-,\mathcal Q_t^-}
\leftarrow\text{\textsc{MoveAndRevoke}}\prn{z_t,\widetilde g_t,\Psi_t,H_t,\mathcal Q_t}
\end{equation}
and then apply the addition step of \cref{alg:ball-oco} if~$s_t^2>0$.
Otherwise, we carry the returned regularizer, Hessian, and active family into the next round.
Writing $\widetilde r_t(u)\coloneqq\inpr{\widetilde g_t,z_t-u}$, we will upper-bound $\sum_{t=1}^T(2\widetilde r_t(u)-\eps\abs{\widetilde r_t(u)})$ using the following properties of \cref{alg:move-revoke}.

\begin{lemma}\label{lem:move-revoke}\label{lem:steps}
Suppose the following conditions hold:~%
 $B\succ0$; 
$z_t$ minimizes~$F_t$; 
the finite active family~%
$\mathcal Q_t$ has~$\gamma_i,h_i>0$ and~%
$\abs{\inpr{q_i,z_t}-m_i}<h_i$ for every~$i\in\mathcal A_t$.
Then, for any input~$\widetilde g_t$, \cref{alg:move-revoke} terminates and satisfies the
following properties. Let~$\mathcal A_t^-$ index the returned family~$\mathcal Q_t^-$,
$\mathcal J_t\coloneqq\mathcal A_t\setminus\mathcal A_t^-$, and
$F_t^-(z)\coloneqq\Psi_t^-(z)+\sum_{i=1}^t\inpr{\widetilde g_i,z}$.
\begin{enumerate}[label=(\roman*),ref=(\roman*),leftmargin=*,nosep]
\item\label[moveproperty]{item:move-revoke-curvature} Each~$i\in\mathcal J_t$ is revoked once at a point~$a_i$, and we have
\begin{equation}\label{eq:revocation-state}
\Psi_t^-(z)=\Psi_t(z)-\sum_{i\in\mathcal J_t}D_{Q_i}(z,a_i)
\qquad(z\in\R^d).
\end{equation}
The returned family is the restriction of~$\mathcal Q_t$ to~$\mathcal A_t^-$,
and the returned Hessian satisfies $H_t^-=\nabla^2\Psi_t^-=B+\sum_{i\in\mathcal A_t^-}\gamma_iq_iq_i^\top$.
\item\label[moveproperty]{item:move-revoke-optimality} The returned point satisfies~$\nabla F_t^-(z_{t+1})=0$, and the returned scalar is~$s_t^2$ from~\eqref{eq:parameters}.
\item\label[moveproperty]{item:move-revoke-decrease} For the linear function~$z\mapsto\inpr{\widetilde g_t,z}$,
define its decrease from~$z_t$ to~$z_{t+1}$ by
$\delta_t\coloneqq\inpr{\widetilde g_t,z_t-z_{t+1}}$. The decreases in the objective and this linear function during the call satisfy
\begin{equation}\label{eq:movement-properties}
0\le F_t(z_t)+\inpr{\widetilde g_t,z_t}-F_t^-(z_{t+1})\le s_t^2/2
\quad\text{and}\quad
0\le\delta_t\le s_t^2.
\end{equation}
\item\label[moveproperty]{item:move-revoke-thresholds} A revoked term has
$h_i\le\abs{\inpr{q_i,a_i}-m_i}<3h_i$; a surviving term has
$\abs{\inpr{q_i,z_{t+1}}-m_i}<h_i$.
\end{enumerate}
\end{lemma}
We prove the lemma in \cref{sec:movement-proof}. \Cref{item:move-revoke-optimality} gives optimality at the returned point. Since~$Q_t$
and its gradient vanish at~$z_{t+1}$, adding~$Q_t$ preserves optimality.
To apply \cref{lem:move-revoke} in round~$t+1$, we also require
$\abs{\inpr{q_i,z_{t+1}}-m_i}<h_i$ for every~$i\in\mathcal A_{t+1}$.
Any new term satisfies this inequality because its residual is zero,
and \cref{item:move-revoke-thresholds} gives it for every surviving term.
Starting from~$z_1=0$, induction therefore establishes that~$z_t$ minimizes~$F_t$
and that \cref{lem:move-revoke} applies in every round.

\subsection{Regret Guarantee on the Ball via Cancellation and Trace Bound}\label{sec:lifetime}
For inputs $\tilde g_t$ with $\norm{\tilde g_t}_2\le1$, we bound $\sum_{t=1}^T\prn{2\widetilde r_t(u)-\eps\abs{\widetilde r_t(u)}}$ independently of~$T$ and transfer the result to $w_t\in\mathcal K$.
Fix unconstrained $u\in\R^d$ and let $\widetilde r_t\coloneqq\widetilde r_t(u)$.
Define the \emph{stability} in round~$t$ by\looseness=-1
\begin{equation}\label{eq:ball-stab}
\mathrm{Stab}_t\coloneqq F_t(z_t)+\inpr{\widetilde g_t,z_t}-F_{t+1}(z_{t+1}).
\end{equation}
We have the FTRL regret identity \citep[Lemma~7.1 and Remark~7.2]{orabona2026online}: 
\begin{equation}\label{eq:ball-pen-stab}
\sum_{t=1}^T\widetilde r_t
=\Psi_{T+1}(u)+\sum_{t=1}^T\mathrm{Stab}_t
-\brc{F_{T+1}(u)-F_{T+1}(z_{T+1})}.
\end{equation}
This identity requires no pointwise monotonicity of~$\Psi_t$ across~$t$.
Summing the definition of~$\mathrm{Stab}_t$,
using~$F_1(z_1)=0$, and subtracting~$\sum_{t=1}^T\inpr{\widetilde g_t,u}$
yields the FTRL regret identity.
Any regularization term
added at~$z_{t+1}$ vanishes there, so $F_{t+1}(z_{t+1})=F_t^-(z_{t+1})$ holds.
\Cref{item:move-revoke-decrease} therefore gives
\begin{equation}\label{eq:ball-stability-displacement}
0\le\mathrm{Stab}_t\le s_t^2/2\quad\text{and}\quad 0\le\delta_t\le s_t^2.
\end{equation}
Summing~\eqref{eq:ball-stability-displacement} gives $\sum_{t=1}^T\mathrm{Stab}_t\le\frac12\sum_{t=1}^Ts_t^2$.
Since $s_t^2=\widetilde g_t^\top\prn{H_t^-}^{-1}\widetilde g_t$, a natural idea would be to bound the sum on the right by an elliptical potential, as in the standard ONS analysis.
Doing so, however, would leave a $\log T$ term in our regret bound.
Also, without revocation, $\Psi_{T+1}(u)$ accumulates the nonnegative regularization terms~$Q_i(u)$ and can grow with~$T$.
To obtain a bound independent of~$T$, we evaluate~$\Psi_{T+1}(u)$ and~$\sum_{t=1}^T\mathrm{Stab}_t$ together with the negative term
$-\brc{F_{T+1}(u)-F_{T+1}(z_{T+1})}$.

Let~$\mathcal I_T$ and~$\mathcal C_T$ index the active and revoked terms, respectively,
after rounds~$1,\ldots,T$. They partition the nonzero-input rounds.
For~$i\in\mathcal I_T\cup\mathcal C_T$, define~$z_i^{\mathrm{end}}$ as the
actual revocation point if~$i\in\mathcal C_T$, and~$z_{T+1}$ otherwise.
Each created~$Q_i$ is added to~$\Psi$, and revocation subtracts its divergence.
\Cref{item:move-revoke-curvature} therefore gives, for every~$z\in\R^d$,
\begin{equation}\label{eq:ball-psi-terminal}
\Psi_{T+1}(z)=\frac12\norm{z}_B^2+\sum_{i\in\mathcal I_T\cup\mathcal C_T}Q_i(z)
-\sum_{i\in\mathcal C_T}D_{Q_i}(z,z_i^{\mathrm{end}}).
\end{equation}
As proved in \cref{sec:movement-properties}, $z_{T+1}$ minimizes~$F_{T+1}$ over~$\R^d$.
Combining this optimality with~\eqref{eq:ball-psi-terminal} gives
\begin{equation}\label{eq:ball-terminal-derivatives}
\nabla F_{T+1}(z_{T+1})=0\quad\text{and}\quad \nabla^2F_{T+1}=H_{T+1}=B+\sum_{i\in\mathcal I_T}\gamma_iq_iq_i^\top.
\end{equation}
Therefore, Taylor's formula, with $D_{Q_i}(u,z)=\gamma_i\inpr{q_i,u-z}^2/2$, gives
\begin{equation}\label{eq:ball-terminal-gap}
F_{T+1}(u)-F_{T+1}(z_{T+1})
=\frac12\norm{u-z_{T+1}}_{H_{T+1}}^2
=\frac12\norm{u-z_{T+1}}_B^2+\sum_{i\in\mathcal I_T}D_{Q_i}(u,z_{T+1}).
\end{equation}
Substituting these expressions into the FTRL regret identity~\eqref{eq:ball-pen-stab} gives
\begin{equation}\label{eq:ball-split}
\sum_{t=1}^T\widetilde r_t+\tfrac12\norm{u-z_{T+1}}_B^2
=\tfrac12\norm{u}_B^2+\sum_{i\in\mathcal I_T\cup\mathcal C_T}
\brc{Q_i(u)-D_{Q_i}(u,z_i^{\mathrm{end}})+\mathrm{Stab}_i}.
\end{equation}
Zero-input rounds have zero linearized excess loss and zero stability.
The following lemma bounds the sum on the right of~%
\eqref{eq:ball-split} by
$O\prn{\eps\sum_{i=1}^T\abs{\widetilde r_i}
+\sum_{i\in\mathcal I_T}s_i^2}$.
\begin{restatable}{lemma}{lifetimebound}\label{lem:lifetime}
For the unconstrained update in \cref{sec:movement-properties} with~%
$\eps\in(0,1]$ and~$B\succ0$, set~$\tau\coloneqq\eps^2/48$.
Then we have, for every input sequence, %
$u\in\R^d$, and~$T\ge1$,
\begin{equation}\label{eq:ball-lifetime}
2\sum_{t=1}^T\widetilde r_t-\eps\sum_{t=1}^T\abs{\widetilde r_t}
+\norm{u-z_{T+1}}_B^2
\le\norm{u}_B^2+\prn{1+\tau}\sum_{i\in\mathcal I_T}s_i^2.
\end{equation}
\end{restatable}
\begin{proof}
For each created term~$i$, set $\rho_i\coloneqq\inpr{q_i,z_i^{\mathrm{end}}}-m_i$,
and recall~$\widetilde r_i=\inpr{q_i,z_i-u}$ and~%
$\delta_i=\inpr{q_i,z_i-z_{i+1}}$.
Using~$m_i=\inpr{q_i,z_{i+1}}$ in the definitions of~$Q_i$ and its Bregman divergence gives
\begin{equation}\label{eq:ball-quadratic-values}
Q_i(u)=\frac{\gamma_i}{2}\prn{\delta_i-\widetilde r_i}^2\quad\text{and}\quad D_{Q_i}(u,z_i^{\mathrm{end}})=
\frac{\gamma_i}{2}\inpr{q_i,u-z_i^{\mathrm{end}}}^2=
\frac{\gamma_i}{2}\prn{\delta_i-\widetilde r_i-\rho_i}^2.
\end{equation}
Subtracting the two expressions above and adding~$\mathrm{Stab}_i$ yields
\begin{equation}\label{eq:ball-bregman-difference}
Q_i(u)-D_{Q_i}(u,z_i^{\mathrm{end}})+\mathrm{Stab}_i
=\frac{\gamma_i}{2}\brc{\delta_i^2-\prn{\delta_i-\rho_i}^2}
+\mathrm{Stab}_i-\gamma_i\rho_i\widetilde r_i.
\end{equation}
Recall~$\gamma_i s_i^2=\tau=\eps^2/48$ and~$h_i=8s_i^2/\eps$
by~\eqref{eq:parameters}, and $0\le\delta_i\le s_i^2$ and $0\le\mathrm{Stab}_i\le s_i^2/2$ by~\eqref{eq:ball-stability-displacement}.
\Cref{item:move-revoke-thresholds} gives~$\abs{\rho_i}<3h_i$ for $i<T$, and $\rho_i=0$ if $i=T$, hence $-\gamma_i\rho_i\widetilde r_i\le3\gamma_i h_i\abs{\widetilde r_i}
=\eps\abs{\widetilde r_i}/2$.

For~$i\in\mathcal C_T$, \cref{item:move-revoke-thresholds} gives~$\abs{\rho_i}\ge h_i$, and~\eqref{eq:parameters} gives~$h_i=8s_i^2/\eps>2s_i^2$.
Since $\rho_i^2-2\rho_i\delta_i\ge\abs{\rho_i}\prn{\abs{\rho_i}-2s_i^2}
\ge h_i\prn{h_i-2s_i^2}$, we obtain
\begin{equation}\label{eq:ball-revoked-drop}
\frac{\gamma_i}{2}\brc{\delta_i^2-\prn{\delta_i-\rho_i}^2}+\mathrm{Stab}_i
\le-\frac{\gamma_i h_i}{2}\prn{h_i-2s_i^2}+\frac{s_i^2}{2}
=\frac{\eps-1}{6}s_i^2\le0.
\end{equation}
For~$i\in\mathcal I_T$, we omit~$-\gamma_i\prn{\delta_i-\rho_i}^2/2$ and use
$\gamma_i\delta_i^2/2+\mathrm{Stab}_i\le\prn{1+\tau}s_i^2/2$.
Adding~$\eps\abs{\widetilde r_i}/2$ to the upper bound for each case, summing in~\eqref{eq:ball-split},
and multiplying by~$2$ proves the lemma.
\end{proof}

To bound~$\sum_{i\in\mathcal I_T}s_i^2$ in \cref{lem:lifetime}, we follow \citet{gupta2026optimal} and track the trace of the inverse Hessian as we add the surviving terms in order of creation.
\begin{lemma}\label{lem:ball-surviving-trace}
For the unconstrained update in \cref{sec:movement-properties} with~$\eps\in(0,1]$, $B\succ0$, and~$\norm{\widetilde g_t}_2\le1$ for every~$t$, let~$\tau\coloneqq\eps^2/48$. For every~$T\ge1$, we have
\begin{equation}\label{eq:survivor-simple}
\sum_{i\in\mathcal I_T}s_i^2\le\frac{1+\tau}{\tau}\tr(B^{-1}).
\end{equation}
\end{lemma}
\begin{proof}
For each~$i\in\mathcal I_T$, define
$\overline H_i\coloneqq B+\sum_{j\in\mathcal I_T,\,j<i}\gamma_jq_jq_j^\top$
and~$\overline s_i^2\coloneqq q_i^\top\overline H_i^{-1}q_i$.
Every term in this sum was present just before~$Q_i$ was created, so
\cref{item:move-revoke-curvature} gives~$0\prec\overline H_i\preceq H_i^-$.
Inverting reverses this order, so we have
$\overline s_i^2\ge q_i^\top\prn{H_i^-}^{-1}q_i=s_i^2$ by~\eqref{eq:parameters}.
Taking traces in the Sherman--Morrison formula gives
\begin{equation}\label{eq:ball-trace-drop}
\tr(\overline H_i^{-1})-\tr\prn*{\prn{\overline H_i+\gamma_iq_iq_i^\top}^{-1}}
=\frac{\gamma_i\norm{\overline H_i^{-1}q_i}_2^2}{1+\gamma_i\overline s_i^2}
\ge\frac{\gamma_i\inpr{q_i,\overline H_i^{-1}q_i}^2}{1+\gamma_i\overline s_i^2}
\ge\frac{\tau}{1+\tau}s_i^2.
\end{equation}
The first inequality uses the Cauchy--Schwarz inequality and~$\norm{q_i}_2\le1$, and the last uses~$\overline s_i^2\ge s_i^2$ and~$\gamma_i s_i^2=\tau$.
Summing this inequality over~$i\in\mathcal I_T$ gives
$\frac{\tau}{1+\tau}\sum_{i\in\mathcal I_T}s_i^2
\le\tr(B^{-1})-\tr(H_{T+1}^{-1})\le\tr(B^{-1})$,
proving the claimed trace bound.
\end{proof}
Substituting~\eqref{eq:survivor-simple} into \cref{lem:lifetime} and dropping the nonnegative term~$\norm{u-z_{T+1}}_B^2$ gives, for every~$u\in\R^d$ and~$T\ge1$,
\begin{equation}\label{eq:ball-balance}
2\sum_{t=1}^T\widetilde r_t(u)-\eps\sum_{t=1}^T\abs{\widetilde r_t(u)}
\le\norm{u}_B^2+\frac{100}{\eps^2}\tr(B^{-1}),
\end{equation}
where we used $\prn{1+\tau}^2/\tau=48/\eps^2+2+\eps^2/48\le100/\eps^2$.

We now prove \cref{thm:ball-oco} by transferring~\eqref{eq:ball-balance} to the predictions~$w_t=\Pi_{\mathcal K}(z_t)$ using the gradient correction of \citet[Theorem~2]{cutkosky2020parameterfree}.
\begin{proof}[\cref{thm:ball-oco}]
For every~$u\in\B_2^d$, the corrected input in~\eqref{eq:ball-correction} satisfies~$\norm{\widetilde g_t}_2\le\norm{g_t}_2\le1$ and~$r_t(u)\le\widetilde r_t(u)$.
Since~$a\mapsto2a-\eps\abs{a}$ is increasing, the left-hand side of~\eqref{eq:ball-balance} bounds $2R_T^{\mathrm{lin}}(u)-\eps R_T^{\mathrm{abs}}(u)$ from above.
\Cref{alg:ball-oco} uses~$B=\lambda I_d$ with~$\lambda\coloneqq10\ceil{\sqrt d}/\eps$. Thus the right-hand side is at most $\lambda+100d/\prn{\lambda\eps^2}\le20\ceil{\sqrt d}/\eps$, proving~\eqref{eq:ball-played-signed}.
\end{proof}

\subsection{Proof of Movement Properties}\label{sec:movement-proof}\label{sec:revocation}
We prove that \cref{alg:move-revoke} terminates and satisfies \crefrange{item:move-revoke-curvature}{item:move-revoke-thresholds}.
Recall~$s_t^2=\widetilde g_t^\top\prn{H_t^-}^{-1}\widetilde g_t$.
Without revocation, $H_t^-=H_t$ holds, and~%
$z_{t+1}=z_t-H_t^{-1}\widetilde g_t$ minimizes~$F_t+\inpr{\widetilde g_t,\cdot}$.
Then, the quantitative bounds in \cref{item:move-revoke-decrease} are intuitive:
$F_t(z_t)+\inpr{\widetilde g_t,z_t}-F_t^-(z_{t+1})=s_t^2/2$ and~$\delta_t=s_t^2$ hold, attaining both upper bounds.
The following proof confirms \crefrange{item:move-revoke-curvature}{item:move-revoke-thresholds} in the presence of revocation.\looseness=-1
\begin{proof}[\cref{lem:move-revoke}]
Revoking term~$i$ at~$a_i$ subtracts~$D_{Q_i}(\cdot,a_i)$ from~$\Psi$
and~$\gamma_iq_iq_i^\top$ from~$H$, and removes entry~$i$ from~$\mathcal Q$.
Summing these updates proves \cref{item:move-revoke-curvature}.

Fix a round~$t$ with~$\widetilde g_t\ne0$. Write~$z_{t,k}$ and~$\alpha_k$ for the iterate and remaining weight before step~$k$, and~$z_{t,k+1}$ and~$\alpha_{k+1}$ for their values after it.
The step uses weight $\alpha_{\mathrm{step}}=\alpha_k-\alpha_{k+1}>0$.
Since~$\alpha_{\mathrm{step}}L\le2$ in \cref{alg:move-revoke}, every term~$i$ active before step~$k$ satisfies
\begin{equation}\label{eq:step-cap}
\abs{\inpr{q_i,z_{t,k+1}-z_{t,k}}}\le 2h_i.
\end{equation}
Before the step, we have $\abs{\inpr{q_i,z_{t,k}}-m_i}<h_i$. By the triangle inequality, the residual at~$z_{t,k+1}$ has absolute value less than~$3h_i$.
The algorithm revokes term~$i$ if this value is at least~$h_i$, proving \cref{item:move-revoke-thresholds}.
We next verify termination. On a nonfinal step, we have~$\alpha_{\mathrm{step}}L=2$. Each maximizing index~$i$ satisfies $\abs{\inpr{q_i,z_{t,k+1}-z_{t,k}}}=2h_i$. Its residual at~$z_{t,k+1}$ then has absolute value greater than~$h_i$, and the algorithm revokes it.
Every nonfinal step therefore removes at least one term from a finite family, and no terms are added during the call. Hence the loop terminates.

To prove \crefrange{item:move-revoke-optimality}{item:move-revoke-decrease}, let~$K_t$ be the number of steps. We have~$\alpha_1=1$ and~$\alpha_{K_t+1}=0$.
Let~$F_{t,k}$ and~$H_{t,k}$ denote the objective and its Hessian before step~$k$, with index~$K_t+1$ denoting the state after the last revocation test and before addition. In particular, we have~$F_{t,1}=F_t+\inpr{\widetilde g_t,\cdot}$ and~$F_{t,K_t+1}=F_t^-$.
Initially, we have~$\nabla F_{t,1}(z_{t,1})=\widetilde g_t$.
Since~$F_{t,k}$ has Hessian~$H_{t,k}$, the move in step~$k$ changes its gradient by~$-\prn{\alpha_k-\alpha_{k+1}}\widetilde g_t$, and revocation preserves it at~$z_{t,k+1}$. By induction, for~$1\le k\le K_t$ we have
\begin{equation}\label{eq:ball-substep-invariant}
\nabla F_{t,k}(z_{t,k})=\alpha_k\widetilde g_t\quad\text{and}\quad z_{t,k+1}-z_{t,k}=-\prn{\alpha_k-\alpha_{k+1}}H_{t,k}^{-1}\widetilde g_t.
\end{equation}
The gradient relation holds also at~$k=K_t+1$, where~%
$\alpha_{K_t+1}=0$ and~$z_{t,K_t+1}=z_{t+1}$. This proves \cref{item:move-revoke-optimality}.
For \cref{item:move-revoke-decrease}, put~$s_{t,k}^2\coloneqq\widetilde g_t^\top H_{t,k}^{-1}\widetilde g_t$.
Revocation preserves the objective value at~$z_{t,k+1}$.
Evaluating the quadratic objective~$F_{t,k}$ at~$z_{t,k+1}$ using~%
\eqref{eq:ball-substep-invariant} therefore gives
\begin{equation}\label{eq:ball-step-drop}
F_{t,k}(z_{t,k})-F_{t,k+1}(z_{t,k+1})
=\frac{\alpha_k^2-\alpha_{k+1}^2}{2}s_{t,k}^2.
\end{equation}
Within the loop, only revocations change the Hessian, so we have~%
$H_{t,k}\succeq H_t^-\succ0$ and~$0<s_{t,k}^2\le s_t^2$.
Summing over~$k$ and using
$\sum_{k=1}^{K_t}(\alpha_k^2-\alpha_{k+1}^2)=1$
proves the first bound in \cref{item:move-revoke-decrease}.
Taking the inner product of~$z_{t,k}-z_{t,k+1}$ with~$\widetilde g_t$ in~\eqref{eq:ball-substep-invariant} and summing gives
$0\le\delta_t=\sum_{k=1}^{K_t}\prn{\alpha_k-\alpha_{k+1}}s_{t,k}^2\le s_t^2$,
since $\sum_{k=1}^{K_t}\prn{\alpha_k-\alpha_{k+1}}=1$.
For zero input, neither the iterate nor the regularizer changes, so both decreases are zero.
\end{proof}

\subsection{Extension to General Domains and Computational Cost}\label{sec:width-extension}\label{sec:cost}
We extend the ball guarantee to general domains.
To this end, we choose coordinates adapted to the shape of~$\mathcal K$ and set the initial regularizer accordingly.
We run the movement, revocation, and addition steps in these coordinates with suitable addition thresholds.
Predictions are projected onto~$\mathcal K$, and the feedback is corrected as in \cref{sec:discrete-step} and transformed to these coordinates.
This gives the regret bound in \cref{thm:main}.
\Cref{app:geometry,app:core-proof} give the coordinate choices, initializations, and analysis.
\begin{theorem}\label{thm:main}
For any fixed~$\eps\in(0,1]$ and inputs with~$\norm{g_t}_2\le G$, the predictions of \cref{alg:oco} in \cref{app:projection} satisfy
\begin{equation}\label{eq:main-signed}
2R_T^{\mathrm{lin}}(u)-\eps R_T^{\mathrm{abs}}(u)
\le cG\width(\mathcal K)/\eps
\end{equation}
for all~$u\in\mathcal K$ and~$T\ge1$, where~$c$ is a universal constant.
\end{theorem}

We then briefly discuss the computational cost; see \cref{app:computation} for the computational model and details.
Let~$N$ be the number of coordinates in the chosen representation, $M_0$ the number of revocable initial regularization terms, and~$J_{\max}$ their maximum number per coordinate.
Through round~$T$, there are at most~$T$ additions and~$T+M_0$ revocations.
Since every nonfinal internal step revokes a term, there are at most~$2T+M_0$ internal steps.
The implementation maintains the inverse Hessian and checks the active terms using
\begin{equation}\label{eq:general-cost-overview}
O\prn*{\prn{T+M_0}N^2\eps^{-2}\brc{1+J_{\max}+\log\prn{T+2}}}
\end{equation}
arithmetic operations through round~$T$ for the updates in these coordinates. 
This count assumes the coordinate map and initial parameters are supplied and excludes projections, conversion between coordinates, and feedback correction.
On the Euclidean unit-ball domain, we have $N=d$ and~$M_0=J_{\max}=0$, so \cref{alg:oco} in \cref{app:projection} uses $O\prn{Td^2\eps^{-2}\log\prn{T+2}}$ operations.

\section{Adaptation to Unknown Comparator Loss}\label{sec:adaptation}
For fixed~$\eps$, \cref{thm:main} gives~$R_T^{\mathrm{lin}}(u)\lesssim GW/\eps+\eps N_T(u)$, where~$W\ge\width(\mathcal K)$. We optimize this tradeoff without knowing~$N_T(u)$ by mixing predictions from base learners with different fixed values of~$\eps$.
The algorithm observes only the subgradient~$g_t$ at the mixture prediction~$w_t$ on each round, and it achieves the regret bound in \cref{thm:adaptive-known-width}.
Below we sketch the proof of \cref{thm:adaptive-known-width} assuming exact mixture weights. \Cref{app:adapt} gives the proof for approximate weights, and \cref{app:adaptive-computation} discusses the arithmetic implementation and its computational cost.

\paragraph{Base Learners.}\leavevmode
For each integer~$i\ge0$, we start learner~$i$, a copy of \cref{alg:oco} in \cref{app:projection}, at round~$2^i$ with parameter~$\eps_i\coloneqq2^{-i}/8$ and input bound~$2G$. A learner is \emph{awake} from its first round onward. We form~$w_t$ as a convex combination of the awake learners' predictions~$w_t^{(j)}$. After observing $g_t$ at $w_t$, we update each awake learner~$j$ with a rescaled input~$g_t^{(j)}$ satisfying~$\norm{g_t^{(j)}}_2\le2G$; see below and \cref{app:master}.
Define $\Lambda_\theta(r)\coloneqq r-\theta\abs{r}$ for $r\in\R$ and $\theta\ge0$. For a learner~$i$ with~$2^i\le T$, \cref{thm:main} gives~$\sum_{t=2^i}^T\Lambda_{\eps_i/2}\prn{\inpr{g_t^{(i)},w_t^{(i)}-u}}\lesssim GW/\eps_i$. For the mixture, consider using the larger coefficient~$2\eps_i$ instead; the reason for this choice is clarified shortly. The identity %
\eqref{eq:convex-transfer} gives
\begin{equation}\label{eq:adaptive-signed-overview}
\sum_{t=1}^T\Lambda_{2\eps_i}(r_t(u))=R_T^{\mathrm{lin}}(u)-2\varepsilon_iR_T^{\mathrm{abs}}(u)=
\prn{1-2\eps_i}R_T^{\mathrm{lin}}(u)-4\eps_iN_T(u).
\end{equation}
Since $1-2\eps_i\ge3/4$, if~$\sum_{t=1}^T\Lambda_{2\eps_i}(r_t(u))\lesssim GW/\eps_i$, then~$R_T^{\mathrm{lin}}(u)\lesssim GW/\eps_i+\eps_iN_T(u)$ holds.

\paragraph{Comparison with $u$ via Learner $i$.}\leavevmode
For each awake learner~$j$, let $r_{t,j}\coloneqq\inpr{g_t,w_t-w_t^{(j)}}$. To form~$g_t^{(j)}$, we multiply~$g_t$ by $\prn{1-\eps_j/2}/\prn{1+\eps_j/2}$ if~$r_{t,j}>0$, by its reciprocal if~$r_{t,j}<0$, and by one otherwise. With these inputs, the comparison with $u$ through learner~$i$ on rounds~$2^i,\ldots,T$ gives
\begin{equation}\label{eq:adaptive-decomposition-overview}
\sum_{t=2^i}^{T}\Lambda_{2\eps_i}(r_t(u))
\le\decompterm{main}{A}{\textstyle\sum_{t=2^i}^{T}\Lambda_{\eps_i/2}(r_{t,i})}+\decompterm{main}{B}{\textstyle\sum_{t=2^i}^{T}\Lambda_{\eps_i/2}\prn{\inpr{g_t^{(i)},w_t^{(i)}-u}}}.
\end{equation}
Term \decompref{main}{A} compares the mixture with learner~$i$. Term \decompref{main}{B} compares learner~$i$ with $u$ on its rescaled inputs. For each round, the multiplier makes the two summands on the right cancel at~$r_t(u)=0$, and \cref{app:scalar} shows that their sum is at least~$\Lambda_{2\eps_i}(r_t(u))$ on either side of zero, giving~\eqref{eq:adaptive-decomposition-overview} after summation over $t$. Since $\norm{g_t^{(i)}}_2\le2G$, the base-learner guarantee in~\eqref{eq:main-signed} implies $\decompref{main}{B}\lesssim GW/\eps_i$. As we see below, using~$2\eps_i$ in~\eqref{eq:adaptive-decomposition-overview} allows us to subtract~$(\eps_i/2)\abs{r_{t,i}}$ from~$r_{t,i}$ in \decompref{main}{A}, and these subtractions cancel the error arising from mixing the learners' predictions.

\paragraph{Recursive Potential and Corrected Losses.}\leavevmode
The \emph{corrected loss} of an awake learner~$j$ at round~$s$ is $-\Lambda_{\eps_j/2}(r_{s,j})$. Let $\widetilde L_{t,j}\coloneqq-\sum_{s=2^j}^t\Lambda_{\eps_j/2}(r_{s,j})$ denote its cumulative corrected loss. At round~$t$, let $K\coloneqq\floor{\log_2t}+2$. Write $\widetilde L\in\R^K$ for the vector of cumulative corrected losses of learners $0,\ldots,K-2$ with a final zero for all learners that have not yet started. We use the learning rates $\eta_k\asymp\eps_k/\prn{GW}$ specified in~\cref{app:master}. Starting from $V_{K-1}(\widetilde L)\coloneqq0$, we define
\begin{equation}\label{eq:mix-potential-overview}
V_k(\widetilde L)\coloneqq-\frac{1}{\eta_k}\log\frac{\mathrm e^{-\eta_k\widetilde L_k}+\mathrm e^{-\eta_kV_{k+1}(\widetilde L)}}{2}\qquad(k=K-2,\ldots,0).
\end{equation}
The recursion aggregates the losses up a binary tree. Its root value after round~$t$ is denoted by~$V_0^{(t)}$.
At each split, we have $V_k(\widetilde L)-\min\set{\widetilde L_k,V_{k+1}(\widetilde L)}\lesssim\eta_k^{-1}$.
Summing along the path to learner~$i$ gives $V_0^{(T)}-\widetilde L_{T,i}\lesssim\sum_{k=0}^i\eta_k^{-1}\lesssim GW/\eps_i$, where the last inequality uses the geometric growth of~$\eps_k^{-1}$.
Since \decompref{main}{A} is the negative cumulative corrected loss, we have
\begin{equation}\label{eq:mix-comparison-overview}
\decompref{main}{A}=-\widetilde L_{T,i}=\prn{V_0^{(T)}-\widetilde L_{T,i}}-V_0^{(T)}\le O\prn{GW/\eps_i}-V_0^{(T)}.
\end{equation}

\paragraph{Cancellation of the Mixing Error.}\leavevmode
To evaluate $-V_0^{(T)}$ in \decompref{main}{A}, we sum the increments of the root potential. Let~$p_{t,j}$ ($0\le j<K$) be the leaf weights for~\eqref{eq:mix-potential-overview}, computed from losses through round~$t-1$ (see \cref{app:recursive-mixture}). The prediction for round $t$ is $w_t=\sum_{j=0}^{K-2}\prn{p_{t,j}/\prn{1-p_{t,K-1}}}w_t^{(j)}$. With $w_t^{(K-1)}\coloneqq w_t$ and~$r_{t,K-1}\coloneqq0$ for the last coordinate, the mixture satisfies $w_t=\sum_{j=0}^{K-1}p_{t,j}w_t^{(j)}$. The \emph{mixing error} is the mean corrected loss minus the potential increment,
\begin{equation}\label{eq:mix-gap-overview}
E_{t,0}\coloneqq-\sum_jp_{t,j}\Lambda_{\eps_j/2}(r_{t,j})-\prn{V_0^{(t)}-V_0^{(t-1)}}.
\end{equation}
When a new learner starts, we split the last leaf into two leaves of value zero, preserving the root potential. Since $V_0^{(0)}=0$, the increments telescope, and the definition of $E_{t,0}$ gives
\begin{equation}\label{eq:mix-telescope-overview}
-V_0^{(T)}=-\sum_{t=1}^T\prn{V_0^{(t)}-V_0^{(t-1)}}=\sum_{t=1}^T\prn{E_{t,0}+\textstyle\sum_jp_{t,j}\Lambda_{\eps_j/2}(r_{t,j})}.
\end{equation}
From $\sum_jp_{t,j}r_{t,j}=\inpr{g_t,w_t-\sum_jp_{t,j}w_t^{(j)}}=0$, we have $\sum_jp_{t,j}\Lambda_{\eps_j/2}(r_{t,j})=-\sum_jp_{t,j}(\eps_j/2)\abs{r_{t,j}}$. As proved in \cref{app:mixing-exact-weights}, the mixing error satisfies $E_{t,0}\le\frac13\sum_jp_{t,j}(\eps_j/2)\abs{r_{t,j}}$. The negative absolute-value terms thus absorb the mixing error in \decompref{main}{A} on each round as follows:
\begin{equation}\label{eq:mix-target}
\decompref{main}{A}=\underbrace{\prn{V_0^{(T)}-\widetilde L_{T,i}}}_{\lesssim GW/\eps_i}+\sum_{t=1}^T\underbrace{\prn{E_{t,0}-\textstyle\sum_jp_{t,j}\prn{\eps_j/2}\abs{r_{t,j}}}}_{\le-\frac23\sum_jp_{t,j}(\eps_j/2)\abs{r_{t,j}}\le0}\lesssim\frac{GW}{\eps_i}.
\end{equation}

\paragraph{Regret over All Rounds.}\leavevmode
Combining the bounds on \decompref{main}{A} and \decompref{main}{B} gives $\sum_{t=2^i}^T\Lambda_{2\eps_i}(r_t(u))\lesssim GW/\eps_i$. Before round~$2^i$, each term satisfies $\Lambda_{2\eps_i}(r_t(u))\le r_t(u)\le G\operatorname{diam}_2(\mathcal K)$. The sum over these rounds is at most~$\prn{2^i-1}G\operatorname{diam}_2(\mathcal K)\lesssim GW/\eps_i$, since $2^i=1/\prn{8\eps_i}$ and $\operatorname{diam}_2(\mathcal K)\lesssim W$ by \cref{item:width-diameter} in \cref{app:width-properties}.
Equation~\eqref{eq:adaptive-signed-overview} now yields $R_T^{\mathrm{lin}}(u)\lesssim GW/\eps_i+\eps_iN_T(u)$ simultaneously for every $i$ with $2^i\le T$. 
When $N_T(u)\lesssim GW$, learner $0$ suffices. For larger $N_T(u)$, the bound~$N_T(u)\lesssim GWT$ ensures that some such $i$ satisfies $\eps_i\asymp\sqrt{GW/N_T(u)}$. Hence
\begin{equation}\label{eq:adaptive-grid-overview}
R_T^{\mathrm{lin}}(u)\lesssim\min_{i\ge0,\,2^i\le T}\set*{\frac{GW}{\eps_i}+\eps_iN_T(u)}\lesssim GW+\sqrt{GW N_T(u)}.
\end{equation}
For subgradient feedback, convexity gives $R_T(u)\le R_T^{\mathrm{lin}}(u)$, and~\eqref{eq:negative-gap} gives $N_T(u)\le\Delta_T(u)$.

\paragraph{Relation to Prior Mixing Methods.}\leavevmode
A standard mixture of expert predictions with initial weights~$\pi_i$ and learning rates~$\eta_i$, such as ML-Prod \citep{gaillard2014second}, can have a term~$\log\prn{1/\pi_i}/\eta_i$ in its regret bound; assigning equal weights to $\Theta(\log T)$ learners introduces a typical $\log\log T$ dependence. Our recursive mixture instead gives the~$O\prn{GW/\eps_i}$ bound in~\eqref{eq:mix-target}, avoiding the dependence on~$T$; see \cref{app:multiple-rates} for a detailed discussion. 
Mixtures over learning rates, e.g., MetaGrad \citep{erven2021metagrad}, also differ in what they adapt to. Ours adapts to~$N_T(u)$, while MetaGrad adapts to~$\sum_{t=1}^T r_t(u)^2$. Its second-order bound still depends logarithmically on~$T$.
Recursive binary mixing is also used in a different context to control switching regret \citep{pasteris2024switching,yang2026logarithmic}.

\section{Applications}\label{sec:applications}
We study two applications of \cref{thm:adaptive-known-width}: contextual search and online inverse linear optimization.
In both applications the learner receives feedback about an unknown linear function without observing its values.
The losses in these applications need not satisfy a self-bounding relation between gradient norms and loss values, so small-loss bounds requiring that relation do not directly apply.

\subsection{Contextual Search with Comparison Noise}\label{sec:search}
Consider the contextual search setting. 
Fix an unknown~$u\in\B_2^d$. In round~$t$, the learner sees a context~$\phi_t\in\B_2^d$,
predicts~$\widehat y_t\in[-1,1]$ for the response~$y_t^\star\coloneqq\inpr{\phi_t,u}$,
and observes $Y_t\in\set{-1,1}$ conveying information about the comparison between~$\widehat y_t$ and~$y_t^\star$.
The context may depend on past observations.
At a tie, the environment chooses either sign for~$\sgn(0)$ before drawing the bit.
Let~$\mathcal F_t$ be the sigma-algebra generated by the history, current context and
prediction, and any tie-breaking sign. For some~$\kappa\in(0,1]$ unknown to the learner,
the probability of a correct comparison satisfies
\begin{equation}\label{eq:noise-model}
\Pr\prn*{Y_t=\sgn(\widehat y_t-y_t^\star)\mid\mathcal F_t}\ge(1+\kappa)/2.
\end{equation}

We run \cref{alg:adaptive} in \cref{app:master} on $\mathcal K=\B_2^d$ with~$G=1$ and $\width(\B_2^d)\le W\lesssim\sqrt d$. From its estimate~$w_t$,
we predict~$\widehat y_t\coloneqq\inpr{\phi_t,w_t}$. 
For the observed comparison bit~$Y_t$, define the loss~%
$f_t(w)\coloneqq Y_t\inpr{\phi_t,w}$ for~$w\in\B_2^d$ and feed its gradient
$g_t\coloneqq\nabla f_t(w_t)=Y_t\phi_t$ to the algorithm.
\begin{restatable}{corollary}{adaptivesearchbound}\label{cor:search-adaptive}
Under~\eqref{eq:noise-model}, \cref{alg:adaptive} in \cref{app:master}
with the parameters, queries, and inputs defined above satisfies
\begin{equation}\label{eq:adaptive-noise-expected}
\sup_{T\ge1}\E\brc*{\sum_{t=1}^T\abs{\widehat y_t-\inpr{\phi_t,u}}}\lesssim\sqrt d/\kappa^2.
\end{equation}
For independent flips with probability~$\prn{1-\kappa}/2$, the expected-error rate is minimax optimal.
\end{restatable}

{\renewcommand{\proofname}{Proof sketch}
\begin{proof}[\cref{cor:search-adaptive}.]
For these linear losses, we have $r_t(u)=Y_t\prn{\widehat y_t-y_t^\star}$.
The cumulative absolute prediction error is therefore
$R_T^{\mathrm{abs}}(u)=\sum_{t=1}^T\abs{\widehat y_t-y_t^\star}$.
Note that only flipped comparisons contribute to~$N_T(u)$.
By~\eqref{eq:noise-model}, we have
$\E\brc{r_t(u)\mid\mathcal F_t}\ge\kappa\abs{\widehat y_t-y_t^\star}$.
Taking expectations and summing over rounds, we apply the bound on the linearized regret
in terms of~$N_T(u)$ from \cref{thm:adaptive-known-width} to obtain
\begin{equation}\label{eq:adaptive-noise-mean}
\kappa\E\brc{R_T^{\mathrm{abs}}(u)}
\le \E\brc{R_T^{\mathrm{lin}}(u)}
\lesssim W+\sqrt{W\E\brc{R_T^{\mathrm{abs}}(u)}}.
\end{equation}
The last inequality uses Jensen's inequality and~$N_T(u)\le R_T^{\mathrm{abs}}(u)$.
Solving the inequality~\eqref{eq:adaptive-noise-mean} for $\E\brc{R_T^{\mathrm{abs}}(u)}$ gives
$\E\brc{R_T^{\mathrm{abs}}(u)}\lesssim W/\kappa^2$.
See \cref{app:noise-adaptive} for a high-probability upper bound and \cref{app:noise-lower}
for the matching lower bound in expectation.
\end{proof}}

The total arithmetic cost through round~$T$ is
$O\prn{Td\prn{d+T}\log\prn{T+2}}$, including the projections onto the ball. See \cref{app:adaptive-total-cost} for the operation counts.
Together with~\eqref{eq:adaptive-noise-expected}, this operation count gives, for fixed~$\kappa>0$, a polynomial-time algorithm in the real-arithmetic model with a $\poly\prn{d}$ bound on the expected cumulative absolute error, resolving the question of
\citet[Section~5.2]{liu2021optimal}.\looseness=-1

\subsection{Online Inverse Linear Optimization}\label{sec:oilo}
We then consider the following OILO setting. Fix the user's unknown utility vector~$u\in\mathcal K$. In round~$t$ the learner receives
a nonempty compact action set~$\mathcal X_t\subset\R^d$ of diameter at most~$G$.
Using its current utility estimate~$w_t\in\mathcal K$, the learner recommends
$\widehat a_t\in\argmax_{a\in\mathcal X_t}\inpr{w_t,a}$. It then observes
the user's feedback action~$y_t\in\mathcal X_t$ and feeds~$g_t\coloneqq\widehat a_t-y_t$ to the algorithm.
The cumulative utility the user forgoes is
$D_T(u)\coloneqq\sum_{t=1}^T\max_{a\in\mathcal X_t}\inpr{u,a-y_t}$.
Define the regret relative to the observed actions~$y_t$ by
\begin{equation}\label{eq:oilo-regrets}
R_T^{\mathrm{obs}}(u)\coloneqq\sum_{t=1}^T\inpr{u,y_t-\widehat a_t}.
\end{equation}

\begin{restatable}{corollary}{oilobound}\label{cor:oilo}
With~$W\ge\width(\mathcal K)$ and the feedback defined in \cref{sec:oilo},
\cref{alg:adaptive} in \cref{app:master} satisfies, for every~$T\ge1$,
\begin{equation}\label{eq:oilo-demo}
R_T^{\mathrm{obs}}(u)\lesssim GW+\sqrt{GW D_T(u)}.
\end{equation}
\end{restatable}

{\renewcommand{\proofname}{Proof sketch}
\begin{proof}[\cref{cor:oilo}.]
To apply \cref{thm:adaptive-known-width}, define the convex \emph{suboptimality loss}
$f_t(w)\coloneqq\max_{a\in\mathcal X_t}\inpr{w,a-y_t}$ for~$w\in\mathcal K$.
The choice of~$\widehat a_t$ gives~$g_t\in\partial f_t(w_t)$.
Since~$y_t\in\mathcal X_t$, each~$f_t$ is nonnegative, so subtracting the per-round minima and using \eqref{eq:negative-gap} give~$N_T(u)\le\Delta_T(u)\le\sum_{t=1}^Tf_t(u)=D_T(u)$.
Moreover, $\inpr{u,y_t-\widehat a_t}=r_t(u)-f_t(w_t)\le r_t(u)$ holds.
Thus we have~$R_T^{\mathrm{obs}}(u)\le R_T^{\mathrm{lin}}(u)$, and the bound in \cref{thm:adaptive-known-width} for $R_T^{\mathrm{lin}}(u)$ proves the claim; see \cref{app:oilo-proof} for the full proof.
\end{proof}}

The learner computes the recommendation with one linear optimization call on~$\mathcal X_t$,
and~$w_t$ can be chosen before~$\mathcal X_t$ is revealed.

To compare the regret relative to optimal actions, define~$R_T^{\mathrm{opt}}(u)\coloneqq D_T(u)+R_T^{\mathrm{obs}}(u)$.
The AM--GM inequality gives
$\sqrt{GW D_T(u)}\le\prn{GW+D_T(u)}/2$, so \cref{cor:oilo} implies
$R_T^{\mathrm{opt}}(u)\lesssim GW+D_T(u)$.
For~$\mathcal K=\B_2^d$, choosing~$W\asymp\sqrt d$ recovers the bound obtained from the corrupted-feedback guarantee of \citet[Section~4]{gupta2026optimal}; see also \cref{app:oilo-suboptimal}.
Our \cref{cor:oilo} gives a finer bound on~$R_T^{\mathrm{obs}}(u)$ with a~$\sqrt{D_T(u)}$ dependence
on the user's cumulative suboptimality.

\label{main-end}

\section*{AI Usage}
The author used GPT-6 Astra, GPT-6.1 Sol, and GPT-6.2 Astra when working on the proofs.
In writing the manuscript, the author used these three models as well as Claude Opus 5.5.

\section*{Acknowledgements}
Shinsaku Sakaue was supported by JST BOOST Program Japan Grant Number
\mbox{JPMJBY24D1}.

\clearpage
\bibliography{references}
\clearpage

\appendix
\crefalias{section}{appendix}
\crefalias{subsection}{appendix}
\crefalias{subsubsection}{appendix}
\section{Maximal Subexponential Mean Width and Initialization}\label{app:geometry}
We show that the maximal subexponential mean width of a domain is comparable to the infimum of its coordinate representation costs over finite admissible descriptions in \cref{app:representation}.
After the transformation rules and diameter bounds in~\cref{app:width-properties}, we prove the width bounds and construct an admissible description for each unit~$\ell_p$ ball in~\cref{app:explicit-initializations}.
We then extend these results to translated and scaled balls, ellipsoids, boxes, and simplices in~\cref{app:domain-extensions}.
Finally, we compare this width with Gaussian and exponential widths in \cref{app:width-comparisons}.
For~$z\in\R^d$, define the support function by $\sigma_{\mathcal K}(z)\coloneqq\sup_{u\in\mathcal K}\inpr{z,u}$.

\subsection{Coordinate Representation Cost and Width}\label{app:representation}
For an integer~$N\ge1$, an \emph{admissible description}~%
$\mathcal D=\prn{w_\circ,A,\ell,\underline h}$ consists of a point~$w_\circ\in\R^d$,
a matrix~$A\in\R^{d\times N}$ with $\mathcal K\subseteq w_\circ+A[-1,1]^N$,
and vectors~$\ell\in(0,\infty)^N$ and~$\underline h\in(0,1]^N$.
For every~$i\in\set{1,\ldots,N}$, the column-length bound satisfies~%
$\ell_i\ge\norm{Ae_i}_2$, and the threshold has the form~%
$\underline h_i=2^{-J_i}$ for an integer~$J_i\ge0$.
Put~$L_{\underline h}\coloneqq\sum_{i=1}^N\ell_i \underline h_i$. For a coefficient vector~$v\in[-1,1]^N$, define its \emph{coefficient cost} by
\begin{equation}\label{eq:geometry}
\Phi_{\ell,\underline h}( v )\coloneqq
\sum_{i=1}^N \ell_i\brc*{\underline h_i+\abs{v_i}
\prn*{1+\log_+\frac{\abs{v_i}}{\underline h_i}}}.
\end{equation}
Here $\log_+x\coloneqq\log\max\set{1,x}$ for~$x\ge0$, so the logarithmic term vanishes at~$v_i=0$; the constant term~$\ell_i \underline h_i$ remains. The \emph{coordinate representation cost} of~$\mathcal D$ on~$\mathcal K$ is the supremum, over~$u\in\mathcal K$, of the minimum coefficient cost among representations of that~$u$:
\begin{equation}\label{eq:representation-cost}
C_{\mathcal D}(\mathcal K)\coloneqq
\sup_{u\in \mathcal K}
  \min\Set*{\Phi_{\ell,\underline h}(v)}{v \in[-1,1]^N,\,A v =u-w_\circ}.
\end{equation}
The inner minimum is attained by compactness and continuity.
The minimizing coefficients of a comparator are used only in the analysis.

We will use the above representation to connect the regret bound achieved by our algorithm and the maximal subexponential mean width.
In \cref{app:general-core}, the coordinate initialization uses the admissible description~$\mathcal D$ chosen by \cref{alg:oco} in \cref{app:projection}.
As we will see in \cref{app:core-proof}, initializing the regularizer using~$\mathcal D$ gives a domain-dependent term
of~$O\prn{GC_{\mathcal D}(\mathcal K)/\eps}$ in the regret bound. 
Recall from~\eqref{eq:width} that the width is given by
\begin{equation}\label{eq:width-recall}
\width(\mathcal K)=\sup_Z\Set*{\E\sigma_{\mathcal K}(Z)}{\E Z=0,\ \E\exp\prn{\abs{\inpr{v,Z}}}\le2\text{ for every }v\in\B_2^d},
\end{equation}
where~$Z$ ranges over~$\R^d$-valued random vectors.
The next proposition compares this width with the best coordinate representation cost.

\begin{proposition}\label{prop:width-equivalence}
For the domain~$\mathcal K$, we have
\begin{equation}\label{eq:width-equivalence}
\width(\mathcal K)\le\inf_{\mathcal D}C_{\mathcal D}(\mathcal K)\le20\,\width(\mathcal K),
\end{equation}
where the infimum is over finite admissible descriptions. An admissible description establishing the upper bound can be chosen
so that the number~$N$ of columns of~$A$ satisfies~$N\le5^r$, where~$r\coloneqq\dim\operatorname{aff}(\mathcal K)$, the dimension of the affine hull of $\mathcal K$.
\end{proposition}
\begin{proof}[\cref{prop:width-equivalence}]
We first bound the width by the cost of any admissible description. Let $\varphi(s)\coloneqq\exp\prn{\abs{s}}-1$. Direct maximization gives its convex conjugate
\begin{equation}\label{eq:exp-conjugate}
\varphi^*\prn{z}=
\begin{cases}
0,&\abs{z}\le1,\\
\abs{z}\log\abs{z}-\abs{z}+1,&\abs{z}>1.
\end{cases}
\end{equation}

Fix an admissible description~$\mathcal D$ and~$z\in\R^d$.
For any~$u\in\mathcal K$ and any~$v\in[-1,1]^N$ satisfying~$Av=u-w_\circ$,
applying the Fenchel--Young inequality to~%
$\inpr{A e_i/\ell_i,z}$ and~$v_i/\underline h_i$ with weight~$\ell_i \underline h_i$, then summing over~$i$, gives
\begin{equation}\label{eq:width-from-representation}
\begin{aligned}
\inpr{z,u-w_\circ}
&\le\sum_{i=1}^N \ell_i \underline h_i\varphi(\inpr{A e_i/\ell_i,z})
 +\sum_{i=1}^N \ell_i \underline h_i\varphi^*\prn{v_i/\underline h_i}\\
&\le\sum_{i=1}^N \ell_i \underline h_i\varphi(\inpr{A e_i/\ell_i,z})
 +\Phi_{\ell,\underline h}( v )-L_{\underline h}.
\end{aligned}
\end{equation}
The second inequality uses $\varphi^*\prn{t}\le\abs{t}\prn{1+\log_+\abs{t}}$ for every~$t\in\R$, which follows from~\eqref{eq:exp-conjugate}.

For each fixed~$u\in\mathcal K$, minimizing~\eqref{eq:width-from-representation}
over its representing coefficients~$v$ leaves the left-hand side unchanged and gives
\begin{equation}\label{eq:width-fixed-point}
\inpr{z,u-w_\circ}
\le\sum_{i=1}^N\ell_i\underline h_i\varphi\prn{\inpr{Ae_i/\ell_i,z}}
 +\min\Set*{\Phi_{\ell,\underline h}(v)}{v\in[-1,1]^N,\,Av=u-w_\circ}
 -L_{\underline h}.
\end{equation}
Taking the supremum of both sides of~\eqref{eq:width-fixed-point} over~%
$u\in\mathcal K$ and using~\eqref{eq:representation-cost} gives
\begin{equation}\label{eq:width-support-from-cost}
\sigma_{\mathcal K}(z)-\inpr{z,w_\circ}
\le\sum_{i=1}^N\ell_i\underline h_i\varphi\prn{\inpr{Ae_i/\ell_i,z}}
 +C_{\mathcal D}(\mathcal K)-L_{\underline h}.
\end{equation}
For every~$Z$ satisfying the constraints in~\eqref{eq:width-recall}, the bound~%
$\norm{Ae_i/\ell_i}_2\le1$ gives
$\E\varphi\prn{\inpr{Ae_i/\ell_i,Z}}\le1$ for each~$i\in\set{1,\ldots,N}$.
Since $\E\inpr{Z,w_\circ}=\inpr{\E Z,w_\circ}=0$, taking expectations in~%
\eqref{eq:width-support-from-cost} yields
\begin{equation}
\E\sigma_{\mathcal K}(Z)
\le\sum_{i=1}^N\ell_i\underline h_i+C_{\mathcal D}(\mathcal K)-L_{\underline h}
=L_{\underline h}+C_{\mathcal D}(\mathcal K)-L_{\underline h}
=C_{\mathcal D}(\mathcal K).
\end{equation}
Taking the supremum over~$Z$ satisfying the constraints in~\eqref{eq:width-recall} and then the infimum over~%
$\mathcal D$ proves the lower bound in \cref{prop:width-equivalence}.

For the upper bound, first suppose~$\mathcal K=-\mathcal K$ and work in its~$r$-dimensional span. A~$1/2$-net of the unit sphere supplies unit vectors~%
$q_1,\ldots,q_m$, $m\le5^r$, such that
$\prn{1/2}\B_2^r\subseteq\conv\set{\pm q_i:1\le i\le m}$;
see \citet[Section 4.2]{vershynin2026hdp} for the standard net bound.
For~$b\in(0,\infty)^m$, define
\begin{equation}\label{eq:moment-value}
\operatorname{val}(b)\coloneqq\sup_\mu\set*{\E_\mu \sigma_{\mathcal K}(Z):
\E_\mu\varphi(\inpr{q_i,Z})\le b_i\ \text{for all }i\in\set{1,\ldots,m}}.
\end{equation}

Here the supremum is over probability laws~$\mu$ on the span of~$\mathcal K$.
We will bound~$\operatorname{val}$ by an affine function of the moment bounds~$b_i$ and use this bound to construct an admissible description with cost~$O\prn{\width(\mathcal K)}$.
For every feasible law, the bounds
$\norm{z}_2\le2\max_{1\le i\le m}\abs{\inpr{q_i,z}}$ and~$\abs{s}\le\varphi(s)$ give
$\E\norm{Z}_2\le2\sum_{i=1}^m b_i<\infty$.
Boundedness of~$\mathcal K$ therefore makes~$\operatorname{val}$ finite.
Mixing feasible laws shows that~$\operatorname{val}$ is concave, and enlarging
the constraint bounds~$b_i$ shows that it is coordinatewise nondecreasing.

Given a feasible random vector~$Z$, define~$\widetilde Z$ conditionally on~$Z$
by choosing~$Z$ and~$-Z$ with probability~$1/2$ each.
Then we have~$\E\brc{\widetilde Z\mid Z}=0$ and hence~$\E\widetilde Z=0$.
Since~$\mathcal K=-\mathcal K$ and~$\varphi$ is even, for every~%
$i\in\set{1,\ldots,m}$ we have
\begin{equation}\label{eq:moment-symmetrization}
\E\sigma_{\mathcal K}(\widetilde Z)=\E\sigma_{\mathcal K}(Z)
\quad\text{and}\quad
\E\varphi\prn{\inpr{q_i,\widetilde Z}}=\E\varphi\prn{\inpr{q_i,Z}}.
\end{equation}
By~\eqref{eq:moment-symmetrization}, the supremum in~\eqref{eq:moment-value} is unchanged if restricted to
symmetric laws, which are centered.

At the constraint vector $\mathbf1\coloneqq(1,\ldots,1)\in\R^m$,
the finite concave function~$\operatorname{val}$ has a supergradient~%
$\beta\in\R^m$ \citep[Section 23]{rockafellar1970convex}.
By definition, $\beta$ satisfies
\begin{equation}\label{eq:moment-supporting}
\operatorname{val}(b)\le\operatorname{val}(\mathbf1)
 +\sum_{i=1}^m\beta_i(b_i-1)
\quad\text{for every }b\in(0,\infty)^m.
\end{equation}
Coordinatewise monotonicity implies~$\beta_i\ge0$ for every~$i\in\set{1,\ldots,m}$.

For a point~$z$ in the span of~$\mathcal K$ and an integer~$k\ge1$, set
$b_i^{(k)}\coloneqq\varphi\prn{\inpr{q_i,z}}+1/k$ for every~$i\in\set{1,\ldots,m}$.
The law concentrated at~$z$ is feasible in~\eqref{eq:moment-value} for~%
$b^{(k)}\in(0,\infty)^m$.
Combining this feasible choice with~\eqref{eq:moment-supporting} gives
\begin{equation}\label{eq:moment-point-mass}
\sigma_{\mathcal K}(z)
\le\operatorname{val}(b^{(k)})
\le\operatorname{val}(\mathbf1)
 +\sum_{i=1}^m\beta_i\brc*{\varphi\prn{\inpr{q_i,z}}+\frac1k-1}.
\end{equation}
Letting~$k\to\infty$ in~\eqref{eq:moment-point-mass} gives
\begin{equation}\label{eq:moment-majorant}
\sigma_{\mathcal K}(z)\le c_0 +\sum_{i=1}^m \beta_i\varphi(\inpr{q_i,z}),\quad\text{where}\;\; c_0\coloneqq \operatorname{val}(\mathbf1)-\sum_{i=1}^m \beta_i.
\end{equation}
At~$z=0$, \eqref{eq:moment-majorant} gives~$0\le c_0$.

For any symmetric law feasible at~$b=\mathbf1$ and any~$v$ in the span of~$\mathcal K$ with~$\norm{v}_2\le1$,
the vector~$v/2$ is a convex combination of~$\pm q_i$.
Convexity of~$\varphi$ gives~$\E\varphi(\inpr{v,Z/2})\le1$.
For a direction in~$\R^d$, its orthogonal projection onto the span of~$\mathcal K$ preserves the inner product with~$Z$ and cannot increase its norm.
Thus~$Z/2$ satisfies the constraints in~\eqref{eq:width-recall}. By homogeneity of the support function, we have $\E\sigma_{\mathcal K}(Z)=2\E\sigma_{\mathcal K}(Z/2)\le2\width(\mathcal K)$. Taking the supremum over the symmetric laws in~\eqref{eq:moment-value} gives
\begin{equation}\label{eq:moment-cost}
 c_0 +\sum_{i=1}^m \beta_i=\operatorname{val}(\mathbf1)\le2\width(\mathcal K).
\end{equation}

We now convert the upper bound on~$\sigma_{\mathcal K}$ in~\eqref{eq:moment-majorant} into an admissible description with cost~$O\prn{\width(\mathcal K)}$.
Let $\mathcal J_+\coloneqq\Set{i\in\set{1,\ldots,m}}{\beta_i>0}$ and retain only these coordinates.
For any nonzero~$u\in\mathcal K$, we have~$\sigma_{\mathcal K}(tu)\ge t\norm{u}_2^2$ for~$t>0$, so~\eqref{eq:moment-majorant} requires~$\mathcal J_+\ne\varnothing$.
Set $N\coloneqq\abs{\mathcal J_+}\le m\le5^r$ and identify~$\R^{\mathcal J_+}$ with~$\R^N$ when forming an admissible description. Put
$\varphi_\beta(z)\coloneqq\sum_{i\in\mathcal J_+} \beta_i\varphi(\inpr{q_i,z})$.
For every~$u\in\mathcal K$ and~$z\in\operatorname{span}(\mathcal K)$, the definition of the support function and~\eqref{eq:moment-majorant} give
$\inpr{u,z}\le\sigma_{\mathcal K}(z)\le c_0+\varphi_\beta(z)$.
Subtracting~$\varphi_\beta(z)$ and taking the supremum over~$z$ gives
\begin{equation}\label{eq:conjugate-bound}
\varphi_\beta^*\prn{u}\coloneqq\sup_{z\in\operatorname{span}(\mathcal K)}\brc{\inpr{u,z}-\varphi_\beta(z)}\le c_0.
\end{equation}
Here~$\varphi_\beta^*$ is the convex conjugate of~$\varphi_\beta$ on~$\operatorname{span}(\mathcal K)$.
Equivalently, there are
coefficients~$y_i\in\R$ for every~$i\in\mathcal J_+$ with
\begin{equation}\label{eq:dual-coefficients}
u=\sum_{i\in\mathcal J_+} y_iq_i\quad\text{and}\quad \sum_{i\in\mathcal J_+} \beta_i\varphi^*\prn{y_i/ \beta_i}\le c_0 .
\end{equation}
To verify~\eqref{eq:dual-coefficients}, we use the conjugacy between sums and infimal convolutions \citep[Section~16]{rockafellar1970convex}. For fixed~$u$, minimize $\sum_{i\in\mathcal J_+}\beta_i\varphi^*\prn{y_i/\beta_i}$ subject
to~$\sum_{i\in\mathcal J_+}y_iq_i=u$. The summands are continuous and
superlinear, so the objective has compact sublevel sets. Hence the minimum is
attained whenever feasible, and its value as a function of~$u$ has a closed
epigraph. Taking its conjugate separates the~$y_i$ and gives~$\varphi_\beta$.
Biconjugacy identifies the minimum with~$\varphi_\beta^*\prn{u}$, which is at most~$c_0$ by~\eqref{eq:conjugate-bound}.

Set~$w_\circ=0$. For~$i\in\mathcal J_+$, let~$\ell_i\coloneqq\mathrm{e}^2(c_0+\beta_i)$, set~$Ae_i\coloneqq\ell_iq_i$, and take~%
$v_i\coloneqq y_i/\ell_i$.
Then we have~$Av=u$ by~\eqref{eq:dual-coefficients}.
If~$\abs{y_i}/\beta_i<\mathrm{e}^2$, we have~$\abs{y_i}<\mathrm{e}^2\beta_i\le\ell_i$.
Otherwise, we have $\varphi^*\prn{y_i/\beta_i}\ge\abs{y_i}/\beta_i$, so~\eqref{eq:dual-coefficients} gives~$\abs{y_i}\le c_0\le\ell_i$.
Thus~$\abs{v_i}\le1$ holds for every~$i\in\mathcal J_+$.
Choose a dyadic~$\underline h_i$ with $\beta_i/\prn{2\ell_i}<\underline h_i\le \beta_i/\ell_i$.
For~$t_i\coloneqq\abs{y_i}/ \beta_i$, the coefficient cost satisfies
\begin{equation}\label{eq:dual-initialization-cost}
\Phi_{\ell,\underline h}( v )
\le\sum_{i\in\mathcal J_+} \beta_i+\sum_{i\in\mathcal J_+} \beta_i t_i(1+\log2+\log_+t_i)
\le4 c_0 +5\sum_{i\in\mathcal J_+} \beta_i.
\end{equation}
The last step in~\eqref{eq:dual-initialization-cost} uses $t(1+\log2+\log_+t)\le4(1+\varphi^*\prn{t})$ for all~$t\ge0$.
For~$0\le t\le1$, this reduces to~$t(1+\log2)\le4$ by~\eqref{eq:exp-conjugate}.
To prove the scalar inequality for~$t\ge1$, it suffices to show that $4(1+\varphi^*\prn{t})-t(1+\log2+\log t)\ge0$.
By~\eqref{eq:exp-conjugate}, this difference equals~$8+3t\log t-\prn{5+\log2}t$ and has minimum~$8-3\exp\prn{\prn{2+\log2}/3}>0$ on~$[1,\infty)$.
Summing the scalar inequality at~$t=t_i$ with weights~$\beta_i$ and using $\sum_{i\in\mathcal J_+}\beta_i\varphi^*\prn{t_i}\le c_0$ from~\eqref{eq:dual-coefficients} gives~\eqref{eq:dual-initialization-cost}.
For this fixed description~$\mathcal D$, every~$u\in\mathcal K$ has coefficients feasible in~\eqref{eq:representation-cost} and satisfying~\eqref{eq:dual-initialization-cost}.
Taking the minimum for each~$u$ and then the supremum over~$u\in\mathcal K$, and using~$c_0\ge0$ and~\eqref{eq:moment-cost}, gives
\begin{equation}\label{eq:symmetric-representation-cost}
C_{\mathcal D}(\mathcal K)
\le4c_0+5\sum_{i\in\mathcal J_+}\beta_i
\le5\prn*{c_0+\sum_{i=1}^m\beta_i}
=5\operatorname{val}(\mathbf1)
\le10\width(\mathcal K).
\end{equation}

For general~$\mathcal K$, choose~$w_\circ\in\mathcal K$ and apply the symmetric construction to~$\mathcal K-\mathcal K$ in its span.
If~$Z$ satisfies the constraints in~\eqref{eq:width-recall}, then so does~$-Z$, giving
\begin{equation}\label{eq:width-difference-body}
\E\sigma_{\mathcal K-\mathcal K}(Z)=\E\sigma_{\mathcal K}(Z)+\E\sigma_{\mathcal K}(-Z)\le2\width(\mathcal K).
\end{equation}
Taking the supremum over~$Z$ in~\eqref{eq:width-difference-body} gives $\width(\mathcal K-\mathcal K)\le2\width(\mathcal K)$.
Since~$\mathcal K-w_\circ\subseteq\mathcal K-\mathcal K$, the same matrix~$A$, length bounds~$\ell$, and thresholds~$\underline h$, with translation~$w_\circ$, give a description of~$\mathcal K$ with cost at most~$20\width(\mathcal K)$.
This completes the proof.
\end{proof}
For canonical domains including balls, simplices, ellipsoids, and boxes, 
we can construct explicit admissible descriptions using at most~$d$ coordinates; see~\cref{app:explicit-initializations,app:domain-extensions}.

\subsection{Basic Properties of Width}\label{app:width-properties}
We show how translation, isometric embedding, and uniform scaling affect~$\width(\mathcal K)$, and compare it with the Euclidean diameter and radius. The transformation rule then extends the unit-ball results to other domains in \cref{app:domain-extensions}. 
The bound~$\operatorname{diam}_2(\mathcal K)\lesssim\width(\mathcal K)$ proved below is used in \cref{sec:adaptation,app:adapt}.

\Needspace{6\baselineskip}
\begin{lemma}\label{lem:width-properties}
The width of~$\mathcal K$ satisfies the following properties.
\begin{enumerate}[label=(\roman*),ref=(\roman*),leftmargin=*,nosep]
\item\label[widthproperty]{item:width-transform} For any integer~$m\ge d$, matrix~$U\in\R^{m\times d}$ with~$U^\top U=I_d$, point~$b\in\R^m$, and~$R>0$, we have
\begin{equation}\label{eq:width-transform}
\width(b+RU\mathcal K)=R\width(\mathcal K).
\end{equation}
\item\label[widthproperty]{item:width-diameter} Writing $\operatorname{diam}_2\mathcal K\coloneqq\max_{u,w\in\mathcal K}\norm{u-w}_2$, we have
\begin{equation}\label{eq:width-radius}
\frac{\log2}{2}\operatorname{diam}_2\mathcal K
\le\width(\mathcal K)\le\sqrt{2d}\inf_{w_\circ\in\R^d}\sup_{u\in\mathcal K}\norm{u-w_\circ}_2.
\end{equation}
\end{enumerate}
\end{lemma}
\begin{proof}
For \cref{item:width-transform}, let~$Z\in\R^m$ satisfy the constraints in~\eqref{eq:width-recall} in dimension~$m$.
Then~$U^\top Z$ satisfies them in dimension~$d$, since~$\E U^\top Z=0$ and~$\norm{Uv}_2=\norm{v}_2$ for every~$v\in\R^d$.
Centering gives $\E\sigma_{b+RU\mathcal K}(Z)=R\E\sigma_{\mathcal K}(U^\top Z)$.
Conversely, if~$Y\in\R^d$ satisfies the constraints in dimension~$d$, then~$UY$ satisfies them in dimension~$m$, since~$\E UY=0$ and~$\norm{U^\top v}_2\le\norm{v}_2$ for every~$v\in\R^m$.
The identity~$U^\top UY=Y$ and the two choices of random vectors give~\eqref{eq:width-transform}.

For the lower bound in \cref{item:width-diameter}, choose diameter endpoints~$u_+,u_-\in\mathcal K$ and set $v\coloneqq(u_+-u_-)/\norm{u_+-u_-}_2$.
The centered vector~$Z=\pm(\log2)v$, with equal probabilities, satisfies the constraints in~\eqref{eq:width-recall}, since~$\abs{\inpr{x,Z}}\le\log2$ for every~$x\in\B_2^d$.
It gives
\begin{equation}
\width(\mathcal K)\ge\E\sigma_{\mathcal K}(Z)
=\frac{\log2}{2}\brc{\sigma_{\mathcal K}(v)+\sigma_{\mathcal K}(-v)}
\ge\frac{\log2}{2}\operatorname{diam}_2\mathcal K.
\end{equation}

For the upper bound, let~$Z\in\R^d$ satisfy the constraints in~\eqref{eq:width-recall}.
For every~$v\in\B_2^d$, the inequality $s^2/2\le\exp\prn{\abs{s}}-1$ gives
$\E\inpr{v,Z}^2\le2\E\brc{\exp\prn{\abs{\inpr{v,Z}}}-1}\le2$.
The resulting second-moment bound and the Cauchy--Schwarz inequality imply
\begin{equation}\label{eq:width-second-moment}
\E ZZ^\top\preceq2I_d
\quad\text{and}\quad
\E\norm{Z}_2\le\sqrt{\E\norm{Z}_2^2}
=\sqrt{\tr(\E ZZ^\top)}\le\sqrt{2d}.
\end{equation}
For every~$w_\circ\in\R^d$, centering and the Cauchy--Schwarz inequality give
\begin{equation}\label{eq:width-radius-expectation}
\E\sigma_{\mathcal K}(Z)
=\E\brc*{\sup_{u\in\mathcal K}\inpr{Z,u-w_\circ}}
\le\E\norm{Z}_2\sup_{u\in\mathcal K}\norm{u-w_\circ}_2.
\end{equation}
Substituting~\eqref{eq:width-second-moment} into~\eqref{eq:width-radius-expectation} and taking the supremum over~$Z$ yields, for every~$w_\circ\in\R^d$,
$\width(\mathcal K)\le\sqrt{2d}\sup_{u\in\mathcal K}\norm{u-w_\circ}_2$.
Taking the infimum over~$w_\circ$ proves~\eqref{eq:width-radius}.
\end{proof}

\subsection{\texorpdfstring{Width and Initialization on Unit $\ell_p$ Balls}{Width and Initialization on Unit lp Balls}}\label{app:explicit-initializations}
We prove matching upper and lower bounds for the maximal subexponential mean width of~$\B_p^r$, then construct an admissible description~$\mathcal D$ with~$N=r$ and~$C_{\mathcal D}(\B_p^r)\lesssim\width(\B_p^r)$.
Fix an integer~$r\ge1$, let~$1\le p\le\infty$ and~$1/p+1/q=1$, and recall that~$\B_p^r=\Set{v\in\R^r}{\norm{v}_p\le1}$.
\begin{proposition}\label{prop:unit-lp-width}
For every~$r\ge1$ and~$1\le p\le\infty$, we have
\begin{equation}\label{eq:unit-lp-width}
\width(\B_p^r)\asymp r^{1/q}\min\set{q,\log\prn{r+1}}.
\end{equation}
All constants are universal, uniformly over~$p$ and~$r$.
\end{proposition}
Recall from~\eqref{eq:width-recall} that~$\width(\B_p^r)$ is the supremum of~$\E\norm{Z}_q$ over random vectors~$Z\in\R^r$ satisfying
\begin{equation}\label{eq:width-vector-constraints}
\E Z=0\quad\text{and}\quad\E\exp\prn{\abs{\inpr{v,Z}}}\le2\quad\text{for every }v\in\B_2^r.
\end{equation}
In~\cref{app:width-upper}, we prove the upper bound for every~$Z$ satisfying~\eqref{eq:width-vector-constraints}. In~\cref{app:width-lower}, we prove the lower bound by constructing such a~$Z$.

\subsubsection{Upper Bounds}\label{app:width-upper}
We bound~$\E\norm{Z}_q$ for every random vector~$Z\in\R^r$ satisfying~\eqref{eq:width-vector-constraints}.
For~$2\le p\le\infty$, we have~$1\le q\le2$.
Jensen's inequality and the coordinate bound~$\E Z_i^2\le2$ from~\eqref{eq:width-second-moment} give
\begin{equation}\label{eq:unit-lp-width-upper}
\E\norm{Z}_q
\le\prn*{\sum_{i=1}^r\E\abs{Z_i}^q}^{1/q}
\le\sqrt2 r^{1/q}.
\end{equation}
Since~$\log2\le\min\set{q,\log\prn{r+1}}\le2$, taking the supremum over~$Z$ proves the upper bound in~\eqref{eq:unit-lp-width} for~$p\ge2$.

For~$1\le p<2$, we have~$q>2$ and~$\sigma_{\B_p^r}(Z)=\norm{Z}_q$.
For finite~$q$, the coordinate moment bound~$\prn{\E\abs{Z_i}^q}^{1/q}\lesssim q$ \citep[Proposition 2.8.1]{vershynin2026hdp} gives
\begin{equation}\label{eq:lp-width-upper-moments}
\E\norm{Z}_q\le\prn*{\sum_{i=1}^r\E\abs{Z_i}^q}^{1/q}\lesssim qr^{1/q}.
\end{equation}
For every~$q>2$, including~$q=\infty$, Jensen's inequality for~$\log$ also gives
\begin{equation}\label{eq:lp-width-upper-maximum}
\E\norm{Z}_q
\le r^{1/q}\E\max_{1\le i\le r}\abs{Z_i}
\le r^{1/q}\log\sum_{i=1}^r\E\exp\prn{\abs{Z_i}}
\le r^{1/q}\log\prn{2r}.
\end{equation}
Taking the smaller upper bound in~\eqref{eq:lp-width-upper-moments} and~\eqref{eq:lp-width-upper-maximum} for finite~$q$, and using~\eqref{eq:lp-width-upper-maximum} for~$q=\infty$, proves
$\width(\B_p^r)\lesssim r^{1/q}\min\set{q,\log\prn{r+1}}$.

\subsubsection{Lower Bounds}\label{app:width-lower}
We prove the lower bound in~\eqref{eq:unit-lp-width} using a random vector with independent Laplace coordinates.
For each~$i\in\set{1,\ldots,r}$, let~$\xi_i$ be an independent symmetric Laplace variable with density~$\exp\prn{-\abs{x}}/2$, and put~$\xi\coloneqq(\xi_1,\ldots,\xi_r)$.
We first verify that~$Z\coloneqq\xi/4$ satisfies~\eqref{eq:width-vector-constraints}.
It is centered.
For~$v\in\B_2^r$, set~$X\coloneqq\inpr{v,\xi}$.
Independence gives~$\E X^2=2\norm{v}_2^2\le2$ and
\begin{equation}\label{eq:laplace-admissible}
\E\exp\prn{X/4}=\prod_{i=1}^r(1-v_i^2/16)^{-1}\le\exp\prn{1/15}.
\end{equation}
The last inequality uses~$-\log\prn{1-u}\le16u/15$ for~$0\le u\le1/16$.
Since~$X$ is symmetric, the inequality~$\exp\prn{\abs{x}}\le\exp\prn{x}+\exp\prn{-x}-1+\abs{x}$ and the Cauchy--Schwarz inequality yield
\begin{equation}\label{eq:laplace-absolute-moment}
\E\exp\prn{\abs{X}/4}
\le2\E\exp\prn{X/4}-1+\E\abs{X}/4
\le2\exp\prn{1/15}-1+\sqrt2/4<2.
\end{equation}
This proves both conditions in~\eqref{eq:width-vector-constraints}.
Consequently, we have
\begin{equation}\label{eq:unit-laplace-width-lower}
\width(\B_p^r)\ge\frac14\E\norm{\xi}_q.
\end{equation}
For~$2\le p\le\infty$, Jensen's inequality and~$\E\abs{\xi_i}=1$ give
\begin{equation}\label{eq:unit-lp-width-lower}
\E\norm{\xi}_q\ge\norm{\E\brc{\prn{\abs{\xi_i}}_{i=1}^r}}_q=r^{1/q}.
\end{equation}
Together with~\eqref{eq:unit-lp-width-upper}, this proves~\eqref{eq:unit-lp-width} for~$p\ge2$.

For~$1\le p<2$, the remaining assertion is~$\E\norm{\xi}_q\gtrsim r^{1/q}\min\set{q,\log\prn{r+1}}$.
First suppose~$r\ge2$ and~$2<q\le(\log r)/2$.
Let~$\nu_q\coloneqq\abs{\Set{i\in\set{1,\ldots,r}}{\abs{\xi_i}\ge q}}$ count the coordinates of magnitude at least~$q$.
Independence gives~$\E\nu_q=r\exp\prn{-q}\ge\sqrt r$ and~$\E\nu_q^2\le(\E\nu_q)^2+\E\nu_q\le2(\E\nu_q)^2$.
For a nonnegative random variable~$Y$ with~$0<\E Y^2<\infty$, the Paley--Zygmund inequality gives~$\Pr(Y\ge\theta\E Y)\ge(1-\theta)^2(\E Y)^2/\E Y^2$ for~$0<\theta<1$.
Applying it with~$Y=\nu_q$ and~$\theta=1/2$ gives~$\Pr(\nu_q\ge\E\nu_q/2)\ge1/8$.
Since~$\norm{\xi}_q\ge q\nu_q^{1/q}$, we obtain
\begin{equation}\label{eq:lp-width-lower-count}
\E\norm{\xi}_q\ge\frac q8\prn*{\frac{r\exp\prn{-q}}2}^{1/q}\ge\frac{qr^{1/q}}{16\mathrm e}.
\end{equation}
This proves the lower bound for~$2<q\le(\log r)/2$.

For~$r\ge2$ and~$q\ge(\log r)/2$, including~$q=\infty$, we use the largest absolute coordinate.
The variables~$\abs{\xi_i}$ are independent exponentials of mean~$1$, so
\begin{equation}\label{eq:laplace-absolute-maximum}
\E\max_{1\le i\le r}\abs{\xi_i}
=\int_0^1\frac{1-\prn{1-s}^r}{s}\,ds
=\sum_{j=1}^r\frac1j\ge\log\prn{r+1}.
\end{equation}
Here the first equality follows by integrating the tail probability and substituting~$s=\exp\prn{-t}$, and the second follows from the finite geometric sum.
Since~$r^{1/q}\le\mathrm e^2$, \eqref{eq:laplace-absolute-maximum} gives
\begin{equation}\label{eq:lp-width-lower-maximum}
\E\norm{\xi}_q\ge\E\max_{1\le i\le r}\abs{\xi_i}
\ge\log\prn{r+1}\ge\mathrm e^{-2}r^{1/q}\min\set{q,\log\prn{r+1}}.
\end{equation}
For~$r=1$, we have~$\E\norm{\xi}_q=\E\abs{\xi_1}=1$ for every~$q$.
Substituting this value for~$r=1$, and~\eqref{eq:lp-width-lower-count} or~\eqref{eq:lp-width-lower-maximum} for~$r\ge2$, into~\eqref{eq:unit-laplace-width-lower} proves the lower bound in~\eqref{eq:unit-lp-width} for~$p<2$.

\subsubsection{Initialization}\label{app:unit-initialization}\label{app:lp-upper}
We construct an admissible description of~$\B_p^r$ whose coordinate representation cost has the order proved in~\eqref{eq:unit-lp-width}.
Take~$N=r$, $w_\circ=0$, $A=I_r$, and~$\ell_i=1$ for every~$i\in\set{1,\ldots,r}$.
The unique coefficient vector representing~$u\in\B_p^r$ is~$v=u$, so~$C_{\mathcal D}(\B_p^r)=\sup_{v\in\B_p^r}\Phi_{\ell,\underline h}(v)$.

For~$p=\infty$, take~$J_i=0$ and~$\underline h_i=1$ for every~$i\in\set{1,\ldots,r}$.
For every~$v\in\B_\infty^r$, \eqref{eq:geometry} gives $\Phi_{\ell,\underline h}(v)=r+\norm{v}_1\le2r$, which has the required order~$r$.

Now let~$1\le p<\infty$. For each~$i\in\set{1,\ldots,r}$, choose
\begin{equation}\label{eq:lp-threshold}
J_i\coloneqq\ceil{\ceil{\log_2 r}/p}\quad\text{and}\quad \underline h_i\coloneqq2^{-J_i}.
\end{equation}
Since~$\log_2r\le\ceil{\log_2r}<\log_2r+1$ and~$p\ge1$, we have $\tfrac14r^{-1/p}<\underline h_i\le r^{-1/p}$ for every~$i\in\set{1,\ldots,r}$. The inner ceiling lets us compute~$J_i$ by doubling and comparisons (see \cref{app:initialization-arithmetic}).
Fix~$v\in\B_p^r$. We have~$L_{\underline h}\le r^{1/q}$ and~$\norm{v}_1\le r^{1/q}$.
To bound the logarithmic sum in~\eqref{eq:geometry}, use~$1/\underline h_i<4r^{1/p}$ from~\eqref{eq:lp-threshold} to obtain
$\log_+\prn{\abs{v_i}/\underline h_i}\le\log_+\prn{r^{1/p}\abs{v_i}}+\log4$ for every~$i\in\set{1,\ldots,r}$.
Multiplying by~$\abs{v_i}$ and summing adds at most~$\prn{\log4}\norm{v}_1$, so~\eqref{eq:geometry} gives
\begin{equation}\label{eq:lp-rounded-cost}
\Phi_{\ell,\underline h}(v)\le\prn{2+\log4}r^{1/q}+\sum_{i=1}^r\abs{v_i}\log_+\prn{r^{1/p}\abs{v_i}}.
\end{equation}

We bound the sum in~\eqref{eq:lp-rounded-cost} in two ways and use the smaller upper bound.
First, $\abs{v_i}\le1$ gives, for every~$1\le p<\infty$,
\begin{equation}\label{eq:lp-logbound-coordinate}
\sum_{i=1}^r\abs{v_i}\log_+\prn{r^{1/p}\abs{v_i}}
\le\frac{\log r}{p}\norm{v}_1\le\frac{r^{1/q}\log r}{p}.
\end{equation}
For~$p=1$, substituting~\eqref{eq:lp-logbound-coordinate} into~\eqref{eq:lp-rounded-cost} gives $\Phi_{\ell,\underline h}(v)\le2+\log4+\log r$.
For~$1<p<\infty$, we also apply $z\log_+z\le z^p/\brc{\mathrm e(p-1)}$ for~$z\ge0$ at~$z=r^{1/p}\abs{v_i}$ and sum to obtain
\begin{equation}\label{eq:lp-logbound}
\sum_{i=1}^r\abs{v_i}\log_+\prn{r^{1/p}\abs{v_i}}
\le\frac{r^{-1/p}}{\mathrm e(p-1)}\sum_{i=1}^r\prn{r^{1/p}\abs{v_i}}^p
\le\frac{r^{1/q}}{\mathrm e(p-1)}.
\end{equation}
Using~\eqref{eq:lp-logbound-coordinate} or~\eqref{eq:lp-logbound} in~\eqref{eq:lp-rounded-cost} bounds~$\Phi_{\ell,\underline h}(v)$ by universal multiples of~$r^{1/q}\log\prn{r+1}$ and~$r^{1/q}q$, respectively, since~$1+1/\prn{p-1}=q$.
Taking the smaller upper bound and including the cases~$p=1$ and~$p=\infty$ proves
\begin{equation}\label{eq:lp-upper}
\sup_{v\in\B_p^r}\Phi_{\ell,\underline h}(v)
\lesssim r^{1/q}\min\set{q,\log\prn{r+1}}.
\end{equation}
For~$r=1$, the threshold is~$\underline h_1=1$ and $\Phi_{\ell,\underline h}(v)=1+\abs{v_1}\le2$.
The constant is uniform over~$r\ge1$ and~$p\in[1,\infty]$.

By~\eqref{eq:unit-lp-width}, the bound in~\eqref{eq:lp-upper} gives~$C_{\mathcal D}(\B_p^r)\lesssim\width(\B_p^r)$, completing the construction.

\subsection{Extensions to Other Domains}\label{app:domain-extensions}
We construct admissible descriptions for translated and scaled norm balls, ellipsoids, boxes, and simplices. As for the unit balls in~\cref{app:unit-initialization}, each description uses at most~$d$ coordinates and has cost proportional to the width of the domain.
\begin{corollary}\label{cor:canonical-width}
Let~$\mathcal K\subseteq\R^d$ be a unit norm ball~$\B_p^d$ with~$1\le p\le\infty$, a Euclidean ball~$R\B_2^d$ with~$R>0$, an ellipsoid or rectangular box of dimension~$1\le r\le d$, or the simplex~$\triangle_d$ with~$d\ge2$.
Then there is an explicit admissible description~$\mathcal D$ with~$N\le d$ and~$C_{\mathcal D}(\mathcal K)\lesssim\width(\mathcal K)$. The implicit constant is universal.
\end{corollary}
To prove~\cref{cor:canonical-width}, we first transfer the unit-ball description under translations, isometric embeddings, and uniform scaling. We then construct descriptions along the principal axes of ellipsoids and boxes and restrict the unit~$\ell_1$ ball construction to the simplex.

\subsubsection{Translated and Scaled Norm Balls}\label{app:scaled-norm-balls}
We transfer the width and admissible description of~$\B_p^r$ to~$\mathcal K=w_\circ+RU\B_p^r$.
Let~$1\le p\le\infty$, $1/p+1/q=1$, $R>0$, $1\le r\le d$, $w_\circ\in\R^d$, and~$U\in\R^{d\times r}$ with~$U^\top U=I_r$.
By~\cref{item:width-transform} and~\eqref{eq:unit-lp-width}, we have
\begin{equation}\label{eq:scaled-lp-width}
\width(\mathcal K)=R\width(\B_p^r)
\asymp Rr^{1/q}\min\set{q,\log\prn{r+1}}.
\end{equation}
Use~$\mathcal D\coloneqq(w_\circ,RU,R\mathbf1,\underline h)$, where~$\underline h$ is chosen in~\cref{app:unit-initialization}.
Each column of~$RU$ has norm~$R$, and every~$u\in\mathcal K$ has unique coefficients~$v\in\B_p^r$ satisfying~$u=w_\circ+RUv$.
Writing~$\mathcal D_0\coloneqq(0,I_r,\mathbf1,\underline h)$ for the unit-ball description, we obtain
\begin{equation}\label{eq:cost-scaling}
\Phi_{R\mathbf1,\underline h}(v)=R\Phi_{\mathbf1,\underline h}(v)
\quad\text{and}\quad
C_{\mathcal D}(\mathcal K)=R C_{\mathcal D_0}(\B_p^r).
\end{equation}
Equations~\eqref{eq:lp-upper} and~\eqref{eq:scaled-lp-width} therefore give~$C_{\mathcal D}(\mathcal K)\lesssim\width(\mathcal K)$ with~$N=r\le d$.
In particular, the Euclidean ball~$R\B_2^d$ has width~$\asymp R\sqrt d$ and an admissible description with this order of coordinate representation cost.

\subsubsection{Ellipsoids and Boxes}\label{app:canonical-initialization}\label{app:ellipsoid-upper}
We bound the width and construct admissible descriptions for ellipsoids and boxes with arbitrary positive axis lengths.
Take~$1\le r\le d$, $w_\circ\in\R^d$, $U\in\R^{d\times r}$ with~$U^\top U=I_r$, and~$a\in(0,\infty)^r$.
The ellipsoid with semiaxis lengths~$a_i$ and the box with half-widths~$a_i$ are, respectively,
\begin{equation}\label{eq:ellipsoid-def}
\mathcal K=w_\circ+U\diag(a_i)\B_2^r
\quad\text{and}\quad
\mathcal K=w_\circ+U\diag(a_i)\B_\infty^r.
\end{equation}
These are affine images of norm balls \citep[Sections 2.2.2 and 2.3.2]{boyd2004convex}.
For their common calculation, let~$p\in\set{2,\infty}$, $1/p+1/q=1$, and write
\begin{equation}\label{eq:norm-ball-domain}
\mathcal K=w_\circ+A\B_p^r,
\qquad A\coloneqq U\diag(a_i).
\end{equation}
The support function satisfies
\begin{equation}\label{eq:norm-ball-support}
\sigma_{\mathcal K}(z)=\inpr{w_\circ,z}+\norm{A^\top z}_q
\qquad(z\in\R^d).
\end{equation}
\begin{proposition}\label{prop:norm-ball-width}
For the ellipsoid and box in~\eqref{eq:norm-ball-domain}, we have
\begin{equation}\label{eq:norm-ball-width}
\width(\mathcal K)\asymp\norm{a}_q.
\end{equation}
The implicit constants are universal.
\end{proposition}
\begin{proof}
By~\cref{item:width-transform}, it suffices to bound~$\width(\diag(a_i)\B_p^r)$.
For every~$Z\in\R^r$ satisfying~\eqref{eq:width-vector-constraints}, the support function in~\eqref{eq:norm-ball-support}, Jensen's inequality, and~$\E Z_i^2\le2$ give
\begin{equation}\label{eq:weighted-lp-width-upper}
\E\sigma_{\diag(a_i)\B_p^r}(Z)
=\E\norm{\diag(a_i)Z}_q
\le\prn*{\sum_{i=1}^r a_i^q\E\abs{Z_i}^q}^{1/q}
\le\sqrt2\norm{a}_q.
\end{equation}
For the lower bound, take~$\xi\in\R^r$ with the independent symmetric Laplace coordinates used in~\cref{app:width-lower}.
The vector~$\xi/4$ satisfies~\eqref{eq:width-vector-constraints}, so~$U\xi/4$ satisfies the corresponding constraints in~$\R^d$ and gives
\begin{equation}\label{eq:laplace-width-lower}
\width(\mathcal K)\ge\E\sigma_{\mathcal K}(U\xi/4)
=\frac14\E\norm{\diag(a_i)\xi}_q.
\end{equation}

Jensen's inequality and~$\E\abs{\xi_i}=1$ give
\begin{equation}\label{eq:weighted-lp-width-lower}
\E\norm{\diag(a_i)\xi}_q
\ge\norm{\E\brc{\prn{a_i\abs{\xi_i}}_{i=1}^r}}_q
=\norm{a}_q.
\end{equation}
Together, \eqref{eq:weighted-lp-width-upper}, \eqref{eq:laplace-width-lower}, and~\eqref{eq:weighted-lp-width-lower} give~$\norm{a}_q/4\le\width(\mathcal K)\le\sqrt2\norm{a}_q$.
This proves~\eqref{eq:norm-ball-width}, with~$\norm{a}_2$ for the ellipsoid and~$\norm{a}_1$ for the box.
\end{proof}

\paragraph{Initialization.}
We use principal-axis directions to construct admissible descriptions with~$C_{\mathcal D}(\mathcal K)\lesssim\width(\mathcal K)$ for ellipsoids and boxes.
For the ellipsoid in~\eqref{eq:ellipsoid-def}, take~$A=U\diag(a_i)$ and, for each~$i\in\set{1,\ldots,r}$, let~$\ell_i=a_i$ and choose a dyadic~$\underline h_i$ satisfying $a_i/\prn{2\norm{a}_2}<\underline h_i\le a_i/\norm{a}_2$.
Fix~$v\in\B_2^r$. The threshold choice gives~$L_{\underline h}\le\norm{a}_2$, and the Cauchy--Schwarz inequality gives $\sum_{i=1}^r a_i\abs{v_i}\le\norm{a}_2$.
Applying~$x\log_+x\le x^2/\mathrm e$ for~$x\ge0$ at~$x=\norm{a}_2\abs{v_i}/a_i$ and summing gives
\begin{equation}\label{eq:ellipsoid-log}
\sum_{i=1}^r a_i\abs{v_i}\log_+\frac{\norm{a}_2\abs{v_i}}{a_i}
\le\frac{\norm{a}_2}{\mathrm{e}}\sum_{i=1}^r v_i^2
\le\frac{\norm{a}_2}{\mathrm{e}}.
\end{equation}
Since~$1/\underline h_i<2\norm{a}_2/a_i$, we have $\log_+\prn{\abs{v_i}/\underline h_i}\le\log_+\prn{\norm{a}_2\abs{v_i}/a_i}+\log2$ for every~$i\in\set{1,\ldots,r}$.
Substituting this inequality and~\eqref{eq:ellipsoid-log} into~\eqref{eq:geometry} yields
\begin{equation}\label{eq:ellipsoid-coefficient-cost}
\Phi_{\ell,\underline h}(v)\le L_{\underline h}+\prn{1+\log2}\sum_{i=1}^r a_i\abs{v_i}+\frac{\norm{a}_2}{\mathrm e}
\le\prn{2+\log2+1/\mathrm e}\norm{a}_2\lesssim\norm{a}_2.
\end{equation}
Every~$u\in\mathcal K$ has a representation~$u=w_\circ+Av$ with~$v\in\B_2^r$.
Taking the supremum over~$v\in\B_2^r$ in~\eqref{eq:ellipsoid-coefficient-cost} and using~\eqref{eq:norm-ball-width} gives $C_{\mathcal D}(\mathcal K)\lesssim\norm{a}_2\lesssim\width(\mathcal K)$.

For the box $\mathcal K=w_\circ+U\diag(a_i)\B_\infty^r$, use~$A=U\diag(a_i)$, $\ell_i=a_i$, and~$\underline h_i=1$ for every~$i\in\set{1,\ldots,r}$.
Every~$u\in\mathcal K$ has a representation~$u=w_\circ+Av$ with~$v\in[-1,1]^r$.
The logarithmic terms in~\eqref{eq:geometry} vanish, so we have $\Phi_{\ell,\underline h}(v)=\sum_{i=1}^r a_i\prn{1+\abs{v_i}}\le2\norm{a}_1$.
Taking the supremum over~$v\in[-1,1]^r$ and using~\eqref{eq:norm-ball-width} yields
\begin{equation}\label{eq:box-representation-cost}
C_{\mathcal D}(\mathcal K)\le2\norm{a}_1\lesssim\width(\mathcal K).
\end{equation}
This gives an admissible description with~$N=r\le d$.

\subsubsection{Simplex}\label{app:simplex-extension}
We bound the simplex width by comparing support functions and use the same initialization as for the unit~$\ell_1$ ball.
The simplex $\triangle_d\coloneqq\Set{u\in[0,1]^d}{\sum_{i=1}^d u_i=1}$ is the intersection of~$\B_1^d$ with the hyperplane~$\Set{u\in\R^d}{\sum_{i=1}^d u_i=1}$.
For~$d\ge2$, its width satisfies
\begin{equation}\label{eq:simplex-comparison}
\frac12\width(\B_1^{d-1})\le\width(\triangle_d)\le\width(\B_1^d).
\end{equation}
The upper bound follows from~$\triangle_d\subseteq\B_1^d$ and the corresponding inequality between support functions.
For the lower bound, let~$Y\in\R^{d-1}$ satisfy the constraints in~\eqref{eq:width-recall} in dimension~$d-1$.
Both~$(Y,0)$ and~$(-Y,0)$ satisfy the constraints in dimension~$d$, and
\begin{equation}
\norm{Y}_\infty
\le\max\set{0,Y_1,\ldots,Y_{d-1}}+\max\set{0,-Y_1,\ldots,-Y_{d-1}}
=\sigma_{\triangle_d}(Y,0)+\sigma_{\triangle_d}(-Y,0).
\end{equation}
Taking expectations and then the supremum over~$Y$ proves the lower bound in~\eqref{eq:simplex-comparison}.
By~\eqref{eq:unit-lp-width} at~$p=1$, the comparison~\eqref{eq:simplex-comparison} gives~$\width(\triangle_d)\asymp\log\prn{d+1}$.
For initialization, take~$w_\circ=0$, $A=I_d$, and~$\ell_j=1$ for every~$j\in\set{1,\ldots,d}$, and use~\eqref{eq:lp-threshold} with~$r=d$ and~$p=1$.
Every~$u\in\triangle_d$ is represented by~$v=u\in\B_1^d$.
Taking the supremum over~$u\in\triangle_d$ and using~\eqref{eq:lp-upper} and~\eqref{eq:simplex-comparison} gives~$C_{\mathcal D}(\triangle_d)\lesssim\width(\triangle_d)$ with~$N=d$.
\subsection{Comparisons with Other Widths}\label{app:width-comparisons}
To explain our use of the maximal subexponential mean width, we examine below whether Gaussian and exponential widths match minimax regret and remain invariant under rotations.
The regret comparisons below use the Euclidean bound~$\norm{g_t}_2\le G$ with~$G=1$.

\paragraph{Gaussian Width.}
Expected support functions are used to express geometric complexity \citep[Sections 7.5 and 2.8]{vershynin2026hdp}.
The \emph{Gaussian width} of~$\mathcal K$ is given by $\E \sigma_{\mathcal K}(Z)$ for a standard Gaussian vector~$Z\in\R^d$.
Neither this width nor its square has the order of the minimax regret for~$b=0$ on both the simplex and the Euclidean unit ball.
On~$\triangle_d$, the Gaussian width is~$\Theta\prn{\sqrt{\log d}}$, which
falls short of the~$\Omega\prn{\log d}$ lower bound in \cref{app:clean-lower}.
Its square has order~$\log d$ on the simplex but order~$d$
on the Euclidean unit ball, where the minimax regret is of order~$\sqrt d$ by \cref{thm:minimax}.
The square would also scale quadratically with the radius of the domain.
For these reasons, we do not adopt the Gaussian width as our geometric complexity measure.

\paragraph{Exponential Width and Rotations.}
The \emph{exponential width} of~$\mathcal K$ is $\E\sigma_{\mathcal K}(\xi)$, where~$\xi\in\R^d$ has independent symmetric Laplace coordinates, each with density~$\exp\prn{-\abs{x}}/2$ on~$\R$ \citep[Section 3.1]{sivakumar2015beyond}.
Below we see that the order of this width can change under rotations of~$\mathcal K$. For a normalized Hadamard matrix~%
$Q$ of dimension~$d$, the moment generating function of each coordinate
of~$Q^\top\xi$ is $\prn{1-t^2/d}^{-d}\le\exp\prn{4t^2/3}$ when~%
$\abs{t}\le\sqrt d/2$. The log-sum-exponential bound with~%
$t\asymp\sqrt{\log d}$ therefore gives
$\E \sigma_{Q\triangle_d}(\xi)=O\prn{\sqrt{\log d}}$.
For the same law of~$\xi$, the variables~$\abs{\xi_i}$ are independent exponentials of mean one. Their largest and smallest values have expectations~$\sum_{j=1}^d1/j$ and~$1/d$, respectively. Symmetry and the pointwise inequality~$\max_i\xi_i-\min_i\xi_i\ge\max_i\abs{\xi_i}-\min_i\abs{\xi_i}$ give, for~$d\ge2$,
\begin{equation}\label{eq:laplace-maximum}
\E\sigma_{\triangle_d}(\xi)=\E\max_i\xi_i
=\frac12\E\brc{\max_i\xi_i-\min_i\xi_i}
\ge\frac12\sum_{j=1}^{d-1}\frac1j\gtrsim\log d.
\end{equation}
Comparing this lower bound with the upper bound for~$Q\triangle_d$ shows that the order of the exponential width can change under rotations. In contrast, the maximal subexponential mean width satisfies $\width(Q\triangle_d)=\width(\triangle_d)$ by~\eqref{eq:width-transform}, and \cref{app:simplex-extension} gives~$\width(\triangle_d)\asymp\log d$.

\section{Extension to General Domains}\label{app:core-proof}
We prove~\cref{thm:main} on general domains, establishing the bound on~$2R_T^{\mathrm{lin}}(u)-\eps R_T^{\mathrm{abs}}(u)$.
To this end, we extend the Euclidean-ball analysis using the coefficient cost~$\Phi_{\ell,\underline h}$ for admissible descriptions from~\cref{app:geometry}.

\subsection{Finite Coordinate Initialization}\label{sec:representation}\label{app:general-core}
We first specify the coordinate initialization and the unconstrained update used on general domains.
Fix~$\eps\in(0,1]$ and an admissible description~$\mathcal D=\prn{w_\circ,A,\ell,\underline h}$ from
\cref{app:representation}. The columns satisfy~$\norm{Ae_j}_2\le\ell_j$ for~$j\in\set{1,\ldots,N}$,
and each comparator has a representation~%
$u=w_\circ+Av$ with~$v\in[-1,1]^N$. For each~$j\in\set{1,\ldots,N}$,
write~$\underline h_j=2^{-J_j}$ with~$J_j\ge0$ an integer.
The unconstrained iterates~$z_t\in\R^N$ start at~$z_1\coloneqq0$.
In each round, the update receives a vector~$\widetilde g_t\in\R^N$.

\subsubsection{Initial Regularization Terms}
The update below uses the fixed positive definite matrix
$B\coloneqq\lambda\diag(\ell_1,\ldots,\ell_N)$, where~%
$\lambda\coloneqq16/\eps$.

For every coordinate~$j\in\set{1,\ldots,N}$ and level
$k\in\set{0,\ldots,J_j-1}$, initialize a quadratic regularization term with direction~$e_j$,
revocation threshold~$h_{jk}$, and coefficient~$\gamma_{jk}$ given by
\begin{equation}\label{eq:initial-quadratics}
 h_{jk}\coloneqq2^k\underline h_j,\qquad
 \gamma_{jk}\coloneqq\frac{\lambda\ell_j}{h_{jk}}\qquad\text{and}\qquad Q^0_{jk}(z)\coloneqq\frac{\gamma_{jk}}2z_j^2.
\end{equation}
There are~$M_0\coloneqq\sum_{j=1}^N J_j$ such terms; a coordinate
with~$J_j=0$ has none. The initial regularizer is
\begin{equation}\label{eq:initial-psi}
 \Psi_1(z)\coloneqq\frac12\norm{z}_B^2+
 \sum_{j=1}^N\sum_{k=0}^{J_j-1}Q^0_{jk}(z).
\end{equation}
Initialize~$H_1\coloneqq\nabla^2\Psi_1$ and~$\mathcal Q_1$ with entries~%
$\prn{e_j,0,\gamma_{jk},h_{jk}}$ indexed by these pairs~$(j,k)$.
The objective~$F_t$ is the sum of~$\Psi_t$ and the past linear losses on~$\R^N$,
as in \cref{sec:discrete-step}. The stability is
$\mathrm{Stab}_t=F_t(z_t)+\inpr{\widetilde g_t,z_t}-F_{t+1}(z_{t+1})$
as in \cref{sec:lifetime}.
In~\cref{alg:move-revoke}, the threshold test for~$Q^0_{jk}$ is~$\abs{z_j}\ge h_{jk}$ at the current point~$z$.
When it holds, the term is revoked by the divergence subtraction in~\eqref{eq:ball-cancellation}.

\subsubsection{Round Updates and Omitted Additions}
To bound the storage and arithmetic costs, we allow an addition to be omitted
when~$s_t^2\le\zeta_t$, with
\begin{equation}\label{eq:general-skip}
 L_{\underline h}\coloneqq\sum_{j=1}^N\ell_j\underline h_j,\qquad
 \zeta_t\coloneqq L_{\underline h}(t+1)^{-4}\quad\text{and}\quad E_T^{\mathrm{skip}}\coloneqq
 \sum_{\substack{1\le t\le T,\ \text{no term}\\\text{added in round }t}}s_t^2
 <\frac{L_{\underline h}}{8}.
\end{equation}
Here we have~$s_t^2=0$ for a zero input. The last inequality follows from
$\sum_{t=1}^{\infty}(t+1)^{-4}<1/8$.
On a round with an addition, use the exact parameters in~%
\eqref{eq:parameters}. They satisfy
\begin{equation}\label{eq:weight-range}
 \tau/2<\gamma_t s_t^2\le\tau\quad\text{and}\quad
 \gamma_th_t=\eps/6,\quad\text{where}\;\; \tau\coloneqq\eps^2/48.
\end{equation}
In particular, we have~$h_t\ge8s_t^2/\eps$. 

\Cref{alg:core} gives the unconstrained updates for arbitrary supplied
vectors, provided that the initial data satisfy the assumptions of~\cref{lem:move-revoke} at~$t=1$. It uses the movement step of \cref{alg:ball-oco} and adds the regularization term whenever~$s_t^2>\zeta_t$. The prediction and feedback rules are
supplied separately, as in the coordinate reduction given in \cref{app:projection}. 
Note that the entries in~$\mathcal Q_1$ for the initial terms~$Q^0_{jk}$ in~$\Psi_1$ use indices~$(j,k)$, distinct from the positive integers indexing terms added during the rounds.

\begin{algorithm2e}
\caption{Unconstrained Updates with Curvature Revocation}\label{alg:core}
\KwIn{$\eps$; thresholds~$\zeta_t\ge0$; initial~$z_1,\Psi_1,H_1,\mathcal Q_1$.}
\KwOut{Unconstrained iterates~$z_1,z_2,\ldots$.}
\For{$t=1,2,\ldots$}{
Output~$z_t$ and receive~$\widetilde g_t$\;
Obtain $z_{t+1},s_t^2,\Psi_t^-,H_t^-,\mathcal Q_t^-$ by calling
\textsc{MoveAndRevoke}$\prn{z_t,\widetilde g_t,\Psi_t,H_t,\mathcal Q_t}$\;
$\prn{\Psi_{t+1},H_{t+1},\mathcal Q_{t+1}}\leftarrow\prn{\Psi_t^-,H_t^-,\mathcal Q_t^-}$\;
\If{$s_t^2>\zeta_t$}{
Set~$q_t\leftarrow\widetilde g_t$, $m_t\leftarrow\inpr{q_t,z_{t+1}}$\;
Set~$\gamma_t\leftarrow\eps^2/\prn{48s_t^2}$, $h_t\leftarrow8s_t^2/\eps$\;
Set $\Psi_{t+1}\leftarrow\Psi_{t+1}+Q_t$\;
Add~$\prn{q_t,m_t,\gamma_t,h_t}$ to~$\mathcal Q_{t+1}$ as entry~$t$\;
Set $H_{t+1}\!\leftarrow\! H_{t+1}\!+\!\gamma_tq_tq_t^\top$\;}}
\end{algorithm2e}

\subsection{Applying Movement and Cancellation Bounds}\label{app:core-setup}\label{app:movement}\label{app:movement-proof}
We extend the movement properties and cancellation bounds to a positive definite fixed matrix~$B$ and a finite collection of initial regularization terms centered at zero.
This includes the coordinate initialization in \cref{app:general-core} and the initialization with~$\Psi_1(z)=\norm{z}_B^2/2$ and~$\mathcal Q_1=\varnothing$.
The arguments through \cref{app:survivor-proof} apply to both choices.

\subsubsection{Movement and Threshold Bounds}
The properties in
\crefrange{item:move-revoke-curvature}{item:move-revoke-thresholds} apply in~$\R^N$ with this finite initial collection. Initially, $z_1=0$ minimizes~$F_1$
and satisfies~$\inpr{q_i,z_1}-m_i=0$ for every initial term~$i$.
At the returned point~$z_{t+1}$, every remaining term satisfies
$\abs{\inpr{q_i,z_{t+1}}-m_i}<h_i$ by \cref{item:move-revoke-thresholds}.
A new term has~$m_t=\inpr{q_t,z_{t+1}}$, so it also satisfies this inequality
and has zero gradient at~$z_{t+1}$.
Together with~$\nabla F_t^-(z_{t+1})=0$ from \cref{item:move-revoke-optimality},
this proves the minimization and threshold assumptions for the next call.
These properties require no norm bound on~$\widetilde g_t$.

Since any term added at~$z_{t+1}$ vanishes there, we have $F_{t+1}\prn{z_{t+1}}=F_t^-\prn{z_{t+1}}$. For every round, including zero inputs, \cref{item:move-revoke-decrease} therefore gives
\begin{equation}\label{eq:movement}
0\le\mathrm{Stab}_t\le s_t^2/2\quad\text{and}\quad 0\le\delta_t\coloneqq\inpr{\widetilde g_t,z_t-z_{t+1}}\le s_t^2.
\end{equation}
\Cref{item:move-revoke-thresholds} gives, at each revocation point~$a_i$,
\begin{equation}\label{eq:overshoot-proof}
h_i\le\abs{\inpr{q_i,a_i}-m_i}<3h_i.
\end{equation}
We next apply the argument of \cref{lem:lifetime} to the terms added during play, using~\eqref{eq:weight-range}.

\subsubsection{Regularization and Stability in Each Round}\label{app:lifetime-proof}
Let~$\mathcal I_T$ and~$\mathcal C_T$ index the surviving and revoked terms added during the rounds, respectively. Fix~$v\in\R^N$ and write~$\widetilde r_i\coloneqq\inpr{\widetilde g_i,z_i-v}$ for the excess loss at~$z_i$ relative to~$v$ in round~$i$.
For each such term~$Q_i$, let~$z_i^{\mathrm{end}}$ be its actual
revocation point if~$i\in\mathcal C_T$, and~$z_{T+1}$ if~%
$i\in\mathcal I_T$. Define
$\rho_i\coloneqq\inpr{\widetilde g_i,z_i^{\mathrm{end}}-z_{i+1}}$.
As in the analysis on the ball, we bound~$Q_i(v)-D_{Q_i}(v,z_i^{\mathrm{end}})+\mathrm{Stab}_i$ for each term added in round~$i$.
The identity~\eqref{eq:ball-bregman-difference} gives
\begin{equation}\label{eq:general-observed-difference}
 Q_i(v)-D_{Q_i}(v,z_i^{\mathrm{end}})+\mathrm{Stab}_i
 =\frac{\gamma_i}{2}\brc{\delta_i^2-\prn{\delta_i-\rho_i}^2}
   +\mathrm{Stab}_i-\gamma_i\rho_i\widetilde r_i.
\end{equation}
For a surviving term with~$i<T$, \cref{item:move-revoke-thresholds} gives~%
$\abs{\rho_i}<h_i$. A term added in round~$T$ has~$\rho_T=0$ since~$z_T^{\mathrm{end}}=z_{T+1}$. For a revoked term, \cref{item:move-revoke-thresholds}
gives~$\abs{\rho_i}<3h_i$. In both cases we have
$\abs{\gamma_i\rho_i}\le3\gamma_i h_i=\eps/2$ by~\eqref{eq:weight-range}. If the term is revoked, %
\eqref{eq:movement} and~$h_i\ge8s_i^2/\eps$ imply
\begin{equation}\label{eq:removal-payment}
 \frac{\gamma_i}{2}\brc{\prn{\delta_i-\rho_i}^2-\delta_i^2}
 \ge\frac{\gamma_i h_i}{2}(h_i-2s_i^2)
 \ge\frac{4-\eps}{6}s_i^2\ge\mathrm{Stab}_i.
\end{equation}
The first inequality follows by minimizing over~%
$\abs{\rho_i}\ge h_i$ and~$0\le\delta_i\le s_i^2$.
For a surviving term, use $-\gamma_i\prn{\delta_i-\rho_i}^2/2\le0$ and~$\gamma_i\delta_i^2\le\tau s_i^2$. 
Thus, as in the proof of \cref{lem:lifetime}, we have
\begin{equation}\label{eq:general-term-goal}
 Q_i(v)-D_{Q_i}(v,z_i^{\mathrm{end}})+\mathrm{Stab}_i
 \le\frac{\eps}{2}\abs{\widetilde r_i}+
 \begin{cases}
 \prn{1+\tau}s_i^2/2,&i\in\mathcal I_T,\\
 0,&i\in\mathcal C_T.
 \end{cases}
\end{equation}

\subsection{Regret Identity with Initial Regularization Terms}\label{sec:initial-proof}
We express the regularizer and the terminal objective difference as sums over initial terms and terms added during the rounds.
Let~$\mathcal I_T^0$ and~$\mathcal C_T^0$ partition the initial pairs~%
$(j,k)$ with~$j\in\set{1,\ldots,N}$ and $k\in\set{0,\ldots,J_j-1}$
according to whether their regularization term survives or has been revoked.
For each such pair, let~$x_{jk}^{\mathrm{end}}$ be the~$j$th coordinate
of its actual revocation point, or~$(z_{T+1})_j$ if it survives.
The divergence of~$Q^0_{jk}$ depends only on this coordinate of its
second argument.
For each initial pair~$(j,k)$, we have
\begin{equation}\label{eq:initial-one-identity}
 Q^0_{jk}(v)-D_{Q^0_{jk}}(v,x_{jk}^{\mathrm{end}}e_j)
 =\frac{\gamma_{jk}}{2}\brc{2x_{jk}^{\mathrm{end}}v_j-\prn{x_{jk}^{\mathrm{end}}}^2}.
\end{equation}
This identity follows directly from
$D_{Q^0_{jk}}(v,x_{jk}^{\mathrm{end}}e_j)=\gamma_{jk}(v_j-x_{jk}^{\mathrm{end}})^2/2$.

To express the FTRL regret identity~\eqref{eq:ball-pen-stab} in terms of the initial and subsequent regularization terms, define
\begin{equation}\label{eq:initial-contribution}
 P_T^{\mathrm{init}}(v)\coloneqq
 \sum_{j=1}^N\sum_{k=0}^{J_j-1}
 \gamma_{jk}\brc{2x_{jk}^{\mathrm{end}}v_j-\prn{x_{jk}^{\mathrm{end}}}^2}.
\end{equation}
Adding the regularization terms created during the rounds to~$\Psi_1$ and subtracting the divergences
of all revoked terms gives
\begin{equation}\label{eq:general-psi-terminal}
\begin{aligned}
 \Psi_{T+1}(v)
 =\frac12\norm{v}_B^2&+
   \sum_{j=1}^N\sum_{k=0}^{J_j-1}Q^0_{jk}(v)
   -\sum_{(j,k)\in\mathcal C_T^0}D_{Q^0_{jk}}(v,x_{jk}^{\mathrm{end}}e_j)\\
 &+\sum_{i\in\mathcal I_T\cup\mathcal C_T}Q_i(v)
   -\sum_{i\in\mathcal C_T}D_{Q_i}(v,z_i^{\mathrm{end}}).
\end{aligned}
\end{equation}
Since~$\nabla F_{T+1}(z_{T+1})=0$, Taylor's formula gives
\begin{equation}\label{eq:general-terminal-gap}
F_{T+1}(v)-F_{T+1}(z_{T+1}) =\frac12\norm{v-z_{T+1}}_B^2+ \sum_{(j,k)\in\mathcal I_T^0}D_{Q^0_{jk}}(v,z_{T+1}) +\sum_{i\in\mathcal I_T}D_{Q_i}(v,z_{T+1}).
\end{equation}
Here, we have~$\nabla^2\prn{Q-D_Q(\cdot,a)}=0$ for every revoked term~$Q$ at~$a$, and hence $\nabla^2F_{T+1}$ is the sum of~$B$ and the Hessians of the surviving terms.
Substituting the expressions for~$\Psi_{T+1}(v)$ and $F_{T+1}(v)-F_{T+1}(z_{T+1})$
into the FTRL identity~\eqref{eq:ball-pen-stab}, using~%
$\Psi_1(z_1)=0$ and the definition~\eqref{eq:initial-contribution}, gives
\begin{equation}\label{eq:general-pen-stab}
\begin{aligned}
 &\sum_{t=1}^T\widetilde r_t+\frac12\norm{v-z_{T+1}}_B^2\\
 &=\frac12\norm{v}_B^2+\frac12P_T^{\mathrm{init}}(v)
 +\sum_{\substack{1\le t\le T\\t\notin\mathcal I_T\cup\mathcal C_T}}\mathrm{Stab}_t
 +\sum_{i\in\mathcal I_T\cup\mathcal C_T}
 \brc{Q_i(v)-D_{Q_i}(v,z_i^{\mathrm{end}})+\mathrm{Stab}_i}.
\end{aligned}
\end{equation}
The rounds~$t\notin\mathcal I_T\cup\mathcal C_T$ are those with~$s_t^2\le\zeta_t$, in which no term was added. Their stability may still be positive.
Applying~\eqref{eq:general-term-goal} to the sum over~$i\in\mathcal I_T\cup\mathcal C_T$ and~%
\eqref{eq:movement} to the rounds with omitted additions in the preceding identity,
then multiplying both sides by~$2$ and subtracting~%
$\eps\sum_{t=1}^T\abs{\widetilde r_t}$, yields
\begin{equation}\label{eq:intermediate}
2\sum_{t=1}^T\widetilde r_t-\eps\sum_{t=1}^T\abs{\widetilde r_t} +\norm{z_{T+1}-v}_B^2 \le\norm{v}_B^2+P_T^{\mathrm{init}}(v) +\prn{1+\tau}\sum_{i\in\mathcal I_T}s_i^2+E_T^{\mathrm{skip}}.
\end{equation}
For each round with an omitted addition, the subtraction leaves~$-\eps\abs{\widetilde r_t}$ on the right. Dropping these nonpositive terms gives the upper bound in~\eqref{eq:intermediate}.
With no initial terms, we have $P_T^{\mathrm{init}}(v)=0$, so
this bound extends \cref{lem:lifetime} to rounds with
omitted additions through~$E_T^{\mathrm{skip}}$. Specifically, we have
\begin{equation}\label{eq:full-lifetime}
 2\sum_{t=1}^T\widetilde r_t-\eps\sum_{t=1}^T\abs{\widetilde r_t}
 +\norm{z_{T+1}-v}_B^2
 \le\norm{v}_B^2+\prn{1+\tau}\sum_{i\in\mathcal I_T}s_i^2
 +E_T^{\mathrm{skip}}.
\end{equation}
We next bound~$\sum_{i\in\mathcal I_T}s_i^2$ in~\eqref{eq:intermediate}
using the surviving initial terms, and then combine this bound with~%
$P_T^{\mathrm{init}}(v)$.

\subsection{Weighted Trace for Surviving Terms}\label{app:survivor-proof}
We bound~$\sum_{i\in\mathcal I_T}s_i^2$ using the Hessian of the initial regularizer with the revoked curvature removed.
Suppose there is a fixed positive semidefinite
matrix~$\Sigma\in\R^{N\times N}$ such that~%
$\widetilde g_t\widetilde g_t^\top\preceq\Sigma$ for every~$t$.
Add the Hessians of surviving initial terms to~$B$ to define
\begin{equation}\label{eq:surviving-initial-matrix}
 B_T\coloneqq B+\sum_{(j,k)\in\mathcal I_T^0}
 \gamma_{jk}e_je_j^\top.
\end{equation}
For~$i\in\mathcal I_T$, define
$\overline H_i\coloneqq B_T+\sum_{j\in\mathcal I_T,\,j<i}
\gamma_j\widetilde g_j\widetilde g_j^\top$ and~%
$\overline s_i^2\coloneqq\widetilde g_i^\top\overline H_i^{-1}\widetilde g_i$.
Every regularization term whose Hessian appears in~$\overline H_i-B$ is active immediately before~%
$Q_i$ is added, so we have~$\overline H_i\preceq H_i^-$ and~$\overline s_i^2\ge s_i^2$.

Multiplying the Sherman--Morrison formula by~$\Sigma$ and taking the trace gives
\begin{equation}\label{eq:trace-drop}
\tr(\Sigma\overline H_i^{-1}) -\tr\prn*{\Sigma(\overline H_i+\gamma_i\widetilde g_i\widetilde g_i^\top)^{-1}} =\frac{\gamma_i\widetilde g_i^\top\overline H_i^{-1} \Sigma\overline H_i^{-1}\widetilde g_i}{1+\gamma_i\overline s_i^2} \ge\frac{\gamma_i\prn{\widetilde g_i^\top\overline H_i^{-1}\widetilde g_i}^2}{1+\gamma_i\overline s_i^2} \ge\frac{\tau}{2+\tau}s_i^2.
\end{equation}
The first inequality uses~$\Sigma\succeq\widetilde g_i\widetilde g_i^\top$;
the last uses~$\overline s_i^2\ge s_i^2$ and~%
$\gamma_i s_i^2>\tau/2$ from~\eqref{eq:weight-range}.
In increasing order of~$i\in\mathcal I_T$, the updates in~\eqref{eq:trace-drop} start at~$B_T$ and add the Hessian of each surviving round term once, giving~$H_{T+1}$. The trace differences therefore sum to~$\tr(\Sigma B_T^{-1})-\tr(\Sigma H_{T+1}^{-1})$. Using~$\tr(\Sigma H_{T+1}^{-1})\ge0$ yields
\begin{equation}\label{eq:survivor-bound}
 \prn{1+\tau}\sum_{i\in\mathcal I_T}s_i^2
 \le c_\eps\tr(\Sigma B_T^{-1}),\quad\text{where}\;\; c_\eps\coloneqq\frac{\prn{1+\tau}\prn{2+\tau}}{\tau}
 \le\frac{100}{\eps^2}.
\end{equation}
With no initial terms and~$\Sigma=I_d$, this coincides with the bound in~\eqref{eq:survivor-simple} for surviving terms on the ball up to universal constants. For inputs~%
$\widetilde g_t=A^\top g'_t$ with~$\norm{g'_t}_2\le1$, we can take~%
$\Sigma=A^\top A$, even if it is singular; no bound on~$\norm{\widetilde g_t}_2$ is needed.

\subsection{One-Dimensional Bound for Initial Regularization and Trace}\label{app:initial-scales}
Substituting~\eqref{eq:survivor-bound} into~\eqref{eq:intermediate}
leaves $P_T^{\mathrm{init}}(v)+c_\eps\tr(\Sigma B_T^{-1})$ to bound.
Below we bound this sum from above coordinate by coordinate.
For~$\widetilde g_t=A^\top g'_t$ with~$\norm{g'_t}_2\le1$,
we have~$g'_t(g'_t)^\top\preceq I_d$. Taking~$\Sigma=A^\top A$ and using~%
$\norm{Ae_j}_2\le\ell_j$ gives
\begin{equation}\label{eq:envelope}
 \widetilde g_t\widetilde g_t^\top\preceq\Sigma\quad\text{and}\quad \Sigma_{jj}\le\ell_j^2
 \quad(j\in\set{1,\ldots,N}).
\end{equation}
Both~$P_T^{\mathrm{init}}(v)$ and~$\tr(\Sigma B_T^{-1})$ depend on which initial thresholds were crossed.

\subsubsection{Trace Bound from Surviving Thresholds}
In each coordinate, the thresholds of revoked initial terms form a prefix:
whenever a threshold is crossed, all active initial terms with smaller
thresholds are revoked as well. Let~$\widehat h_j$ be the smallest
threshold among surviving initial terms in coordinate~$j$,
or one if none survives. Summing the remaining geometric series in the definition of~$B_T$ gives
\begin{equation}\label{eq:compressed-initial}
 (B_T)_{jj}=\lambda\ell_j(2/\widehat h_j-1)
 \ge\lambda\ell_j/\widehat h_j.
\end{equation}
The prefix property also implies
\begin{equation}\label{eq:initial-threshold-prefix}
  \widehat h_j=\underline h_j+\sum_{\substack{0\le k<J_j,\\(j,k)\in\mathcal C_T^0}}h_{jk}.
\end{equation}
Since~$B_T$ is diagonal, each trace summand is
$\Sigma_{jj}/(B_T)_{jj}\le\ell_j\widehat h_j/\lambda$
by~$\Sigma_{jj}\le\ell_j^2$ and~\eqref{eq:compressed-initial}. 
Summing over~$j\in\set{1,\ldots,N}$ and using~\eqref{eq:initial-threshold-prefix} gives
\begin{equation}\label{eq:initial-trace}
 \tr(\Sigma B_T^{-1})
 \le\lambda^{-1}\brc*{L_{\underline h}+
 \sum_{(j,k)\in\mathcal C_T^0}\ell_jh_{jk}}.
\end{equation}
\subsubsection{Summation over Dyadic Thresholds}
We combine the expression of $P_T^{\mathrm{init}}(v)$ in~\eqref{eq:initial-contribution} with the trace bound~\eqref{eq:initial-trace} multiplied by~$c_\eps$ and sum over the dyadic thresholds in each coordinate.
Set~$c_1\coloneqq39/64$ and
$\lg_+x\coloneqq\max\set{0,\log_2x}$ for~$x>0$.
For~$(j,k)\in\mathcal C_T^0$, we have~$\abs{x_{jk}^{\mathrm{end}}}\ge h_{jk}$
and~$c_\eps/\lambda^2\le25/64$, so we have
$-\gamma_{jk}\prn{x_{jk}^{\mathrm{end}}}^2
+\prn{c_\eps/\lambda}\ell_jh_{jk}
\le-c_1\gamma_{jk}\prn{x_{jk}^{\mathrm{end}}}^2$.
The following lemma combines this inequality with the linear term~%
$2\gamma_{jk}x_{jk}^{\mathrm{end}}v_j$ in~\eqref{eq:initial-contribution}
and sums over thresholds.

\begin{lemma}\label{lem:initial-scales}
Let~$J\ge0$ be an integer, set~$h\coloneqq2^{-J}$ and~%
$h_k\coloneqq2^kh$ for~$k\in\set{0,\ldots,J-1}$, and fix~%
$v\in[-1,1]$ and~$\beta\in[0,25/64]$.
Let $\mathcal C\subseteq\set{0,\ldots,J-1}$ and let the numbers~%
$x_k\in\R$, for~$k\in\set{0,\ldots,J-1}$, satisfy
\begin{equation}\label{eq:scalar-endpoints}
 h_k\le\abs{x_k}<3h_k\quad(k\in\mathcal C)\quad\text{and}\quad \abs{x_k}<h_k\quad(k\in\set{0,\ldots,J-1}\setminus\mathcal C).
\end{equation}
Then, with~$a\coloneqq\abs{v}$, we have
\begin{equation}\label{eq:scalar-scale-bound}
 \sum_{k=0}^{J-1}\frac{2x_kv-x_k^2}{h_k}
 +\beta\sum_{k\in\mathcal C}h_k
 \le6a\lg_+(a/h)+\prn{6+2/c_1}a.
\end{equation}
At~$a=0$, the term~$6a\lg_+(a/h)$ is defined as zero; when~$J=0$
all sums are empty.
\end{lemma}
\begin{proof}
Fix~$k\in\mathcal C$ and put~$\vartheta\coloneqq x_k/h_k$.
By~\eqref{eq:scalar-endpoints}, we have~$1\le\abs{\vartheta}<3$.
Since $\beta\le25/64=1-c_1$, the quadratic and trace terms satisfy
\begin{equation}\label{eq:scalar-negative-square}
 -\vartheta^2h_k+\beta h_k\le-c_1\vartheta^2h_k.
\end{equation}
Adding~$2\vartheta v$ to~\eqref{eq:scalar-negative-square} gives the first inequality below. For the second, $\abs{\vartheta}<3$ gives the upper bound~$6a$, and completing the square gives~$a^2/\prn{c_1h_k}$. Thus we have
\begin{equation}\label{eq:initial-one-scale}
 \frac{2x_kv-x_k^2}{h_k}+\beta h_k
 \le2\vartheta v-c_1\vartheta^2h_k
 \le\min\set*{6a,\frac{a^2}{c_1h_k}}.
\end{equation}
For $k\in\set{0,\ldots,J-1}\setminus\mathcal C$, the trace term
is absent. Equation~\eqref{eq:scalar-endpoints} gives~%
$\abs{x_k/h_k}<1$, so the same two bounds apply to~%
$\prn{2x_kv-x_k^2}/h_k$ alone.

For~$a>0$, there are at most~$1+\lg_+(a/h)$ thresholds with~%
$h_k<a$. Use the bound~$6a$ for these thresholds. For the remaining thresholds, the values~$a^2/\prn{c_1h_k}$
halve at each successive scale and their first value is at most~%
$a/c_1$. Thus the sum over thresholds satisfies
\begin{equation}\label{eq:geometric-sum}
 \sum_{k=0}^{J-1}\min\set*{6a,\frac{a^2}{c_1h_k}}
 \le6a\lg_+(a/h)+\prn{6+2/c_1}a.
\end{equation}
Summing the upper bounds for the individual thresholds and using~\eqref{eq:geometric-sum}
proves~\eqref{eq:scalar-scale-bound}. If~$a=0$, both bounds in the minimum are zero, so the left-hand side is nonpositive. If~$J=0$, both sums in~%
\eqref{eq:scalar-scale-bound} vanish. Therefore, the bound holds in all cases.
\end{proof}

\subsection{Regret Bound in Terms of the Width}\label{app:coordinate-sum}
We sum the one-dimensional bounds to obtain the regret guarantee in terms of~%
$\Phi_{\ell,\underline h}$, then use its relation to the width.
\subsubsection{Summing Coordinate Bounds}
To bound $P_T^{\mathrm{init}}(v)+c_\eps\tr(\Sigma B_T^{-1})$ for~$v\in[-1,1]^N$, we apply \cref{lem:initial-scales} to each coordinate~%
$j\in\set{1,\ldots,N}$ with~$J=J_j$, $h=\underline h_j$, $v=v_j$,
$x_k=x_{jk}^{\mathrm{end}}$ for $k\in\set{0,\ldots,J_j-1}$, and
\begin{equation}
  \mathcal C=\Set{k\in\set{0,\ldots,J_j-1}}{(j,k)\in\mathcal C_T^0}.
\end{equation}
For a surviving term,~$x_k=(z_{T+1})_j$ satisfies~$\abs{x_k}<h_k$ by
\cref{item:move-revoke-thresholds}; for a revoked term,~$x_k$ is the~$j$th coordinate
of its revocation point and satisfies~$h_k\le\abs{x_k}<3h_k$ by \cref{item:move-revoke-thresholds}.
These are the conditions on~$x_k$ in \cref{lem:initial-scales}. Set~$\beta\coloneqq c_\eps/\lambda^2$;
equation~\eqref{eq:survivor-bound} and~$\lambda=16/\eps$ give~%
$0<\beta\le25/64$.
Multiplying~\eqref{eq:scalar-scale-bound} by~$\lambda\ell_j$,
summing from~$j=1$ to~$N$, and using~%
\eqref{eq:initial-contribution} and~\eqref{eq:initial-trace} yields
\begin{equation}\label{eq:initial-joint-bound}
 P_T^{\mathrm{init}}(v)+c_\eps\tr(\Sigma B_T^{-1})
 \le\frac{c_\eps}{\lambda}L_{\underline h}
 +\lambda\sum_{j=1}^N\ell_j
 \brc*{6\abs{v_j}\lg_+(\abs{v_j}/\underline h_j)
 +\prn{6+2/c_1}\abs{v_j}}.
\end{equation}

In~\eqref{eq:intermediate}, we use~\eqref{eq:survivor-bound} for the surviving terms
and~\eqref{eq:initial-joint-bound} for
$P_T^{\mathrm{init}}(v)+c_\eps\tr(\Sigma B_T^{-1})$.
Since~$B=\lambda\diag(\ell_1,\ldots,\ell_N)$ and~$v\in[-1,1]^N$, we have
$\norm{v}_B^2\le\lambda\sum_{j=1}^N\ell_j\abs{v_j}$.
Using also~$E_T^{\mathrm{skip}}<L_{\underline h}/8$ from~\eqref{eq:general-skip}, we obtain coefficients of at most~$51/\prn{8\eps}$, $96/\prn{\eps\log2}$, and~$16(7+128/39)/\eps$ for~$L_{\underline h}$, the logarithmic term in~\eqref{eq:geometry}, and its linear term, respectively. Since each coefficient is below~$192/\eps$, the finite initialization~\eqref{eq:initial-quadratics} gives
\begin{equation}\label{eq:internal-bound}
 2\sum_{t=1}^T\widetilde r_t-\eps\sum_{t=1}^T\abs{\widetilde r_t}
 +\norm{z_{T+1}-v}_B^2
 \le\frac{192}{\eps}\Phi_{\ell,\underline h}(v).
\end{equation}

\subsubsection{Choosing an Admissible Description}\label{app:width-bound}
By \cref{prop:width-equivalence}, we can choose an admissible description~$\mathcal D$ depending only on~$\mathcal K$ such that $C_{\mathcal D}(\mathcal K)\asymp\width(\mathcal K)$.
Minimizing the coefficient cost in~%
\eqref{eq:internal-bound} over each comparator's representations gives
the following guarantee.

\begin{restatable}{lemma}{geometrybound}\label{lem:geometry}
Fix~$\eps\in(0,1]$ and an admissible description~$\mathcal D$ with~$C_{\mathcal D}(\mathcal K)\le20\width(\mathcal K)$.
Run \cref{alg:core} with the coordinate initialization and addition thresholds in \cref{app:general-core}.
Every~$u\in\mathcal K$ has a representation~$u=w_\circ+Av$ with~$v\in[-1,1]^N$, fixed independently
of the input sequence, such that for every sequence~%
$\widetilde g_t=A^\top g'_t$ with~$\norm{g'_t}_2\le1$, we have
\begin{equation}\label{eq:geometry-internal}
 2\sum_{t=1}^T\inpr{\widetilde g_t,z_t-v}
 -\eps\sum_{t=1}^T\abs{\inpr{\widetilde g_t,z_t-v}}
 \le\frac{3840}{\eps}\width(\mathcal K)
 \quad(T\ge1).
\end{equation}
\end{restatable}
\begin{proof}
For each~$u$, choose~$v\in[-1,1]^N$ minimizing~$\Phi_{\ell,\underline h}(v)$ subject to~$u=w_\circ+Av$.
This minimum is at most~$C_{\mathcal D}(\mathcal K)$ by~\eqref{eq:representation-cost}.
Use this~$v$ in~\eqref{eq:internal-bound} with~%
$\Sigma=A^\top A$ and discard~$\norm{z_{T+1}-v}_B^2\ge0$ from the left-hand side.
This proves~\eqref{eq:geometry-internal}.
\end{proof}

\subsection{\texorpdfstring{Simplified Initialization for $p\ge2$}{Simplified Initialization for p >= 2}}\label{app:fixed-initialization}\label{app:ball}
For~$\ell_p$ balls with~$p\ge2$, we can omit the revocable initial terms and retain the same width dependence in the regret guarantee.
This simplification also applies after scaling along coordinate axes, isometric embedding, and translation.
As shown below, we initialize with a single fixed regularization term and use the same rules for additions and revocations after feedback.
We first specialize~\eqref{eq:full-lifetime} and~\eqref{eq:survivor-bound} to this initialization, then choose the fixed matrix to achieve the same width dependence.

\begin{lemma}\label{lem:fixed-initialization}
Fix~$\eps\in(0,1]$, an integer~$N\ge1$, a positive definite matrix~$B\in\R^{N\times N}$, a positive semidefinite matrix~$\Sigma\in\R^{N\times N}$, and thresholds~$\zeta_t\ge0$ for~$t\ge1$.
Run \cref{alg:core} with~$z_1=0$, $\Psi_1(z)=\norm{z}_B^2/2$, $H_1=B$, and~$\mathcal Q_1=\varnothing$.
Suppose the inputs~$\widetilde g_t\in\R^N$ satisfy~$\widetilde g_t\widetilde g_t^\top\preceq\Sigma$ for every~$t$.
For any~$v\in\R^N$, write~$\widetilde r_t\coloneqq\inpr{\widetilde g_t,z_t-v}$.
Then, for every~$T\ge1$, we have
\begin{equation}\label{eq:fixed-distance}
2\sum_{t=1}^T\widetilde r_t-\eps\sum_{t=1}^T\abs{\widetilde r_t}
+\norm{z_{T+1}-v}_B^2
\le\norm{v}_B^2+\frac{100}{\eps^2}\tr(\Sigma B^{-1})+\sum_{t=1}^T\zeta_t.
\end{equation}
\end{lemma}
\begin{proof}
Since there are no revocable initial terms, we have~$P_T^{\mathrm{init}}(v)=0$ and~$B_T=B$.
Substituting~\eqref{eq:survivor-bound} into~\eqref{eq:full-lifetime} therefore bounds the left-hand side of~\eqref{eq:fixed-distance} by
$\norm{v}_B^2+100\tr(\Sigma B^{-1})/\eps^2+E_T^{\mathrm{skip}}$.
On every round with an omitted addition, we have~$s_t^2\le\zeta_t$, so~$E_T^{\mathrm{skip}}\le\sum_{t=1}^T\zeta_t$.
This proves~\eqref{eq:fixed-distance}.
\end{proof}

\paragraph{Scaling along the Coordinate Axes.}
We then explain specifications for an $\ell_p$ ball scaled along its coordinate axes and then mapped isometrically and translated. We choose the fixed matrix and addition thresholds to bound the left-hand side of~\eqref{eq:fixed-distance} by a constant multiple of~$\width(\mathcal K)/\eps$.
Fix~$\eps\in(0,1]$, $2\le p\le\infty$, $1\le r\le d$, $w_\circ\in\R^d$, $U\in\R^{d\times r}$ with~$U^\top U=I_r$, and~$a\in(0,\infty)^r$.
Use~$N=r$ coordinates, write~$A\coloneqq U\diag(a_i)$ and~$\mathcal K\coloneqq w_\circ+A\B_p^r$, and let~$1/p+1/q=1$.
Let~$S\coloneqq\norm{a}_q$ and set
\begin{equation}\label{eq:fixed-initial-matrix}
B\coloneqq\frac{10}{\eps}\diag\prn{S^{q-1}a_i^{2-q}}
\quad
\text{and}
\quad \Psi_1(z)\coloneqq\frac12 z^\top Bz.
\end{equation}
Take~$z_1=0$, $H_1=B$, and~$\mathcal Q_1=\varnothing$.
After each update, add a regularization term only if~$s_t^2>\zeta_t$, where~$\zeta_t\coloneqq S(t+1)^{-4}$.
For this choice of~$B$, we claim
\begin{equation}\label{eq:fixed-initial-balance}
\sup_{v\in\B_p^r}v^\top Bv=\frac{10S}{\eps}
\quad\text{and}\quad
\tr(A^\top AB^{-1})=\frac{\eps S}{10}.
\end{equation}
For~$2<p<\infty$, we bound the supremum in~\eqref{eq:fixed-initial-balance} using H\"older's inequality with conjugate exponents~$p/\prn{p-2}$ and~$p/2$.
Since~$\prn{2-q}p/\prn{p-2}=q$, we have, for every~$v\in\B_p^r$,
\begin{equation}\label{eq:fixed-initial-holder}
\sum_{i=1}^r a_i^{2-q}v_i^2
\le\prn*{\sum_{i=1}^r a_i^q}^{(p-2)/p}\norm{v}_p^2
=S^{2-q}\norm{v}_p^2.
\end{equation}
Taking~$v_i=\prn{a_i/S}^{q/p}$ for every~$i\in\set{1,\ldots,r}$ gives~$\norm{v}_p=1$ and attains equality in~\eqref{eq:fixed-initial-holder}. Multiplication by~$10S^{q-1}/\eps$ then gives the first equality in~\eqref{eq:fixed-initial-balance}.
At~$p=2$, we have~$B=\prn{10S/\eps}I_r$, and at~$p=\infty$, we have~$B=\prn{10/\eps}\diag(a_i)$, so the same equality holds at both endpoints.
For the trace, $A^\top A=\diag(a_i^2)$ gives
\begin{equation}\label{eq:fixed-initial-trace}
\tr(A^\top AB^{-1})
=\frac{\eps}{10S^{q-1}}\sum_{i=1}^r a_i^q
=\frac{\eps S}{10}.
\end{equation}
For inputs~$\widetilde g_t=A^\top g'_t$ with~$\norm{g'_t}_2\le1$, we have~$\widetilde g_t\widetilde g_t^\top\preceq A^\top A$.
Write~$\widetilde r_t(v)\coloneqq\inpr{\widetilde g_t,z_t-v}$.
Applying~\cref{lem:fixed-initialization} with~$\Sigma=A^\top A$, substituting~\eqref{eq:fixed-initial-balance}, and using~$\sum_{t=1}^{\infty}(t+1)^{-4}<1/8$ gives, for every~$v\in\B_p^r$ and~$T\ge1$,
\begin{equation}\label{eq:fixed-norm-ball-regret}
2\sum_{t=1}^T\widetilde r_t(v)-\eps\sum_{t=1}^T\abs{\widetilde r_t(v)}
+\norm{z_{T+1}-v}_B^2
\le\frac{20S}{\eps}+\frac S8\le\frac{21S}{\eps}.
\end{equation}
To relate~$S$ to the width, take~$\xi\in\R^r$ with the independent symmetric Laplace coordinates from~\cref{app:width-lower}.
By~\eqref{eq:laplace-absolute-moment} and~\cref{item:width-transform}, the vector~$U\xi/4$ satisfies the width constraints in~$\R^d$.
The support function and Jensen's inequality give
\begin{equation}\label{eq:fixed-width-comparison}
\width(\mathcal K)\ge\E\sigma_{\mathcal K}(U\xi/4)
=\frac14\E\norm{\diag(a_i)\xi}_q
\ge\frac14\norm{a}_q=\frac S4.
\end{equation}
Thus the left-hand side of~\eqref{eq:fixed-norm-ball-regret} is at most~$84\width(\mathcal K)/\eps$.
For~$\ell_p$ balls with~$p\ge2$, ellipsoids, and boxes, the procedure in~\cref{app:initialization-arithmetic} computes a matrix~$\widehat B$ satisfying~$B\preceq\widehat B\preceq3B$.
With the addition thresholds specified there, \eqref{eq:rounded-fixed-guarantee} bounds the right-hand side of~\eqref{eq:fixed-distance} for~$\widehat B$ by~$41S/\eps$. Equation~\eqref{eq:fixed-width-comparison} therefore gives a bound of order~$\width(\mathcal K)/\eps$.

\paragraph{Implementation on the Unit Euclidean Ball.}\phantomsection\label{app:ball-implementation}
We apply~\eqref{eq:fixed-distance} to bound the regret of \cref{alg:ball-oco} with a positive addition threshold.
Let~$k\coloneqq\ceil{\sqrt d}$ and set
\begin{equation}\label{eq:ball-base}
 B\coloneqq\lambda I_d
 \quad
 \text{and}
 \quad\lambda\coloneqq10k/\eps.
\end{equation}
Replace the addition condition~$s_t^2>0$ in \cref{alg:ball-oco} by~$s_t^2>\zeta_t$, where~$\zeta_t\coloneqq\prn{\eps/k}\prn{t+1}^{-4}$.
For inputs with~$\norm{\widetilde g_t}_2\le1$, take~$\Sigma=I_d$ in~\eqref{eq:fixed-distance}.
For every~$u\in\B_2^d$, write~$\widetilde r_t(u)\coloneqq\inpr{\widetilde g_t,z_t-u}$.
Using~$d\le k^2$ and~$\sum_{t=1}^{\infty}(t+1)^{-4}<1/8$ in~\eqref{eq:fixed-distance} gives
\begin{equation}\label{eq:ball-distance}
2\sum_{t=1}^T\widetilde r_t(u)-\eps\sum_{t=1}^T\abs{\widetilde r_t(u)}+\norm{z_{T+1}-u}_B^2
\le\lambda\norm{u}_2^2+\frac{100}{\eps^2}\tr(B^{-1})+\frac{\eps}{8k}
\le\frac{20k}{\eps}+\frac{\eps}{8k}\le\frac{21k}{\eps}.
\end{equation}
Discarding the terminal-distance term yields
\begin{equation}\label{eq:ball-signed}
2\sum_{t=1}^T\widetilde r_t(u)-\eps\sum_{t=1}^T\abs{\widetilde r_t(u)}
\le\frac{20\ceil*{\sqrt d}}{\eps}+\frac{\eps}{8\ceil*{\sqrt d}}
\le\frac{21\ceil*{\sqrt d}}{\eps}
\qquad(u\in\B_2^d,\ T\ge1).
\end{equation}
Applying the gradient correction~\eqref{eq:ball-correction} as in the proof of \cref{thm:ball-oco} to~\eqref{eq:ball-signed} with~$\norm{g_t}_2\le1$ gives~\eqref{eq:ball-played-signed} with~$21$ in place of~$20$.

\paragraph{\texorpdfstring{Obstacles for $p<2$.}{Obstacles for p < 2.}}
For~$1\le p<2$, the H\"older step in~\eqref{eq:fixed-initial-holder} fails because its exponents~$p/2$ and~$p/\prn{p-2}$ do not satisfy the hypotheses of H\"older's inequality.
\Cref{lem:fixed-initialization} still applies, but even a different fixed matrix cannot give the width order through~\eqref{eq:fixed-distance}.
Consider~$\B_p^N$ with~$\Sigma=I_N$. For every~$B\succ0$, we have
\begin{equation}\label{eq:fixed-matrix-limitation}
\sup_{v\in\B_p^N}\norm{v}_B^2+\frac{100}{\eps^2}\tr(B^{-1})
\ge\frac{\tr B}{N}+\frac{100N^2}{\eps^2\tr B}
\ge\frac{20\sqrt N}{\eps}.
\end{equation}
The first inequality uses~$e_j\in\B_p^N$ for every~$j\in\set{1,\ldots,N}$ and~$\tr B\,\tr(B^{-1})\ge N^2$ from the Cauchy--Schwarz inequality; the second is the arithmetic--geometric mean inequality.
For each fixed~$p<2$, \eqref{eq:unit-lp-width} gives~$\width(\B_p^N)=o\prn{\sqrt N}$ as~$N\to\infty$.
Thus~\eqref{eq:fixed-matrix-limitation} shows that optimizing~$B$ in~\eqref{eq:fixed-distance} cannot recover the width order attained by the multiscale initialization.

\subsection{Regret after Projection onto the Domain}\label{app:projection}
We transfer the unconstrained guarantee to the regret of predictions in~$\mathcal K$ by projection and gradient correction.
Given the unconstrained iterate~$z_t\in\R^N$, form the unprojected point
$\widetilde w_t\coloneqq w_\circ+Az_t\in\R^d$, play~%
$w_t\coloneqq\Pi_{\mathcal K}(\widetilde w_t)\in\mathcal K$, and observe~$g_t$ there.
We correct~$g_t$ for the projection using the reduction of
\citet[Theorem~2]{cutkosky2020parameterfree}, building on
\citet{cutkosky2018blackbox}.\footnote{
  \citet[Section~3.2, Lemma~3]{wang2026lightons} use this conversion in LightONS to defer expensive Mahalanobis projections.
} 
Here each prediction uses Euclidean projection. Put~$n_t\coloneqq\widetilde w_t-w_t$ and define

\begin{equation}\label{eq:correction}
 \bar g_t\coloneqq
 \begin{cases}
 g_t,&n_t=0,\\
 g_t-\dfrac{\min\set{0,\inpr{g_t,n_t}}}{\norm{n_t}_2^2}n_t,&n_t\ne0,
 \end{cases}
 \quad\text{and}\quad\widetilde g_t\coloneqq A^\top\bar g_t/G.
\end{equation}
The projection correction gives~$\bar g_t\in\R^d$. We then apply~$A^\top/G$ to obtain the input~$\widetilde g_t\in\R^N$ supplied to \cref{alg:core}.
If~$\inpr{g_t,n_t}<0$, the correction removes the component of~$g_t$
along~$n_t$. The resulting linearized excess loss at~$\widetilde w_t$ is at least
that at~$w_t$ for every~$u\in\mathcal K$, as shown below.

\Cref{alg:oco} combines this projection rule with the unconstrained update.
The domain representation and initialization are chosen once, before observing feedback.
For a representation~$\mathcal K=w_\circ+U\diag(a_i)\B_p^r$ with~$U^\top U=I_r$, $a_i>0$ for~$i\in\set{1,\ldots,r}$, and~$p\ge2$, we use the simplified initialization in \cref{app:fixed-initialization}: a single fixed regularization term and no revocable initial terms, with the fixed matrix and addition thresholds specified there.
For~$\ell_p$ balls with~$p<2$ and for the probability simplex~$\triangle_d$, use the dyadic thresholds in \cref{app:scaled-norm-balls,app:simplex-extension} and initialize the coordinate terms as in \cref{app:general-core}.
For an arbitrary domain, choose an admissible description~$\mathcal D$ with~$C_{\mathcal D}(\mathcal K)\le20\,\width(\mathcal K)$ (\cref{prop:width-equivalence}) and initialize as in \cref{app:general-core}.
For the canonical domains in \cref{cor:canonical-width}, the procedure in \cref{app:initialization-arithmetic} computes the fixed matrix for~$p\ge2$ and the dyadic coordinate thresholds for~$p<2$ and the simplex.

\begin{algorithm2e}
\caption{Main Procedure on General Domains}\label{alg:oco}
\KwIn{Domain~$\mathcal K$ from \cref{sec:protocol}, norm bound~$G>0$, parameter~$\eps\in(0,1]$.}
\uIf{a representation from \cref{app:fixed-initialization} with~$p\ge2$ is supplied}{
Initialize with the single fixed regularization term and no revocable initial terms as in \cref{app:fixed-initialization}, with the implementation in \cref{app:initialization-arithmetic} for canonical domains\;}
\uElseIf{$\mathcal K$ is an $\ell_p$ ball with~$p<2$ or $\mathcal K=\triangle_d$}{
Choose~$\mathcal D$ as in \cref{app:scaled-norm-balls,app:simplex-extension} and initialize as in \cref{app:general-core}\;}
\Else{
Choose~$\mathcal D$ as in \cref{prop:width-equivalence} and initialize as in \cref{app:general-core}\;}
\For{$t=1,2,\ldots$}{
Set $w_t\leftarrow\Pi_{\mathcal K}(w_\circ+Az_t)$, play~$w_t$, and receive~$g_t$\;
Compute~$\widetilde g_t$ by~\eqref{eq:correction}\;
Run one round of~\cref{alg:core} with this input, $\eps$, and the
addition threshold~$\zeta_t$ specified by the initialization\;}
\end{algorithm2e}

\begin{restatable}{lemma}{projectionbound}\label{lem:projection}
For every~$u\in\mathcal K$, the corrected vector satisfies
\begin{equation}\label{eq:projection-facts}
 \norm{\bar g_t}_2\le\norm{g_t}_2\le G\quad\text{and}\quad \inpr{\bar g_t,\widetilde w_t-u}\ge\inpr{g_t,w_t-u}.
\end{equation}
\end{restatable}
\begin{proof}
Recall that~$n_t=\widetilde w_t-w_t$ and~$w_t=\Pi_{\mathcal K}(\widetilde w_t)$.
Projection optimality gives~$\inpr{n_t,w_t-u}\ge0$.
If~$\inpr{g_t,n_t}\ge0$, we have~$\bar g_t=g_t$ and
$\inpr{\bar g_t,\widetilde w_t-u}-\inpr{g_t,w_t-u}=\inpr{g_t,n_t}\ge0$.
Otherwise~$n_t\ne0$ holds, and~\eqref{eq:correction} removes the component of~$g_t$
along~$n_t$, so we have~$\norm{\bar g_t}_2\le\norm{g_t}_2$ and~$\inpr{\bar g_t,n_t}=0$.
Consequently, we have
$\inpr{\bar g_t,\widetilde w_t-u}-\inpr{g_t,w_t-u}
=-\inpr{g_t,n_t}\inpr{n_t,w_t-u}/\norm{n_t}_2^2\ge0$.
The first case includes~$n_t=0$, completing the proof.
\end{proof}

Now we are ready to prove \cref{thm:main}.

\begin{proof}[\cref{thm:main}]
By~\eqref{eq:correction} and \cref{lem:projection}, the inputs have the form~$\widetilde g_t=A^\top(\bar g_t/G)$ with $\norm{\bar g_t/G}_2\le1$, as required by the unconstrained arguments.
For each initialization and comparator~$u\in\mathcal K$, we choose a representation~$u=w_\circ+Av$ satisfying
\begin{equation}\label{eq:initialization-excess-bound}
\sum_{t=1}^T\prn*{2\inpr{\widetilde g_t,z_t-v}-\eps\abs{\inpr{\widetilde g_t,z_t-v}}}
\lesssim\frac{\width(\mathcal K)}\eps.
\end{equation}
For the fixed regularizer on~$\mathcal K=w_\circ+A\B_p^r$ with~$p\ge2$ and nonzero, mutually orthogonal columns of~$A$, use the coefficient vector~$v\in\B_p^r$ representing~$u$.
Equations~\eqref{eq:fixed-norm-ball-regret} and~\eqref{eq:fixed-width-comparison} give~\eqref{eq:initialization-excess-bound}.
For the fixed-matrix initializations of~$\ell_p$ balls with~$p\ge2$, ellipsoids, and boxes, the computed matrix~$\widehat B$ in \cref{app:initialization-arithmetic} satisfies~$B\preceq\widehat B\preceq3B$.
With the addition thresholds specified there, \eqref{eq:rounded-fixed-guarantee} bounds the right-hand side of~\eqref{eq:fixed-distance}, with~$B=\widehat B$, by~$41S/\eps$.
Together with~\eqref{eq:fixed-width-comparison}, this gives~\eqref{eq:initialization-excess-bound}.
For an~$\ell_p$ ball with~$p<2$ or for~$\triangle_d$, \cref{app:scaled-norm-balls,app:simplex-extension} give~$\Phi_{\ell,\underline h}(v)\lesssim\width(\mathcal K)$, so~\eqref{eq:internal-bound} implies~\eqref{eq:initialization-excess-bound}.
For the initialization on an arbitrary domain, choose~$v$ as in \cref{lem:geometry} and use~\eqref{eq:geometry-internal}.
Each choice depends on~$u$ and the initialization and is independent of the input sequence.

Equation~\eqref{eq:correction} and \cref{lem:projection} give
\begin{equation}\label{eq:general-excess-transfer}
 \inpr{\widetilde g_t,z_t-v}
 =\frac{\inpr{\bar g_t,\widetilde w_t-u}}G
 \ge\frac{r_t(u)}G.
\end{equation}
The function~$r\mapsto2r-\eps\abs{r}$ is increasing and positively homogeneous.
Applying it to~\eqref{eq:general-excess-transfer}, summing over~$t$, and using~\eqref{eq:initialization-excess-bound} proves~\eqref{eq:main-signed} after multiplication by~$G$.
\end{proof}

The computational cost of \cref{alg:oco} is analyzed in \cref{app:computation}.

\section{Adaptation to Unknown Shifted Comparator Loss}\label{app:adapt}
We combine base learners with different fixed values of~$\eps$ to obtain a regret bound in terms of~$N_T(u)$ without knowing~$N_T(u)$ in advance.
\restate{\adaptivewidthbound}
We first bound the linearized regret for inputs with~$\norm{g_t}_2\le G$ when \cref{alg:adaptive} uses a supplied~$M\ge G\operatorname{diam}_2\mathcal K$. For~$W\ge\width(\mathcal K)$, the choice~$M\coloneqq2GW/\log2$ satisfies this condition by~\eqref{eq:width-radius} and gives \cref{thm:adaptive-known-width}.
All base learners share the coordinate representation and initialization rule of \cref{alg:oco}. Each uses its own parameter~$\eps$ to form its initial matrix and run its updates. For the fixed initialization in \cref{app:fixed-initialization}, learners on~$\ell_p$ balls with~$p\ge2$, ellipsoids, or boxes may use the computed matrix and addition scale in \cref{app:initialization-arithmetic}. For the multiscale initialization, they share the admissible description~$\mathcal D$.

\subsection{Recursive Mixture of Base Learners}\label{app:master}
We mix the predictions of fixed-parameter learners and scale their feedback to compare the mixture with any comparator~$u$ through each learner (see \cref{app:scalar}).
\subsubsection{Base Learners and Scaled Feedback}
For each integer~$i\ge0$, start a copy of \cref{alg:oco} at global round~$t_i\coloneqq2^i$ with parameter~$\eps_i\coloneqq2^{-i}/8$ and input bound~$2G$. At global round~$t\ge t_i$, the awake learner~$i$ predicts~$w_t^{(i)}\in\mathcal K$ and runs its update at local round~$t-t_i+1$. The factor~$\prn{t+1}^{-4}$ in its addition threshold becomes~$\prn{t-t_i+2}^{-4}$ with any of the three scales, $S$ in \cref{app:fixed-initialization}, $\widehat S$ in~\eqref{eq:rounded-addition-scale}, and~$L_{\underline h}$ in~\eqref{eq:general-skip}.

The mixture predicts a convex combination~$w_t$ of the awake learners' predictions (see \cref{app:prediction-feedback}). Write~$r_{t,i}\coloneqq\inpr{g_t,w_t-w_t^{(i)}}$ for its linearized loss difference from each awake learner~$i$. For a loss difference~$a\in\R$ and~$0<\theta<1$, define the feedback multiplier by
\begin{equation}\label{eq:scalar-correction}
\operatorname{scale}_\theta(a)\coloneqq\begin{cases}
\prn{1-\theta}/\prn{1+\theta},&a>0,\\
1,&a=0,\\
\prn{1+\theta}/\prn{1-\theta},&a<0.
\end{cases}
\end{equation}
After observing~$g_t$ at~$w_t$, we supply~$g_t^{(i)}\coloneqq\operatorname{scale}_{\eps_i/2}(r_{t,i})g_t$ to each awake learner~$i$.
Since $0<\operatorname{scale}_{\eps_i/2}(r_{t,i})\le17/15<2$, this input has norm at most~$2G$.
At that local round, \cref{alg:oco} applies~\eqref{eq:correction} to~$g_t^{(i)}$ using learner~$i$'s projection and norm bound~$2G$, then supplies the resulting vector in~$\R^N$ to \cref{alg:core}.

\subsubsection{Recursive Mixture}\label{app:recursive-mixture}
\paragraph{Aggregating a Loss Vector.}\leavevmode
For an integer~$K\ge2$, let~$\widetilde L=\prn{\widetilde L_0,\ldots,\widetilde L_{K-1}}\in\R^K$ be a loss vector. We map~$\widetilde L$ to mixture weights by a recursion independent of the round, with fixed learning rates~$\eta_k\coloneqq\eps_k/\prn{128M}$ for~$k\ge0$. The potential~$V_k(\widetilde L)$ aggregates the losses~$\widetilde L_k,\ldots,\widetilde L_{K-1}$ for~$0\le k<K$.
Set~$V_{K-1}(\widetilde L)\coloneqq\widetilde L_{K-1}$ and, for~$k=K-2,\ldots,0$, define
\begin{equation}\label{eq:mix-recursion}
\begin{aligned}
V_k(\widetilde L)&\coloneqq-\eta_k^{-1}\log\frac{\exp\prn{-\eta_k\widetilde L_k}+\exp\prn{-\eta_kV_{k+1}(\widetilde L)}}{2}\qquad\text{and}\\
p_k^{\mathrm L}(\widetilde L)&\coloneqq\frac{\exp\prn{-\eta_k\widetilde L_k}}{\exp\prn{-\eta_k\widetilde L_k}+\exp\prn{-\eta_kV_{k+1}(\widetilde L)}}.
\end{aligned}
\end{equation}
At split~$k$, the mass assigned to coordinates~$k,\ldots,K-1$ is divided in proportions~$p_k^{\mathrm L}(\widetilde L)$ and~$1-p_k^{\mathrm L}(\widetilde L)$ between coordinate~$k$ and the remaining coordinates~$k+1,\ldots,K-1$.
Thus~$p_k^{\mathrm L}(\widetilde L)>1/2$ holds exactly when~$\widetilde L_k<V_{k+1}(\widetilde L)$.
Let~$P_k(\widetilde L)$ denote the mass assigned to coordinates~$k,\ldots,K-1$ for~$0\le k\le K-1$. Starting from~$P_0(\widetilde L)\coloneqq1$, define the coordinate weights recursively by
\begin{equation}\label{eq:mix-leaf-weights}
p_k(\widetilde L)\coloneqq P_k(\widetilde L)p_k^{\mathrm L}(\widetilde L)\quad\text{and}\quad
P_{k+1}(\widetilde L)\coloneqq P_k(\widetilde L)\prn{1-p_k^{\mathrm L}(\widetilde L)}\quad(0\le k\le K-2),
\end{equation}
and $p_{K-1}(\widetilde L)\coloneqq P_{K-1}(\widetilde L)$.
The weights define~$p(\widetilde L)\coloneqq\prn{p_0(\widetilde L),\ldots,p_{K-1}(\widetilde L)}\in\triangle_K$.

\paragraph{\texorpdfstring{Weights at Round~$t$.}{Weights at Round t.}}\leavevmode
For~$r\in\R$ and~$\theta\ge0$, let~$\Lambda_\theta(r)\coloneqq r-\theta\abs{r}$. At round~$s\ge t_i$, learner~$i$ has corrected loss~$-\Lambda_{\eps_i/2}(r_{s,i})$ and cumulative corrected loss~$\widetilde L_{s,i}\coloneqq-\sum_{h=t_i}^s\Lambda_{\eps_i/2}(r_{h,i})$. Set~$\widetilde L_{s,i}\coloneqq0$ for~$0\le s<t_i$.
At round~$t$, let~$i_{\max}(t)\coloneqq\floor{\log_2t}$ and~$K\coloneqq i_{\max}(t)+2$. Applying~$p$ to the cumulative corrected losses through round~$t-1$ of learners awake at round~$t$, with a final zero for all learners yet to start, gives the exact weights
\begin{equation}\label{eq:mix-round-weights}
p_t\coloneqq p\bigl(\prn{\widetilde L_{t-1,0},\ldots,\widetilde L_{t-1,i_{\max}(t)},0}\bigr)\in\triangle_K.
\end{equation}

\subsubsection{Prediction and Feedback}\label{app:prediction-feedback}
To implement the mixture using arithmetic operations and comparisons, we allow the use of approximate weights $\widehat p_t=\prn{\widehat p_{t,0},\ldots,\widehat p_{t,K-1}}\in\triangle_K$ satisfying
\begin{equation}\label{eq:mix-approximation}
\sum_{i=0}^{i_{\max}(t)}\widehat p_{t,i}>0\quad\text{and}\quad
\norm{\widehat p_t-p_t}_1\le\delta_t^{\mathrm{mix}}\coloneqq(t+1)^{-2}.
\end{equation}
\Cref{app:adaptive-computation} gives an arithmetic procedure satisfying these conditions.

We normalize the approximate weights and combine the awake learners' predictions by
\begin{equation}\label{eq:master-prediction}
\pi_{t,i}\coloneqq\frac{\widehat p_{t,i}}{\sum_{j=0}^{i_{\max}(t)}\widehat p_{t,j}}\quad(0\le i\le i_{\max}(t))\quad\text{and}\quad w_t\coloneqq\sum_{i=0}^{i_{\max}(t)}\pi_{t,i}w_t^{(i)}.
\end{equation}
Recall $r_{t,i}=\inpr{g_t,w_t-w_t^{(i)}}$ for~$0\le i\le i_{\max}(t)$.
After observing~$g_t$, update the cumulative corrected losses of the awake learners by
\begin{equation}\label{eq:master-update}
\widetilde L_{t,i}\leftarrow\widetilde L_{t-1,i}-\Lambda_{\eps_i/2}(r_{t,i})\quad(0\le i\le i_{\max}(t)).
\end{equation}

To bound the mixing error, we average the corrected losses with weights~$\widehat p_t$.
For each learner~$i>i_{\max}(t)$, set~$w_t^{(i)}\coloneqq w_t$ and~$r_{t,i}\coloneqq0$.
By~\eqref{eq:master-prediction}, the awake learners' predictions have weighted sum~$\prn{1-\widehat p_{t,K-1}}w_t$. Adding~$\widehat p_{t,K-1}w_t$ and using the definition of~$r_{t,i}$, we obtain
\begin{equation}\label{eq:mix-zero-mean}
\sum_{i=0}^{K-1}\widehat p_{t,i}w_t^{(i)}=w_t
\qquad\text{and}\qquad
\sum_{i=0}^{K-1}\widehat p_{t,i}r_{t,i}=0.
\end{equation}
Canceling the linear terms by~\eqref{eq:mix-zero-mean} gives~$-\sum_i\widehat p_{t,i}\Lambda_{\eps_i/2}(r_{t,i})=\sum_i\widehat p_{t,i}\prn{\eps_i/2}\abs{r_{t,i}}$.
We use this sum in \cref{app:mixing-analysis} to offset the mixing error in the regret bound.

\begin{algorithm2e}[tb]
\caption{Adaptation to Unknown Shifted Comparator Loss}\label{alg:adaptive}
\KwIn{Domain~$\mathcal K$, norm bound~$G>0$, width bound~$W\ge\width(\mathcal K)$.}
\KwOut{Predictions~$w_1,w_2,\ldots\in\mathcal K$.}
Set~$M\leftarrow2GW/\log2$ and fix the coordinate representation and initialization rule of \cref{alg:oco} for all base learners\;
\For{global rounds~$t=1,2,\ldots$}{
Set~$i_{\max}(t)\mathrel{\leftarrow}\floor{\log_2t}$ and~$K\mathrel{\leftarrow} i_{\max}(t)+2$\;
\If{$t=2^{i_{\max}(t)}$}{Set~$i\leftarrow i_{\max}(t)$, $\eps_i\leftarrow2^{-i}/8$, and~$\eta_i\leftarrow\eps_i/\prn{128M}$\;
Start learner~$i$ at local round~$1$ of \cref{alg:oco} with~$\eps_i$ and input bound~$2G$. Set~$\widetilde L_{t-1,i}\leftarrow0$\;}
Obtain~$w_t^{(i)}$ from each base learner~$0\le i\le i_{\max}(t)$\;
Approximate~$p_t$ in~\eqref{eq:mix-round-weights} by~$\widehat p_t\in\triangle_K$ satisfying~\eqref{eq:mix-approximation}\;
Play~$w_t$ from~\eqref{eq:master-prediction} and observe~$g_t$\;
\For{$i=0,\ldots,i_{\max}(t)$}{
Set $r_{t,i}\mathrel{\leftarrow}\inpr{g_t,w_t-w_t^{(i)}}$ and
$\widetilde L_{t,i}\mathrel{\leftarrow} \widetilde L_{t-1,i}-\Lambda_{\eps_i/2}(r_{t,i})$\;
Supply~$\operatorname{scale}_{\eps_i/2}(r_{t,i})g_t$ to base learner~$i$ at its local round~$t-2^i+1$\;}}
\end{algorithm2e}

\subsection{Sufficient Condition for the Regret Bound}\label{app:base-property}
For~$u\in\mathcal K$ and~$T\ge1$, we bound~$R_T^{\mathrm{lin}}(u)$ through the sum of~$\Lambda_{2\eps_i}(r_t(u))$ over all rounds for each learner~$i$ with~$t_i\le T$.
Using~$r_t(u)=\inpr{g_t,w_t-u}$ and~\eqref{eq:convex-transfer}, we obtain
\begin{equation}\label{eq:adaptive-loss-identity}
\sum_{t=1}^T\Lambda_{2\eps_i}(r_t(u))
=\prn{1-2\eps_i}R_T^{\mathrm{lin}}(u)-4\eps_iN_T(u).
\end{equation}
For the fixed-parameter bound, set~$C_0\coloneqq cG\width(\mathcal K)+M$ with a universal~$c\ge1$ satisfying~\eqref{eq:main-signed}. For adaptation to unknown~$N_T(u)$, set~$C_{\mathrm{ad}}\coloneqq C_0+256M$.
Since~$1-2\eps_i\ge3/4$, an upper bound of order~$C_{\mathrm{ad}}/\eps_i$ on the sum in~\eqref{eq:adaptive-loss-identity} yields~$R_T^{\mathrm{lin}}(u)\lesssim C_{\mathrm{ad}}/\eps_i+\eps_iN_T(u)$.
Below we bound the sum on rounds~$t_i,\ldots,T$ by comparing the mixture with~$u$ through learner~$i$. We then include the earlier rounds and minimize the regret bound over all~$i$ with~$t_i\le T$.

\subsection{\texorpdfstring{Comparison with~$u$ via a Base Learner}{Comparison with u via a Base Learner}}\label{app:scalar}
We compare the mixture with~$u$ through each base learner~$i$ and apply its fixed-parameter guarantee.
For~$T\ge t_i$, we show that the scaled feedback in~\eqref{eq:scalar-correction} yields
\begin{equation}\label{eq:adaptive-decomposition}
\sum_{t=t_i}^T\Lambda_{2\eps_i}(r_t(u))
\le\decompterm{appendix}{A}{\sum_{t=t_i}^T\Lambda_{\eps_i/2}(r_{t,i})}
+\decompterm{appendix}{B}{\sum_{t=t_i}^T\Lambda_{\eps_i/2}\prn{\inpr{g_t^{(i)},w_t^{(i)}-u}}}.
\end{equation}
Term~\decompref{appendix}{A} compares the mixture with learner~$i$, and term~\decompref{appendix}{B} compares learner~$i$ with~$u$ on its scaled inputs.

To prove~\eqref{eq:adaptive-decomposition}, we establish the following inequality.
For~$0<\theta\le1/5$ and all~$r,a\in\R$, we have
\begin{equation}\label{eq:scalar-bridge}
\Lambda_{4\theta}(r)\le\Lambda_\theta(a)
+\Lambda_\theta\prn{\operatorname{scale}_\theta(a)\prn{r-a}}.
\end{equation}
The right-hand side vanishes at~$r=0$ by~\eqref{eq:scalar-correction}, since
$\Lambda_\theta\prn{-\operatorname{scale}_\theta(a)a}=-\operatorname{scale}_\theta(a)\prn{a+\theta\abs{a}}=-\Lambda_\theta(a)$.
The right-hand side has the following slope bounds in~$r$, with~$+$ for~$r<a$ and~$-$ for~$r>a$:
\begin{equation}\label{eq:scalar-slopes}
1-4\theta\le\frac{\prn{1-\theta}^2}{1+\theta}
\le\operatorname{scale}_\theta(a)\prn{1\pm\theta}
\le\frac{\prn{1+\theta}^2}{1-\theta}\le1+4\theta.
\end{equation}
Integrating from~$0$ to~$r$ gives the lower bound~$\prn{1-4\theta}r$ for~$r\ge0$ and~$\prn{1+4\theta}r$ for~$r\le0$, proving~\eqref{eq:scalar-bridge}.
For~$\theta=\eps_i/2$, $r=r_t(u)$, and~$a=r_{t,i}$, the scaled excess satisfies
$\operatorname{scale}_{\eps_i/2}(r_{t,i})\prn{r_t(u)-r_{t,i}}=\inpr{g_t^{(i)},w_t^{(i)}-u}$.
Summing~\eqref{eq:scalar-bridge} with these substitutions for $t=t_i,\ldots,T$ proves~\eqref{eq:adaptive-decomposition}.

We now bound~\decompref{appendix}{B} using~\eqref{eq:main-signed}.
Fix~$\theta\in(0,1/16]$. On local round~$s$, let~$w'_s$ be the prediction of \cref{alg:oco} with parameter~$2\theta$ and input bound~$2G$, and let~$g'_s$ be its input with~$\norm{g'_s}_2\le2G$.
With~$\eps=2\theta$ and~$2G$ in place of~$G$, \eqref{eq:main-signed} implies, for every~$n\ge1$,
\begin{equation}\label{eq:base-property}
\sum_{s=1}^n\Lambda_\theta\prn{\inpr{g'_s,w'_s-u}}
\le\frac{cG\width(\mathcal K)}{2\theta}\le\frac{C_0}{\theta}.
\end{equation}
For learner~$i$, the global rounds~$t=t_i,\ldots,T$ are its first~$T-t_i+1$ local rounds, and its inputs satisfy~$\norm{g_t^{(i)}}_2\le2G$.
Applying~\eqref{eq:base-property} with~$\theta=\eps_i/2$ therefore gives
\begin{equation}\label{eq:adaptive-comparator-bound}
\decompref{appendix}{B}\le2C_0/\eps_i.
\end{equation}
It remains to bound~\decompref{appendix}{A}, using the recursive potential.

\subsection{Comparison with a Base Learner and Cancellation of the Mixing Error}\label{app:mixing-analysis}
We bound~\decompref{appendix}{A} by comparing the root potential with learner~$i$'s cumulative corrected loss.
We then see how the weighted absolute-value terms offset the mixing error when we subtract the root increments.
We first show this cancellation for exact weights, then use the same variance bound for the approximate weights in \cref{alg:adaptive}.

\subsubsection{Recursive Potential and Cumulative Corrected Losses}
For any integer~$K\ge2$ and loss vector~$\widetilde L\in\R^K$, \eqref{eq:mix-recursion} gives the following comparison at each split:
\begin{equation}\label{eq:mix-path}
V_k(\widetilde L)-\min\set{\widetilde L_k,V_{k+1}(\widetilde L)}
\le\frac{\log2}{\eta_k}\qquad(0\le k\le K-2).
\end{equation}
Applying~\eqref{eq:mix-path} along the path to coordinate~$i$ gives
\begin{equation}\label{eq:mix-path-bound}
V_0(\widetilde L)-\widetilde L_i
\le(\log2)\sum_{k=0}^{\min\set{i,K-2}}\eta_k^{-1}
\le\frac{256M\log2}{\eps_i}\qquad(0\le i\le K-1).
\end{equation}
The last inequality uses~$\eta_k=\eps_k/\prn{128M}$ and~$\sum_{k=0}^i\eps_k^{-1}\le2\eps_i^{-1}$.

To sum the root increments through~$T\ge1$, fix~$K\coloneqq\floor{\log_2T}+2$ coordinates.
For~$0\le t\le T$, write~$\widetilde L_t\coloneqq\prn{\widetilde L_{t,0},\ldots,\widetilde L_{t,K-1}}$ with $\widetilde L_{t,j}=0$ for $t<t_j$, so~$\widetilde L_0=0$ holds.
Extend both weight vectors in~\eqref{eq:mix-approximation} to~$K$ coordinates by repeatedly splitting the last leaf's weight equally between its children as needed. Both child losses are zero, so~\eqref{eq:mix-recursion} gives equal weights and preserves the root value.
Splitting exact mass~$x$ and approximate mass~$y$ preserves the~$\ell_1$ distance between the weight vectors since~$2\abs{x/2-y/2}=\abs{x-y}$.
In this dimension we write~$p_t$ and~$\widehat p_t$ for the extended weights, with~$p_t=p(\widetilde L_{t-1})$.
Because~$r_{t,j}=0$ for $t<t_j$, \eqref{eq:mix-zero-mean} continues to hold.
Let~$V_0^{(t)}\coloneqq V_0(\widetilde L_t)$ denote the root potential after round~$t$.
For a learner~$i$ with $t_i\le T$, its cumulative corrected loss gives
\begin{equation}\label{eq:mix-final-comparison}
\decompref{appendix}{A}=-\widetilde L_{T,i}
=\prn{V_0^{(T)}-\widetilde L_{T,i}}-V_0^{(T)}
\quad\text{and}\quad
V_0^{(T)}-\widetilde L_{T,i}\le\frac{256M\log2}{\eps_i},
\end{equation}
where the upper bound follows from~\eqref{eq:mix-path-bound}.

\subsubsection{Cancellation for Exact Weights}\label{app:mixing-exact-weights}
To evaluate~$-V_0^{(T)}$ in~\eqref{eq:mix-final-comparison}, define the round-$t$ mixing error by
\begin{equation}\label{eq:mix-round-gap}
E_{t,0}\coloneqq-\sum_{j=0}^{K-1}p_{t,j}\Lambda_{\eps_j/2}(r_{t,j})
-\prn{V_0^{(t)}-V_0^{(t-1)}}.
\end{equation}
The following lemma bounds this error for exact and approximate weights.
\begin{lemma}\label{lem:mixing-one-round}
Fix an integer~$K\ge2$, $\widetilde L\in\R^K$, and~$a_j\in[-M,M]$ for~$0\le j\le K-1$.
Set~$\widetilde L'_j\coloneqq\widetilde L_j-\Lambda_{\eps_j/2}(a_j)$ for each~$j$.
Suppose~$\widehat p\in\triangle_K$ and~$\delta\ge0$ satisfy~$\norm{\widehat p-p(\widetilde L)}_1\le\delta$ and~$\sum_{j=0}^{K-1}\widehat p_j a_j=0$.
Then, with~$U\coloneqq\sum_{j=0}^{K-1}\widehat p_j(\eps_j/2)\abs{a_j}$, we have
\begin{equation}\label{eq:mix-one-round-bound}
-\sum_{j=0}^{K-1}p_j(\widetilde L)\Lambda_{\eps_j/2}(a_j)
-\prn{V_0(\widetilde L')-V_0(\widetilde L)}
\le\frac13\prn{U+M\delta}.
\end{equation}
\end{lemma}
We will prove \cref{lem:mixing-one-round} in \cref{sec:proof-lem-mixing-one-round}.
Below we bound~\decompref{appendix}{A} when~$\widehat p_t=p_t$ on every round.
In this case, canceling the linear terms by~\eqref{eq:mix-zero-mean} gives
\begin{equation}\label{eq:mix-exact-mean}
\sum_{j=0}^{K-1}p_{t,j}\Lambda_{\eps_j/2}(r_{t,j})
=\sum_{j=0}^{K-1}p_{t,j}r_{t,j}-\sum_{j=0}^{K-1}p_{t,j}\frac{\eps_j}{2}\abs{r_{t,j}}
=-\sum_{j=0}^{K-1}p_{t,j}\frac{\eps_j}{2}\abs{r_{t,j}}.
\end{equation}
Since~$V_0^{(0)}=0$, summing the increments in~\eqref{eq:mix-round-gap} gives
\begin{equation}\label{eq:mix-root-telescope}
-V_0^{(T)}
=-\sum_{t=1}^T\prn{V_0^{(t)}-V_0^{(t-1)}}
=\sum_{t=1}^T\prn{E_{t,0}+\textstyle\sum_{j=0}^{K-1}p_{t,j}\Lambda_{\eps_j/2}(r_{t,j})}.
\end{equation}
Since~$\abs{r_{t,j}}\le G\operatorname{diam}_2(\mathcal K)\le M$, applying~\eqref{eq:mix-one-round-bound} with~$\widetilde L=\widetilde L_{t-1}$, $a_j=r_{t,j}$, $\widehat p=p_t$, and~$\delta=0$ gives~$E_{t,0}\le\frac13\sum_jp_{t,j}(\eps_j/2)\abs{r_{t,j}}$.
Using~\eqref{eq:mix-exact-mean} in~\eqref{eq:mix-root-telescope} and substituting into~\eqref{eq:mix-final-comparison} gives
\begin{equation}\label{eq:mix-exact-cancellation}
\decompref{appendix}{A}
\le\frac{256M\log2}{\eps_i}
+\sum_{t=1}^T\underbrace{\prn{E_{t,0}-\textstyle\sum_{j=0}^{K-1}p_{t,j}\prn{\eps_j/2}\abs{r_{t,j}}}}_{\le-\frac23\sum_{j=0}^{K-1}p_{t,j}(\eps_j/2)\abs{r_{t,j}}\le0}
\le\frac{256M\log2}{\eps_i}.
\end{equation}

\subsubsection{Common Variance Bound}\label{sec:proof-lem-mixing-one-round}
We prove \cref{lem:mixing-one-round} by bounding the mixing error through summation of variance bounds over the splits.
At each split we use the two-loss case of \citet[Lemma~4]{derooij2014follow}, derived there from \citet[Lemma~A.5]{cesabianchi2006prediction}.
For~$b\ge0$, $\eta>0$ with~$2\eta b\le1$, $x,y\in[-b,b]$, and~$q\in[0,1]$, put~$d\coloneqq-\eta^{-1}\log\prn{q\mathrm e^{-\eta x}+\prn{1-q}\mathrm e^{-\eta y}}$.
Then we have
\begin{equation}\label{eq:binary-gap}
0\le qx+\prn{1-q}y-d\le\eta q\prn{1-q}\prn{x-y}^2.
\end{equation}
Let~$X$ take values~$x,y$ with probabilities~$q,1-q$. Jensen's inequality gives~$d\le\E X$. For the upper bound, write~$\eta\prn{\E X-d}=\log\E\mathrm e^{-\eta\prn{X-\E X}}$. Since~$\abs{\eta\prn{X-\E X}}\le1$, the inequalities~$\mathrm e^{-z}\le1-z+z^2$ for~$\abs{z}\le1$ and~$\log\prn{1+z}\le z$ for~$z\ge0$ give~$\E X-d\le\eta\operatorname{Var}(X)$.

Fix an arbitrary loss vector~$\widetilde L\in\R^K$ and an update~$v\in[-2M,2M]^K$.
Abbreviate~$p_j\coloneqq p_j(\widetilde L)$ and~$P_j\coloneqq P_j(\widetilde L)$ for~$0\le j\le K-1$, and~$p_k^{\mathrm L}\coloneqq p_k^{\mathrm L}(\widetilde L)$ for~$0\le k\le K-2$.
For this pair~$\prn{\widetilde L,v}$, define the mean update, the potential increment, and the mixing error at each node~$0\le k<K$ by
\begin{equation}\label{eq:mix-tail}
\mu_k\coloneqq\frac{\sum_{j=k}^{K-1}p_jv_j}{P_k},\qquad
d_k\coloneqq V_k(\widetilde L+v)-V_k(\widetilde L)
\quad\text{and}\quad E_k\coloneqq\mu_k-d_k.
\end{equation}
Taking~$\widetilde L=\widetilde L_{t-1}$ and~$v_j=-\Lambda_{\eps_j/2}(r_{t,j})$ gives~$E_0=E_{t,0}$ by~\eqref{eq:mix-round-gap}. The positive weights give~$\abs{\mu_k}\le2M$. Monotonicity and translation equivariance of~\eqref{eq:mix-recursion} give~$\abs{d_k}\le2M$.
For~$0\le k\le K-2$, set~$b_k\coloneqq p_k^{\mathrm L}v_k+\prn{1-p_k^{\mathrm L}}d_{k+1}-d_k$.
Dividing the exponential sum in~\eqref{eq:mix-recursion} for~$\widetilde L+v$ by the sum for~$\widetilde L$ gives~$\mathrm e^{-\eta_k d_k}=p_k^{\mathrm L}\mathrm e^{-\eta_k v_k}+\prn{1-p_k^{\mathrm L}}\mathrm e^{-\eta_k d_{k+1}}$. Since~$4M\eta_k\le4M\eta_0=1/256$, \eqref{eq:binary-gap} applies to~$b_k$. Using~$\mu_k=p_k^{\mathrm L}v_k+\prn{1-p_k^{\mathrm L}}\mu_{k+1}$ in~\eqref{eq:mix-tail} gives
\begin{equation}\label{eq:mix-gap-recursion}
E_{K-1}=0,\qquad E_k=\prn{1-p_k^{\mathrm L}}E_{k+1}+b_k
\quad\text{and}\quad E_0=\sum_{k=0}^{K-2}P_kb_k.
\end{equation}
Since~$b_k\ge0$, the recursion gives~$P_{k+1}E_{k+1}\le E_0$ and~$E_k\ge0$. Also, $E_k\le4M$ holds by~\eqref{eq:mix-tail}.
For~$0\le k\le K-2$, weighting the variance of the two child means by~$P_k$ gives~$D_k\coloneqq P_kp_k^{\mathrm L}\prn{1-p_k^{\mathrm L}}\prn{v_k-\mu_{k+1}}^2$.
Using~$\prn{v_k-d_{k+1}}^2\le2\prn{v_k-\mu_{k+1}}^2+2E_{k+1}^2$ and~$E_{k+1}^2\le4ME_{k+1}$ in~\eqref{eq:binary-gap}, then summing by~\eqref{eq:mix-gap-recursion}, gives
\begin{equation}\label{eq:mix-gap-variance}
E_0\le2\sum_{k=0}^{K-2}\eta_kD_k
+8M\sum_{k=0}^{K-2}\eta_kp_k^{\mathrm L}P_{k+1}E_{k+1}
\le2\sum_{k=0}^{K-2}\eta_kD_k+E_0/64.
\end{equation}
For the last inequality, each~$p_k^{\mathrm L}P_{k+1}E_{k+1}$ is at most~$E_0$ by~\eqref{eq:mix-gap-recursion}, and~$8M\sum_{k=0}^{K-2}\eta_k\le16M\eta_0=1/64$.

To bound the variance sum, put~$m\coloneqq\mu_0$ and~$J\coloneqq\sum_{j=0}^{K-1}p_j\eps_j\abs{v_j}$.
We prove the bound
\begin{equation}\label{eq:mix-joint-variance}
\sum_{k=0}^{K-2}\eps_kD_k\le4MJ+2M\eps_0\abs{m}.
\end{equation}
The identity~$D_k=p_kv_k^2+P_{k+1}\mu_{k+1}^2-P_k\mu_k^2$, multiplied by~$\eps_k$ and summed over~$k$, gives
\begin{equation}\label{eq:mix-variance-telescope}
\sum_{k=0}^{K-2}\eps_kD_k
=\sum_{j=0}^{K-2}\eps_jp_jv_j^2+\eps_{K-2}p_{K-1}v_{K-1}^2
+\sum_{k=1}^{K-2}(\eps_{k-1}-\eps_k)P_k\mu_k^2-\eps_0m^2.
\end{equation}
Using~$\abs{\mu_k}\le2M$ and writing the tail sum as~$m-\sum_{j=0}^{k-1}p_jv_j$, we obtain
\begin{equation}\label{eq:mix-prefix}
P_k\mu_k^2\le2M\abs*{\sum_{j=k}^{K-1}p_jv_j}
\le2M\prn*{\abs{m}+\sum_{j=0}^{k-1}p_j\abs{v_j}}.
\end{equation}
Substitute~\eqref{eq:mix-prefix} and~$v_j^2\le2M\abs{v_j}$ into~\eqref{eq:mix-variance-telescope}.
For~$0\le j\le K-2$, the coefficient of~$2Mp_j\abs{v_j}$ is~$\eps_j+\sum_{k=j+1}^{K-2}\prn{\eps_{k-1}-\eps_k}$ and is at most~$2\eps_j$ by telescoping.
For~$j=K-1$, the coefficient equals~$\eps_{K-2}$, which is~$2\eps_{K-1}$.
The coefficient of~$2M\abs{m}$ is at most~$\eps_0$.
Dropping~$-\eps_0m^2$ proves~\eqref{eq:mix-joint-variance}.

To finish the proof of \cref{lem:mixing-one-round}, take~$v_j=-\Lambda_{\eps_j/2}(a_j)$ for~$0\le j\le K-1$.
Then we have~$\abs{v_j}\le2\abs{a_j}\le2M$. Since~$\sum_j\widehat p_ja_j=0$, we also have~$\sum_j\widehat p_jv_j=U$.
By the hypothesis~$\sum_j\abs{p_j-\widehat p_j}\le\delta$, we have~$\abs{m-U}=\abs{\sum_j\prn{p_j-\widehat p_j}v_j}\le2M\delta$.
Since~$\eps_j\le1$, the same weight bound gives~$J\le\sum_j\widehat p_j\eps_j\abs{v_j}+2M\delta\le4U+2M\delta$.
With~$\eta_k=\eps_k/\prn{128M}$, \eqref{eq:mix-joint-variance} therefore gives
\begin{equation}\label{eq:mix-variance-absorption}
\sum_{k=0}^{K-2}\eta_kD_k
\le J/32+\eps_0\abs{m}/64\le\frac9{64}\prn{U+M\delta}.
\end{equation}
For this update, \eqref{eq:mix-gap-variance} and~\eqref{eq:mix-variance-absorption} give~$E_0\le2\prn{U+M\delta}/7\le\prn{U+M\delta}/3$, proving~\eqref{eq:mix-one-round-bound}.

\subsubsection{Comparison with a Base Learner Using Approximate Weights}
We use \cref{lem:mixing-one-round} to bound~\decompref{appendix}{A} for the approximate weights~$\widehat p_t$ of \cref{alg:adaptive}. Recall from~\eqref{eq:mix-approximation} that~$\norm{\widehat p_t-p_t}_1\le\delta_t^{\mathrm{mix}}=\prn{t+1}^{-2}$.
For the extended vectors on~$K$ coordinates, define
$U_t\coloneqq\sum_{j=0}^{K-1}\widehat p_{t,j}(\eps_j/2)\abs{r_{t,j}}$.
By~\eqref{eq:mix-zero-mean}, we have~$U_t=-\sum_j\widehat p_{t,j}\Lambda_{\eps_j/2}(r_{t,j})\ge0$.
Since~$\abs{\Lambda_{\eps_j/2}(r_{t,j})}\le2M$, replacing~$\widehat p_t$ by~$p_t$ in this mean gives~$-\sum_jp_{t,j}\Lambda_{\eps_j/2}(r_{t,j})\ge U_t-2M\delta_t^{\mathrm{mix}}$.
Applying \cref{lem:mixing-one-round} with~$\widetilde L=\widetilde L_{t-1}$, $a_j=r_{t,j}$, $\widehat p=\widehat p_t$, and~$\delta=\delta_t^{\mathrm{mix}}$ yields~$E_{t,0}\le(U_t+M\delta_t^{\mathrm{mix}})/3$.
Thus the increments in~\eqref{eq:mix-round-gap} satisfy
\begin{equation}\label{eq:mix-approximate-increment}
V_0^{(t)}-V_0^{(t-1)}
\ge U_t-2M\delta_t^{\mathrm{mix}}-E_{t,0}
\ge\frac23U_t-\frac73M\delta_t^{\mathrm{mix}}.
\end{equation}
Since~$V_0^{(0)}=0$, summing~\eqref{eq:mix-approximate-increment} over~$t=1,\ldots,T$ and substituting into~\eqref{eq:mix-final-comparison} gives
\begin{equation}\label{eq:mix-bound}
\decompref{appendix}{A}+\frac23\sum_{t=1}^TU_t
\le\frac{256M\log2}{\eps_i}+\frac73M\sum_{t=1}^T\delta_t^{\mathrm{mix}}.
\end{equation}
Since~$U_t\ge0$ and~$\sum_{t=1}^\infty\delta_t^{\mathrm{mix}}<1$, \eqref{eq:mix-bound} gives~$\decompref{appendix}{A}\lesssim M/\eps_i$. With~$M=2GW/\log2$, we recover the final~$O(GW/\eps_i)$ bound in~\eqref{eq:mix-target}. We use~\eqref{eq:mix-bound} in \cref{app:master-analysis}.

\subsection{Regret over All Rounds}\label{app:master-analysis}\label{app:master-optimization}
We bound the mixture's linearized regret over all~$T$ rounds in terms of~$N_T(u)$. For~$t_i\le T$, the sum in~\eqref{eq:adaptive-decomposition} starts at~$t_i$.
For $t<t_i$, $\Lambda_{2\eps_i}(r_t(u))\le r_t(u)\le M$ gives
$\sum_{t=1}^{t_i-1}\Lambda_{2\eps_i}(r_t(u))\le M\prn{t_i-1}\le M/\prn{8\eps_i}$.
Combining this inequality with~\eqref{eq:adaptive-decomposition}, \eqref{eq:adaptive-comparator-bound}, and~\eqref{eq:mix-bound} yields
\begin{equation}\label{eq:master-signed}
\sum_{t=1}^T\Lambda_{2\eps_i}(r_t(u))\le2C_{\mathrm{ad}}/\eps_i,\quad\text{where } C_{\mathrm{ad}}=C_0+256M=cG\width(\mathcal K)+257M.
\end{equation}
Indeed, the mixture term and the sum over $t<t_i$ are together bounded by $512M/\eps_i$, since~$256\log2+7/24+1/8<512$.
Since~$1-2\eps_i\ge3/4$, \eqref{eq:adaptive-loss-identity} gives
\begin{equation}\label{eq:master-ratewise}
R_T^{\mathrm{lin}}(u)\le\frac{8C_{\mathrm{ad}}}{3\eps_i}+\frac{16}{3}\eps_iN_T(u)\qquad(T\ge2^i).
\end{equation}

\begin{theorem}\label{thm:adapt}
With a supplied~$M\ge G\operatorname{diam}_2\mathcal K$ in place of~$2GW/\log2$, \cref{alg:adaptive}
satisfies, for arbitrary inputs $g_t$ with~$\norm{g_t}_2\le G$, simultaneously for all~$u\in\mathcal K$ and~$T\ge1$,
\begin{equation}\label{eq:adaptive-representation}
R_T^{\mathrm{lin}}(u)\le24\prn*{C_{\mathrm{ad}}+\sqrt{C_{\mathrm{ad}}N_T(u)}},
\quad\text{where}\;\;C_{\mathrm{ad}}=cG\width(\mathcal K)+257M.
\end{equation}
For subgradient feedback, $R_T(u)\le R_T^{\mathrm{lin}}(u)$ holds, and the bound on~$R_T^{\mathrm{lin}}(u)$ also holds with~$\Delta_T(u)$ in place of~$N_T(u)$.
\end{theorem}
\begin{proof}
Put~$n\coloneqq N_T(u)$.
If~$n<C_{\mathrm{ad}}$, choosing~$i=0$ in~\eqref{eq:master-ratewise} gives a bound below~$22C_{\mathrm{ad}}$.
Otherwise choose~$i\coloneqq\floor{\log_2\sqrt{n/C_{\mathrm{ad}}}}$.
Since~$n\le MT$ and~$C_{\mathrm{ad}}\ge M$, we have~$2^i\le\sqrt T\le T$, so learner~$i$ is awake at round~$T$.
Moreover, $\sqrt{C_{\mathrm{ad}}/n}/8\le\eps_i\le\sqrt{C_{\mathrm{ad}}/n}/4$ holds.
Substitution into~\eqref{eq:master-ratewise} gives~$R_T^{\mathrm{lin}}(u)\le(68/3)\sqrt{C_{\mathrm{ad}}n}$.
This proves the bound on the linearized regret for arbitrary bounded inputs.
For subgradients, convexity gives~$R_T(u)\le R_T^{\mathrm{lin}}(u)$, and~\eqref{eq:negative-gap} gives~$N_T(u)\le\Delta_T(u)$.
\end{proof}
Taking~$M=2GW/\log2$ with~$W\ge\width(\mathcal K)$ gives~$C_{\mathrm{ad}}\lesssim GW$. Thus \cref{thm:adapt} implies \cref{thm:adaptive-known-width}.
If~$M\asymp G\operatorname{diam}_2\mathcal K$ is supplied directly, \eqref{eq:width-radius} instead gives~$C_{\mathrm{ad}}\asymp G\width(\mathcal K)$.

\section{Implementation and Arithmetic Complexity}\label{app:computation}
We bound the number of arithmetic operations used by \cref{alg:oco} in \cref{prop:cost} and by \cref{alg:adaptive} in \cref{app:adaptive-computation}.
For \cref{alg:oco}, we bound the cost of one internal step of \cref{alg:move-revoke}, the number of active added regularization terms, and the number of internal steps.

In what follows, we count arithmetic operations and comparisons on exact real numbers. When evaluating a projection onto a ball, taking a square root counts as one operation.
We remark that these counts do not bound the encoding lengths of the numbers stored by the algorithms or returned by the projection oracle.
Projections onto~$\mathcal K$, products with~$A$ and~$A^\top$, and application-specific optimization calls are counted separately.
In general, the coordinate map~$A$ and the initial parameters are supplied, and computing them is not counted.
For an arbitrary domain, \cref{prop:width-equivalence} allows us to choose a representation in which the number~$N$ of columns of~$A$ satisfies~$N\le5^r$, where~$r=\dim\operatorname{aff}(\mathcal K)$.
For the domains in \cref{cor:canonical-width}, \cref{app:initialization-arithmetic} shows how to compute the initial parameters from the domain representation using~$O\prn{N+\log\prn{N+1}}$ operations, and the bounds below include this cost.

For the coordinate initialization in \cref{app:general-core}, recall that~$\underline h_j=2^{-J_j}$ holds for~$j\in\set{1,\ldots,N}$ and that there are~$M_0=\sum_{j=1}^NJ_j$ initial terms, and put~$J_{\max}\coloneqq\max_{1\le j\le N}J_j$.
The fixed initialization in \cref{app:fixed-initialization} has no revocable initial terms, and we set~$M_0=J_{\max}=0$ for it.
Its matrix and addition scale are either~$B$ and~$S$ from \cref{app:fixed-initialization} or~$\widehat B$ and~$\widehat S$ from \cref{app:initialization-arithmetic}.

\begin{restatable}{proposition}{costbound}\label{prop:cost}
With the coordinate initialization in \cref{app:general-core} or the fixed initialization in \cref{app:fixed-initialization}, \cref{alg:oco} on inputs satisfying~$\norm{g_t}_2\le G$ uses
\begin{equation}\label{eq:cost-total}
O\prn*{\prn{T+M_0}N^2\eps^{-2}\brc{1+J_{\max}+\log\prn{T+2}}}
\end{equation}
arithmetic operations through round~$T$ and~$O\prn{N^3\eps^{-4}\brc{1+J_{\max}+\log\prn{t+2}}^2}$ in round~$t$. Both bounds exclude the following costs per round: one projection onto~$\mathcal K$, one product with each of~$A$ and~$A^\top$, and~$O\prn{d}$ operations for the translation and for the gradient correction in~\eqref{eq:correction}.
\end{restatable}
\Cref{app:quadratic-objective} bounds the cost of one internal step, one addition, and one revocation.
\Cref{app:active} bounds the number of active added regularization terms, and \cref{app:initialization-arithmetic} shows how to compute the initial parameters for the domains in \cref{cor:canonical-width}.
\Cref{app:operation-counts} counts the internal steps and proves \cref{prop:cost}.

\subsection{Stored Data and Cost of One Internal Step}\label{app:quadratic-objective}
We describe the data stored by \cref{alg:oco} and bound the cost of one internal step of \cref{alg:move-revoke}, one addition, and one revocation.

\paragraph{Regularizer and Inverse Hessian.}
We store the regularizer as~$\Psi(z)=z^\top Hz/2+b^\top z+c$, where~$H=\nabla^2\Psi$, $b\in\R^N$ is the vector of linear coefficients, and~$c\in\R$ is the constant term. We also store~$H^{-1}$ and the current point~$z$.
By~\eqref{eq:ball-cancellation}, revoking~$Q_i$ at~$a$ subtracts~$\gamma_i\inpr{q_i,z-a}^2/2$ from~$\Psi$, so the coefficients change by
\begin{equation}\label{eq:tangent-coefficients}
H\leftarrow H-\gamma_iq_iq_i^\top,\qquad
b\leftarrow b+\gamma_i\inpr{q_i,a}q_i\qquad\text{and}\qquad c\leftarrow c-\frac{\gamma_i}{2}\inpr{q_i,a}^2.
\end{equation}
At~$a$, the changes to~$Ha$ and~$b$ sum to zero, so~$\nabla\Psi(a)$ is preserved, as stated in \cref{sec:discrete-step}.
Adding~$Q_t$ in~\eqref{eq:ball-quadratic} adds~$\gamma_tq_tq_t^\top$, $-\gamma_tm_tq_t$, and~$\gamma_tm_t^2/2$ to~$H$, $b$, and~$c$, respectively.
For an addition or a revocation with coefficient~$\gamma$ and direction~$q$, the Sherman--Morrison formula gives
\begin{equation}\label{eq:rank-one-inverse}
\prn{H\pm\gamma qq^\top}^{-1}=H^{-1}\mp\frac{\gamma\prn{H^{-1}q}\prn{H^{-1}q}^\top}{1\pm\gamma q^\top H^{-1}q}.
\end{equation}
For~$x,y\in\R^N$, the matrix determinant lemma gives~$\det\prn{H+xy^\top}=\det(H)\prn{1+y^\top H^{-1}x}$.
Taking~$x=\pm\gamma q$ and~$y=q$, we obtain~$1\pm\gamma q^\top H^{-1}q=\det\prn{H\pm\gamma qq^\top}/\det H>0$.
For an addition we have~$H+\gamma qq^\top\succeq H\succeq B\succ0$. For a revocation we have~$H-\gamma qq^\top\succeq B$ because the remaining active terms have positive semidefinite Hessians.
Each addition or revocation updates~$H$, $H^{-1}$, $b$, and~$c$ in~$O\prn{N^2}$ operations.
Initially, $H_1$ is diagonal: it equals the supplied~$B$ or the computed~$\widehat B$ for the fixed initialization, and its~$j$th diagonal entry is~$\lambda\ell_j\prn{2/\underline h_j-1}$ for the coordinate initialization by~\eqref{eq:initial-quadratics}.
Hence~$H_1^{-1}$ is computed in~$O\prn{N}$ operations.

\paragraph{Compressed Initial Terms.}
For the coordinate initialization, we do not store the initial entries of~$\mathcal Q$ individually.
In coordinate~$j$, the initial terms~$Q^0_{jk}$, indexed by~$0\le k<J_j$, have direction~$e_j$, residual~$z_j$ at~$z$, threshold~$h_{jk}=2^k\underline h_j$, and coefficient~$\gamma_{jk}=\lambda\ell_j/h_{jk}$ by~\eqref{eq:initial-quadratics}.
As in \cref{app:initial-scales}, let~$\widehat h_j$ be the smallest threshold among the active initial terms in coordinate~$j$, or one if none is active.
The revoked thresholds form a prefix in each coordinate, so we store only~$\widehat h_j$. If~$\widehat h_j<1$, the largest ratio in~$L$ among the active initial terms in coordinate~$j$ is~$\abs{\prn{\widetilde g_{t,H}}_j}/\widehat h_j$.
After each move, for each coordinate~$j$, while~$\widehat h_j<1$ and~$\abs{z_j}\ge\widehat h_j$, we revoke the initial term with threshold~$\widehat h_j$ and coefficient~$\lambda\ell_j/\widehat h_j$ using~\eqref{eq:ball-cancellation} and double~$\widehat h_j$.
This loop revokes exactly the active initial terms with~$\abs{z_j}\ge h_{jk}$, as \cref{alg:move-revoke} requires.

\paragraph{Cost of One Internal Step.}
Each internal step computes~$\widetilde g_{t,H}=H^{-1}\widetilde g_t$ once in~$O\prn{N^2}$ operations and uses it both to evaluate~$L$ and to move~$z$.
At the beginning of round~$t$, the terms added in earlier rounds that remain active are indexed by~$\mathcal I_{t-1}$ (see \cref{app:lifetime-proof}). No new term is added during the call to \cref{alg:move-revoke}.
For each such term~$Q_i$ still active, the step evaluates~$\abs{\inpr{q_i,\widetilde g_{t,H}}}/h_i$ before the move and tests~$\abs{\inpr{q_i,z}-m_i}\ge h_i$ after it, using~$O\prn{N\abs{\mathcal I_{t-1}}}$ operations in total.
With coordinate initialization, scanning the~$N$ thresholds takes~$O\prn{N}$ operations. Thus the arithmetic cost per internal step, excluding~$O\prn{N^2}$ per revocation, is
\begin{equation}\label{eq:step-cost}
O\prn*{N^2+N\abs{\mathcal I_{t-1}}}.
\end{equation}
At the end of round~$t$, we use~$O\prn{N^2}$ operations to compute~$s_t^2=\widetilde g_t^\top\prn{H_t^-}^{-1}\widetilde g_t$ and, if~$s_t^2>\zeta_t$, add~$Q_t$.

\subsection{Number of Added Terms Remaining Active}\label{app:active}
At the beginning of round~$t$, the Hessian~$H$ contains one rank-one term for each~$i\in\mathcal I_{t-1}$. We bound the number~$\abs{\mathcal I_{t-1}}$ of active terms added in earlier rounds by comparing upper and lower bounds on~$\det(H)/\det(B)$.
The following argument applies to the coordinate initialization, the fixed initialization, and \cref{alg:ball-oco} with the positive addition thresholds in \cref{app:ball-implementation}.
In each case, we write the addition threshold as~$\zeta_s=\bar\zeta(s+1)^{-4}$: the scale is~$\bar\zeta=L_{\underline h}$ for the coordinate initialization by~\eqref{eq:general-skip}, $\bar\zeta=S$ for the fixed initialization, and~$\bar\zeta=\eps/\ceil{\sqrt d}$ for \cref{alg:ball-oco}.
When the fixed initialization uses the computed matrix~$\widehat B$, we use~$\widehat B$ in place of~$B$ throughout and take~$\bar\zeta=\widehat S$.
Recall from~\eqref{eq:surviving-initial-matrix} that~$B_{t-1}$ is the sum of~$B$ and the Hessians of the initial terms active at the start of round~$t$. If there are no active initial terms, we have~$B_{t-1}=B$.
The Hessian at the start of round~$t$ is therefore
\begin{equation}\label{eq:active-hessian}
H=B_{t-1}+\sum_{s\in\mathcal I_{t-1}}\gamma_s\widetilde g_s\widetilde g_s^\top.
\end{equation}

\paragraph{Lower Bound.}
For~$i\in\mathcal I_{t-1}$, take~$\overline H_i=B_{t-1}+\sum_{s\in\mathcal I_{t-1},\,s<i}\gamma_s\widetilde g_s\widetilde g_s^\top$ and~$\overline s_i^2=\widetilde g_i^\top\overline H_i^{-1}\widetilde g_i$ from \cref{app:survivor-proof} with~$T=t-1$.
All terms in~$\overline H_i-B$ were active just before~$Q_i$ was added, so we have~$\overline H_i\preceq H_i^-$ and hence~$\overline s_i^2\ge s_i^2$. By~\eqref{eq:weight-range}, we have~$\gamma_is_i^2>\eps^2/96$.
The matrix determinant lemma therefore gives
\begin{equation}\label{eq:active-det-ratio}
\frac{\det\prn{\overline H_i+\gamma_i\widetilde g_i\widetilde g_i^\top}}{\det\overline H_i}=1+\gamma_i\overline s_i^2\ge1+\frac{\eps^2}{96}.
\end{equation}
Adding the terms in~\eqref{eq:active-hessian} to~$B_{t-1}$ in increasing order yields~$H$. The product of the ratios in~\eqref{eq:active-det-ratio} is therefore~$\det H/\det B_{t-1}$, with the empty product equal to~$1$.
Since~$B_{t-1}\succeq B$, we obtain
\begin{equation}\label{eq:active-det-lower}
\frac{\det H}{\det B}=\frac{\det B_{t-1}}{\det B}\prod_{i\in\mathcal I_{t-1}}\prn{1+\gamma_i\overline s_i^2}\ge\prn*{1+\frac{\eps^2}{96}}^{\abs{\mathcal I_{t-1}}}.
\end{equation}

\paragraph{Upper Bound.}
The eigenvalues of~$B^{-1/2}HB^{-1/2}$ are positive, and their product and sum are~$\det(H)/\det(B)$ and~$\tr(B^{-1/2}HB^{-1/2})$, respectively.
Applying the arithmetic--geometric mean inequality to these eigenvalues gives
\begin{equation}\label{eq:active-det-upper}
\frac{\det H}{\det B}\le\prn*{\frac{\tr(B^{-1/2}HB^{-1/2})}{N}}^N.
\end{equation}
For~$i\in\mathcal I_{t-1}$, we have~$s_i^2>\zeta_i$ by the addition rule. Hence~\eqref{eq:parameters} gives~$\gamma_i=\eps^2/\prn{48s_i^2}<\eps^2(i+1)^4/\prn{48\bar\zeta}$.
Multiplying~\eqref{eq:active-hessian} by~$B^{-1}$ and taking the trace, we obtain
\begin{equation}\label{eq:active-trace}
\tr(B^{-1/2}HB^{-1/2})
\le\tr(B^{-1}B_{t-1})+\frac{\eps^2}{48\bar\zeta}\sum_{s\in\mathcal I_{t-1}}(s+1)^4\,\widetilde g_s^\top B^{-1}\widetilde g_s.
\end{equation}
Combining~\eqref{eq:active-det-lower} with~\eqref{eq:active-det-upper}, and using~$\log\prn{1+\eps^2/96}\ge\eps^2/97$ for~$\eps\in(0,1]$, we obtain
\begin{equation}\label{eq:active-log}
\abs{\mathcal I_{t-1}}\le\frac{97N}{\eps^2}\log\frac{\tr(B^{-1/2}HB^{-1/2})}{N}.
\end{equation}

\paragraph{Three Initializations.}
For each initialization, we bound~$\tr(B^{-1}B_{t-1})$ and~$\widetilde g_s^\top B^{-1}\widetilde g_s/\bar\zeta$ in~\eqref{eq:active-trace}, using~$\sum_{s\in\mathcal I_{t-1}}(s+1)^4\le(t-1)t^4\le(t+1)^5$.
\begin{itemize}[leftmargin=*,nosep]
\item For the coordinate initialization, we have~$B=\lambda\diag(\ell_1,\ldots,\ell_N)$ with~$\lambda=16/\eps$. By~\eqref{eq:compressed-initial}, we have~$\tr(B^{-1}B_{t-1})=\sum_{j=1}^N\prn{2/\widehat h_j-1}\le2^{J_{\max}+1}N$.
The inputs~$\widetilde g_s=A^\top\bar g_s/G$ satisfy~$\abs{\widetilde g_{s,j}}\le\norm{Ae_j}_2\le\ell_j$, so we have~$\widetilde g_s^\top B^{-1}\widetilde g_s\le\lambda^{-1}\sum_{j=1}^N\ell_j\le\prn{\eps/16}2^{J_{\max}}L_{\underline h}$, where the last inequality uses~$\underline h_j\ge2^{-J_{\max}}$.
Thus the right-hand side of~\eqref{eq:active-trace} is at most~$2^{J_{\max}+1}N+2^{J_{\max}}\eps^3(t+1)^5/768\le2^{J_{\max}}N\brc{2+(t+1)^5}$.
\item For the fixed initialization, $B_{t-1}=B$ gives~$\tr(B^{-1}B_{t-1})=N$. Since~$\widetilde g_s\widetilde g_s^\top\preceq A^\top A$, we have~$\widetilde g_s^\top B^{-1}\widetilde g_s\le\tr(A^\top AB^{-1})=\eps S/10$ by~\eqref{eq:fixed-initial-balance}. For the computed matrix $\widehat B$, \eqref{eq:rounded-addition-scale} gives~$\tr(A^\top A\widehat B^{-1})=\eps\widehat S/10$.
Thus the right-hand side of~\eqref{eq:active-trace} is at most~$N+\eps^3(t+1)^5/480$.
\item For \cref{alg:ball-oco}, we have~$N=d$, $B=\prn{10\ceil{\sqrt d}/\eps}I_d$, $B_{t-1}=B$, and~$\norm{\widetilde g_s}_2\le1$. Hence we have~$\tr(B^{-1}B_{t-1})=d$ and~$\widetilde g_s^\top B^{-1}\widetilde g_s\le\eps/\prn{10\ceil{\sqrt d}}=\bar\zeta/10$.
Thus the right-hand side of~\eqref{eq:active-trace} is at most~$d+\eps^2(t+1)^5/480$.
\end{itemize}
In each case, the trace divided by~$N$ is at most~$2^{J_{\max}}\brc{2+(t+1)^5}\le2^{J_{\max}}(t+2)^5$, with~$J_{\max}=0$ for the last two cases.
Substituting this bound into~\eqref{eq:active-log}, and noting that at most one term is added per round, gives
\begin{equation}\label{eq:active-population}
\abs{\mathcal I_{t-1}}\le\frac{97N}{\eps^2}\brc{J_{\max}+5\log\prn{t+2}}\quad\text{and}\quad\abs{\mathcal I_{t-1}}\le t-1.
\end{equation}
In particular, we have~$\abs{\mathcal I_{t-1}}=O\prn{N\eps^{-2}\brc{1+J_{\max}+\log\prn{t+2}}}$.

\subsection{Computing Initial Parameters}\label{app:initialization-arithmetic}
For the domains in \cref{cor:canonical-width}, we compute the initial parameters of \cref{alg:oco} using~$O\prn{N+\log\prn{N+1}}$ arithmetic operations and comparisons, and we show that the regret bound of the fixed initialization holds with the computed matrix.
We use the supplied geometric description: the translation and isometry, together with the radius for a ball, the semiaxis lengths for an ellipsoid, or the half-widths for a box.
Let~$m\coloneqq\ceil{\log_2N}$. Starting at one and doubling until the result is at least~$N$ gives~$m$ as the number of doublings, in~$O\prn{\log\prn{N+1}}$ operations.

\paragraph{\texorpdfstring{Balls with $p<2$ and the Simplex.}{Balls with p < 2 and the Simplex.}}
For a ball~$w_\circ+RU\B_p^N$ with~$1\le p<2$, the coordinate initialization uses the admissible description in \cref{app:scaled-norm-balls}, with~$\ell=R\mathbf1$ and~$\underline h_i=2^{-J_i}$. By~\eqref{eq:lp-threshold}, we have~$J_i=\ceil{m/p}$ for every coordinate~$i$.
Comparing~$pk$ with~$m$ for~$k=0,1,\ldots$ finds~$\ceil{m/p}$ as the smallest integer~$k\ge0$ with~$pk\ge m$, using~$O\prn{\log\prn{N+1}}$ operations, and forming~$\ell$ and~$\underline h$ uses~$O\prn{N+\log\prn{N+1}}$ additional operations.
By \cref{app:simplex-extension}, the simplex uses the same thresholds with~$p=1$ and~$N=d$.

\paragraph{\texorpdfstring{Fixed Matrices for $p\ge2$.}{Fixed Matrices for p >= 2.}}
The diagonal entries of~$B$ in~\eqref{eq:fixed-initial-matrix} involve the powers~$S^{q-1}$ and~$a_i^{2-q}$ with real exponents.
We instead compute a diagonal matrix~$\widehat B$ with~$B\preceq\widehat B\preceq3B$.
\begin{itemize}[leftmargin=*,nosep]
\item For a ball~$w_\circ+RU\B_p^N$ with~$2\le p<\infty$, we have~$a=R\mathbf1$ and~$B=\prn{10RN^{1/p}/\eps}I_N$. With~$k\coloneqq\ceil{m/p}$, computed as above, take~$\widehat B\coloneqq\prn{10R2^k/\eps}I_N$.
Since~$m\ge\log_2N$, we have~$2^k\ge2^{m/p}\ge N^{1/p}$. Since~$m<\log_2N+1$ and~$k<m/p+1$, we also have~$2^k<2^{1+1/p}N^{1/p}\le2\sqrt2N^{1/p}<3N^{1/p}$.
Thus~$B\preceq\widehat B\preceq3B$ holds. For~$p=\infty$, the ball is the box with~$a=R\mathbf1$.
\item For an ellipsoid with semiaxis lengths~$a_1,\ldots,a_N>0$, we have~$p=2$ and~$B=\prn{10\norm{a}_2/\eps}I_N$. Compute~$\sum_{i=1}^Na_i^2$ and~$\max_ia_i$. Starting from~$c=\max_ia_i$, double~$c$ until~$c^2\ge\sum_{i=1}^Na_i^2$, and take~$\widehat B\coloneqq\prn{10c/\eps}I_N$.
Since~$\max_ia_i\ge\norm{a}_2/\sqrt N$, at most~$\ceil{\log_2\sqrt N}$ doublings occur, so the computation uses~$O\prn{N+\log\prn{N+1}}$ operations, independently of the ratios between the semiaxis lengths. The final value satisfies~$\norm{a}_2\le c<2\norm{a}_2$, which gives~$B\preceq\widehat B\preceq2B$.
\item For a box with half-widths~$a_1,\ldots,a_N>0$, take the exact matrix~$\widehat B\coloneqq B=\prn{10/\eps}\diag(a_i)$ in~$O\prn{N}$ operations.
\end{itemize}
For each of these matrices, we compute the addition scale
\begin{equation}\label{eq:rounded-addition-scale}
\widehat S\coloneqq\frac{10}{\eps}\tr(A^\top A\widehat B^{-1}),
\end{equation}
in~$O\prn{N}$ operations: since~$A^\top A=\diag(a_i^2)$ and~$\widehat B$ is diagonal, the trace is~$\sum_{i=1}^N a_i^2/\widehat B_{ii}$.
The exact matrix satisfies~$\tr(A^\top AB^{-1})=\eps S/10$ by~\eqref{eq:fixed-initial-balance}, so~$B\preceq\widehat B\preceq3B$ implies~$S/3\le\widehat S\le S$.
\Cref{alg:oco} uses~$\widehat B$ in place of~$B$ and the addition threshold~$\widehat S(t+1)^{-4}$.
The entries of~$\eps\widehat B$ and the scale~$\widehat S$ do not depend on~$\eps$, so they are computed once and shared by the base learners in \cref{alg:adaptive}.

\paragraph{Regret Bound with the Computed Matrix.}
We explain how the above approximations affect the regret bound.
Since the inputs satisfy~$\widetilde g_t\widetilde g_t^\top\preceq A^\top A$ by~\eqref{eq:correction} and \cref{lem:projection}, we apply \cref{lem:fixed-initialization} with~$B=\widehat B$, $\Sigma=A^\top A$, and~$\zeta_t=\widehat S(t+1)^{-4}$.
We bound the three terms on the right-hand side of~\eqref{eq:fixed-distance}.
First, $\widehat B\preceq3B$ and~\eqref{eq:fixed-initial-balance} give~$\sup_{v\in\B_p^N}\norm{v}_{\widehat B}^2\le30S/\eps$.
Second, \eqref{eq:rounded-addition-scale} and~$\widehat S\le S$ give~$\prn{100/\eps^2}\tr(A^\top A\widehat B^{-1})=10\widehat S/\eps\le10S/\eps$.
Third, $\sum_{t\ge1}(t+1)^{-4}<1/8$ gives~$\sum_{t=1}^T\zeta_t<\widehat S/8\le S/8$.
Adding these bounds and using~$\eps\le1$, we obtain
\begin{equation}\label{eq:rounded-fixed-guarantee}
\sup_{v\in\B_p^N}\norm{v}_{\widehat B}^2+\frac{100}{\eps^2}\tr(A^\top A\widehat B^{-1})+\frac{\widehat S}{8}
\le\prn*{\frac{40}{\eps}+\frac18}S\le\frac{41S}{\eps}.
\end{equation}
For every~$v\in\B_p^N$, the left-hand side of~\eqref{eq:fixed-distance} with~$B=\widehat B$ is therefore at most~$41S/\eps\le164\width(\mathcal K)/\eps$ by~\eqref{eq:fixed-width-comparison}.

\subsection{Total Operation Counts of \texorpdfstring{\cref{alg:oco}}{Algorithm \ref{alg:oco}}}\label{app:operation-counts}
To prove \cref{prop:cost}, we count the internal steps of \cref{alg:move-revoke}. We then use \cref{app:quadratic-objective,app:active} to bound their cost and add the initialization cost from \cref{app:initialization-arithmetic}.

\paragraph{Internal Steps.}
Let~$K_t$ be the number of internal steps in round~$t$, with~$K_t=0$ for a zero input, and recall from \cref{lem:move-revoke} the set~$\mathcal J_t$ of terms revoked in round~$t$.
Recall also that~$M_0$ is the number of revocable initial terms.
By the termination argument in \cref{sec:movement-proof}, every nonfinal step revokes a term, so we have~$K_t\le1+\abs{\mathcal J_t}$.
Each term is revoked at most once. Through round~$T$, there are at most~$M_0+T$ terms: the initial terms and at most one added term per round. Summing over rounds therefore gives
\begin{equation}\label{eq:general-step-count}
\sum_{t=1}^TK_t\le2T+M_0.
\end{equation}

\begin{proof}[\cref{prop:cost}]
Initialization uses~$O\prn{N}$ operations for~$H_1^{-1}$ and, for the domains in \cref{cor:canonical-width}, $O\prn{N+\log\prn{N+1}}$ operations for the initial parameters.
Predicting~$w_t$ uses the product with~$A$, the translation, and the projection listed in \cref{prop:cost}. After correcting~$g_t$ to obtain~$\bar g_t$, we apply~$A^\top$ and divide the~$N$ coordinates by~$G$. Of these prediction and input operations, only the~$N$ divisions count toward~\eqref{eq:cost-total}.
By~\eqref{eq:general-step-count} and~\eqref{eq:step-cost}, the internal steps through round~$T$ use~$O\prn{\prn{2T+M_0}\prn{N^2+N\max_{t\le T}\abs{\mathcal I_{t-1}}}}$ operations.
There are at most~$T$ additions, at most~$T+M_0$ revocations, and at most~$T$ evaluations of~$s_t^2$. Each costs~$O\prn{N^2}$ operations.
Substituting~\eqref{eq:active-population}, we bound the total by
\begin{equation}
\begin{aligned}
&\prn{2T+M_0}\,O\prn*{N^2+N\cdot N\eps^{-2}\brc{1+J_{\max}+\log\prn{T+2}}}+\prn{3T+M_0}\,O\prn{N^2}\\
&=O\prn*{\prn{T+M_0}N^2\eps^{-2}\brc{1+J_{\max}+\log\prn{T+2}}},
\end{aligned}
\end{equation}
which proves~\eqref{eq:cost-total} in \cref{prop:cost}.
In round~$t$, every step except the last revokes at least one term, and the terms that can be revoked are the~$\abs{\mathcal I_{t-1}}$ active added terms and at most~$NJ_{\max}$ active initial terms. Hence we have~$K_t\le1+\abs{\mathcal I_{t-1}}+NJ_{\max}$.
Including its revocations, round~$t$ therefore uses~$O\prn{\prn{1+\abs{\mathcal I_{t-1}}+NJ_{\max}}\prn{N^2+N\abs{\mathcal I_{t-1}}}}$ operations.
Substituting~\eqref{eq:active-population} and~$NJ_{\max}\le N\eps^{-2}J_{\max}$ gives the bound~$O\prn{N^3\eps^{-4}\brc{1+J_{\max}+\log\prn{t+2}}^2}$ for round~$t$.
\end{proof}

\paragraph{Cost on the Ball.}\phantomsection\label{app:ball-cost}
We bound the number of active terms and the total arithmetic cost of the variant of \cref{alg:ball-oco} in \cref{app:ball-implementation}.
For \cref{alg:ball-oco} with the addition thresholds~$\zeta_t=\prn{\eps/\ceil{\sqrt d}}(t+1)^{-4}$ in \cref{app:ball-implementation}, we have~$N=d$, $A=I_d$, and~$M_0=J_{\max}=0$.
The active family at the start of round~$t$ consists of the added terms indexed by~$\mathcal I_{t-1}$, so~\eqref{eq:active-population} gives~$\abs{\mathcal A_t}\le\prn{485d/\eps^2}\log\prn{t+2}$.
The preceding count then gives~$O\prn{Td^2\eps^{-2}\log\prn{T+2}}$ arithmetic operations through round~$T$ and~$O\prn{d^3\eps^{-4}\log^2\prn{t+2}}$ in round~$t$.
The projection and the correction~\eqref{eq:ball-correction} use~$O\prn{d}$ additional operations per round.

\paragraph{\texorpdfstring{Bound Uniform in $\eps$.}{Bound Uniform in epsilon.}}
To bound the total cost of \cref{alg:adaptive} in \cref{app:adaptive-total-cost}, we bound each base learner's cost independently of its parameter~$\eps$.
Using only the second bound in~\eqref{eq:active-population}, namely~$\abs{\mathcal I_{t-1}}\le t-1$, in the proof of \cref{prop:cost}, we bound the number of operations through round~$T$ by
\begin{equation}\label{eq:all-store}
O\prn*{\prn{T+M_0}\prn{N^2+NT}}.
\end{equation}
The exclusions in \cref{prop:cost} still apply. At most one term is added per round for any thresholds~$\zeta_t\ge0$. The bound therefore holds when their values are supplied; if they must be computed, their evaluation cost is additional.

\subsection{Arithmetic Implementation of \texorpdfstring{\cref{alg:adaptive}}{Algorithm \ref{alg:adaptive}}}\label{app:adaptive-computation}
We give an arithmetic implementation of \cref{alg:adaptive} and bound its total cost.

\subsubsection{Evaluating the Recursive Mixture}
We construct weights~$\widehat p_t$ satisfying~\eqref{eq:mix-approximation} using~$O\prn{\log^2(t+2)}$ arithmetic operations and comparisons in round~$t$.
At round~$t$, write~$\widetilde L\coloneqq\prn{\widetilde L_{t-1,0},\ldots,\widetilde L_{t-1,i_{\max}(t)},0}$ for the input to~$p$ in~\eqref{eq:mix-round-weights}.
For $0\le i\le i_{\max}(t)+1$, set~$\bar L_i\coloneqq \widetilde L_i/M$ and~$\bar V_i\coloneqq V_i(\widetilde L)/M$.
For the split indices~$0\le i\le i_{\max}(t)$, set $\bar\eta_i\coloneqq M\eta_i=\eps_i/128$ and abbreviate~$p_i^{\mathrm L}\coloneqq p_i^{\mathrm L}(\widetilde L)$.
Since~$\abs{r_{s,i}}\le G\operatorname{diam}_2\mathcal K\le M$, each corrected loss satisfies~$\abs{\Lambda_{\eps_i/2}(r_{s,i})}\le2M$, so we have~$\abs{\bar L_i}\le2t$ for~$0\le i\le i_{\max}(t)+1$.
The last coordinate, which represents all learners that have not yet started, has value~$\bar V_{i_{\max}(t)+1}=\bar L_{i_{\max}(t)+1}=0$.
For~$i=i_{\max}(t),\ldots,0$, the exact recursion in~\eqref{eq:mix-recursion} gives
\begin{equation}\label{eq:mix-stable-recursion}
x_i\coloneqq\bar\eta_i\abs{\bar L_i-\bar V_{i+1}}\quad\text{and}\quad \bar V_i=\min\set{\bar L_i,\bar V_{i+1}}+
\frac{\log2-\log\prn{1+\mathrm e^{-x_i}}}{\bar\eta_i}.
\end{equation}
The exact split weight satisfies~$p_i^{\mathrm L}=\mathrm e^{-x_i}/\prn{1+\mathrm e^{-x_i}}$ if~%
$\bar L_i\ge \bar V_{i+1}$ and~$p_i^{\mathrm L}=1/\prn{1+\mathrm e^{-x_i}}$ otherwise.

We approximate the exponential and logarithms in~\eqref{eq:mix-stable-recursion} using~$b\coloneqq\ceil{\log_2(t+2)}\ge2$, found by repeated doubling.
There are at most~$b+1$ splits, and for each awake learner~$i$ we have~$\bar\eta_i^{-1}=1024\cdot2^i\le1024t$. At each split, we use the computed normalized loss and child potential both to approximate~$x_i$ by~$\widehat x_i$ and to choose the branch for the split weight.
For~$\widehat x_i\ge32b$, we approximate~$\mathrm e^{-\widehat x_i}$ by zero, with error at most~$\mathrm e^{-32b}<2^{-32b}$.
Otherwise, we choose the smallest integer~$r\ge0$ with~$\widehat x_i/2^r\le1$, evaluate the degree-$128b$ Taylor polynomial of~$\mathrm e^{-x}$ at~$x=\widehat x_i/2^r$, clip the value to~$[0,1]$, and square it~$r$ times.
The absolute Taylor remainder on~$[0,1]$ is at most~$1/\prn{128b+1}!\le2^{-128b}$, and squaring on this interval at most doubles the error. Since~$2^r<64b$, the error is at most~$2^{-32b}$.
For~$y\in[0,1]$, we have~$y/(2+y)\le1/3$, and we use the first~$128b$ terms of the series
\begin{equation}\label{eq:mix-log-series}
\log\prn{1+y}=2\sum_{j=0}^{\infty}\frac{1}{2j+1}\prn*{\frac{y}{2+y}}^{2j+1},
\end{equation}
whose tail after~$128b$ terms is bounded by~$2\sum_{j=128b}^{\infty}3^{-2j-1}<2^{-128b}$. We approximate~$\log2$ once per round at~$y=1$, and at each split we evaluate the series at the approximation of~$\mathrm e^{-\widehat x_i}$.
Evaluating either series uses~$O\prn{b}$ operations by updating each term from its predecessor.
We also use the approximation of~$\log2$ at~$y=1$ to compute~$\bar L_i=\widetilde L_i\log2/\prn{2GW}$ from the stored loss~$\widetilde L_i$. Since~$\abs{\bar L_i}\le2t$ and~$\log2>1/2$, the error in each normalized loss is at most~$4t\,2^{-128b}\le2^{-25b}$. If~$M$ is supplied directly, the division~$\widetilde L_i/M$ is exact.

We clip each computed split weight to~$[\underline p,1-\underline p]$, where~$\underline p\coloneqq2^{-8b}$, and form the leaf masses~$\widehat p_t$ from the clipped split weights by~\eqref{eq:mix-leaf-weights}, so that their sum is exactly one.
Clipping affects only the weights; the parent uses the computed potential in~\eqref{eq:mix-stable-recursion}.
When all leaves below a node have zero cumulative corrected loss, we set its value to zero and its split weight to~$1/2$ without approximation, since these are the exact values.

\subsubsection{Accuracy of Computed Weights}
We verify both conditions in~\eqref{eq:mix-approximation}: the total weight of the awake learners is positive, and~$\norm{\widehat p_t-p_t}_1\le(t+1)^{-2}$ holds.
With the computed inputs fixed, approximating~$\mathrm e^{-\widehat x_i}$, $\log 2$, and~$\log\prn{1+\mathrm e^{-\widehat x_i}}$ incurs a total error of at most~$3\bar\eta_i^{-1}2^{-32b}\le2^{-25b}$ in~\eqref{eq:mix-stable-recursion}. Indeed, the truncation error for each logarithm is at most~$2^{-128b}$, and~$y\mapsto\log\prn{1+y}$ is $1$-Lipschitz on~$[0,1]$.
The partial derivatives of~$\bar V_i$ with respect to~$\bar L_i$ and~$\bar V_{i+1}$ are~$p_i^{\mathrm L}$ and~$1-p_i^{\mathrm L}$, so perturbing both inputs changes the exact potential by at most the larger input error. Including the error in the normalized losses and summing the evaluation errors over at most~$b+1$ splits gives a potential error of at most~$\prn{b+2}2^{-25b}\le2^{-23b}$.
The exact split weight is~$1/\prn{1+\exp\prn{\bar\eta_i\prn{\bar L_i-\bar V_{i+1}}}}$, which is~$\bar\eta_i/4$-Lipschitz in the difference~$\bar L_i-\bar V_{i+1}$. For either branch, replacing~$y=\mathrm e^{-\widehat x_i}$ by its approximation changes~$y/\prn{1+y}$ or~$1/\prn{1+y}$ by at most~$2^{-32b}$. Thus before clipping, the split weight differs from~$p_i^{\mathrm L}$ by at most~$\prn{\bar\eta_i/4}\,2^{-22b}+2^{-32b}\le2^{-22b}$.

After clipping to~$[\underline p,1-\underline p]$, each split weight differs from its exact value by at most~$\underline p+2^{-22b}\le2\underline p$.
To compare the exact and computed leaf distributions, couple two random descents by using the same uniform draw on~$[0,1]$ at each split, independently across splits.
Conditional on both descents reaching a split, the probability of taking different branches equals the absolute difference between its two split weights, at most~$2\underline p$. By a union bound over at most~$b+1$ splits, the probability of reaching different leaves is at most~$2(b+1)\underline p$.
The~$\ell_1$ distance between the two leaf distributions is at most twice this probability. Thus we have
\begin{equation}\label{eq:mix-distribution-error}
\norm{\widehat p_t-p_t}_1\le4(b+1)2^{-8b}
\le2^{-2b}\le(t+1)^{-2}.
\end{equation}
The first leaf, which corresponds to learner~$0$, has mass at least~$\underline p>0$, so the denominator defining~$\pi_{t,i}$ in~\eqref{eq:master-prediction} is positive.

\subsubsection{Total Cost of \texorpdfstring{\cref{alg:adaptive}}{Algorithm \ref{alg:adaptive}}}\label{app:adaptive-total-cost}
In round~$t$, the recursion evaluates at most~$b+1$ nodes with~$O\prn{b}$ operations each, so computing~$\widehat p_t$ uses~$O\prn{\log^2(t+2)}$ arithmetic operations and stores~$O\prn{\log\prn{t+2}}$ reals and counters.
For the at most~$1+\log_2t$ awake base learners, forming~$w_t$, the loss differences~$r_{t,i}$, and the scaled inputs uses~$O\prn{d\log\prn{t+2}}$ operations.
Each base learner runs \cref{alg:oco} for at most~$T$ rounds. Its own parameter~$\eps_i$ does not enter~\eqref{eq:all-store}, so with~$N$ coordinates and~$M_0$ initial terms, it uses~$O\prn{\prn{T+M_0}\prn{N^2+NT}}$ operations through round~$T$.
Summing over the~$1+\floor{\log_2T}$ base learners started by round~$T$ gives
\begin{equation}\label{eq:master-cost}
O\prn*{\prn{T+M_0}\prn{N^2+NT}\log\prn{T+2}}
\end{equation}
arithmetic operations, plus~$O\prn{T\log\prn{T+2}}$ projections and as many products with each of~$A$ and~$A^\top$. The translations, convex combinations, loss differences, scaled inputs, and corrections in~\eqref{eq:correction} use~$O\prn{Td\log\prn{T+2}}$ further operations.
The bound~\eqref{eq:master-cost} includes the~$O\prn{T\log^2\prn{T+2}}$ operations for the mixture weights, since~$N\ge1$ and~$\log\prn{T+2}\le T+1$.
For contextual search on~$\B_2^d$ in \cref{sec:search}, we may take the smallest power of two~$W$ with~$W^2\ge2d$. Repeated doubling finds this value in~$O\prn{\log\prn{d+1}}$ operations, and~\eqref{eq:width-radius} gives~$\width(\B_2^d)\le\sqrt{2d}\le W<2\sqrt{2d}$. The fixed initialization has~$N=d$, $A=I_d$, and~$M_0=0$, and each projection uses~$O\prn{d}$ operations.
Then~\eqref{eq:master-cost} and the preceding counts give the total~$O\prn{Td\prn{d+T}\log\prn{T+2}}$ stated in \cref{sec:search}.

\section{Minimax Lower Bounds}\label{app:lower}\label{sec:sharp}
We prove matching lower bounds for $\ell_p$ balls, the simplex, ellipsoids, and rectangular boxes, even with full loss-function feedback.
In this section, losses are convex and~$G$-Lipschitz on~$\R^d$.
A \emph{comparator-loss budget}~$b\ge0$ requires~$\Delta_T^\star\le b$ on every realization. Let~$R_\infty^\star(\mathcal K,G,b)$ denote the infimum, over anytime strategies knowing~$b$, of the worst-case expected regret against the best fixed comparator over all horizons and loss sequences satisfying these conditions.

\begin{restatable}{theorem}{minimaxbound}\label{thm:minimax}
For every domain in \cref{cor:canonical-width}, $G>0$, and~$b\ge0$, we have
\begin{equation}\label{eq:minimax-width}
R_\infty^\star(\mathcal K,G,b)=
\Theta\prn*{G\width(\mathcal K)+\sqrt{G\width(\mathcal K)b}}.
\end{equation}
One subgradient per round suffices for the upper bound. The lower bound holds even with
full loss-function feedback. The implicit constants are universal, uniformly over~$p\in[1,\infty]$.
\end{restatable}
Below we write~$S_{\mathcal K}\coloneqq G\width(\mathcal K)$ for brevity. \Cref{thm:adaptive-known-width} gives the upper bound with~$W=\width(\mathcal K)$ without knowing~$b$. For the lower bound, we draw a loss sequence before play, independently of the learner's randomness. It satisfies~$\Delta_T^\star\le b$ on every realization. For each learner, its worst-case expected regret is therefore at least its expected regret under this distribution.

The rest of this section is organized as follows. \Cref{app:clean-lower} proves a lower bound of order~$S_{\mathcal K}$ even at~$b=0$. For the term~$\sqrt{S_{\mathcal K}b}$, \cref{app:coordinate-budget} proves a lower bound for ellipsoids and boxes. \Cref{app:sparse} constructs losses that force large expected regret while keeping the best expert's loss small. \Cref{app:sparse-embedding} uses these losses to prove~$R_\infty^\star(\mathcal K,G,b)\gtrsim\sqrt{S_{\mathcal K}b}$ for the simplex and~$\ell_p$ balls. Taking the larger of the two bounds gives the lower bound in~\eqref{eq:minimax-width}.

\subsection{Lower Bounds with a Common Minimizer}\label{app:clean-lower}
We prove a lower bound of order~$G\width(\mathcal K)$ for any learner by constructing losses with a common minimizer. Such losses satisfy every budget~$b\ge0$.
On~$\mathcal K=\B_p^d$ with~$d\ge2$, let~$1/p+1/q=1$ and fix an integer~$1\le k<d$. Write~$a\coloneqq k^{-1/p}$, with~$a=1$ for~$p=\infty$, and~$L\coloneqq ka=k^{1/q}$. For a set of coordinates~$\mathcal A$ with~$\abs{\mathcal A}\ge k$, define
\begin{equation}\label{eq:hypersimplex}
\mathcal S_{\mathcal A}\coloneqq\Set*{w\in[0,a]^d}{\operatorname{supp}(w)\subseteq \mathcal A,\ \sum_{i=1}^d w_i=L}.
\end{equation}
For finite~$p$, we have $\sum_{i=1}^d w_i^p\le a^{p-1}\sum_{i=1}^d w_i=1$ on~$\mathcal S_{\mathcal A}$. For~$p=\infty$, the coordinate cap is~$a=1$. Thus~$\mathcal S_{\mathcal A} \subseteq \mathcal K$ holds in both cases.

Draw a uniformly random permutation of the $d$ coordinates before play. Initially set~$\mathcal A\leftarrow\set{1,\ldots,d}$.
While~$\abs{\mathcal A}>k$, let~$i$ be the next coordinate in the permutation and use the loss $f_t(w)\coloneqq G\dist_2(w,\mathcal S_{\mathcal A\setminus\set{i}})$, then remove~$i$ from~$\mathcal A$. The point~$u$ with value~$a$ on the last~$k$ coordinates of the permutation and zero elsewhere belongs to every set, so it incurs zero loss.

To bound the loss, condition on the past and~$w_t$ with~$\abs{\mathcal A}=m>k$. Let~$\bar w_t\coloneqq\Pi_{\mathcal S_{\mathcal A}}(w_t)$. On the next face~$\mathcal C\coloneqq\mathcal S_{\mathcal A\setminus\set{i}}$, every~$z$ satisfies~$\Pi_{\mathcal S_{\mathcal A}}(z)=z$ and~$z_i=0$. Nonexpansiveness gives~$\dist_2(w_t,\mathcal C)\ge\dist_2(\bar w_t,\mathcal C)$. Since~$z_i=0$ on~$\mathcal C$, the latter distance is at least~$\bar w_{t,i}$. Past loss functions depend only on the coordinates already removed, so the next coordinate~$i$ is still uniform on~$\mathcal A$ given the past and~$w_t$. Averaging over~$i$ shows that the conditional expected loss is at least~$\prn{G/m}\sum_{j\in\mathcal A}\bar w_{t,j}=GL/m$. Summing from~$m=k+1$ to~$d$ gives
\begin{equation}\label{eq:clean-lp-lower}
R_\infty^\star(\B_p^d,G,0)
\ge G k^{1/q}(\mathrm H_d-\mathrm H_k),\quad\text{where}\;\; \mathrm H_j\coloneqq\sum_{i=1}^j i^{-1}.
\end{equation}
We choose~$k$ so that~$k^{1/q}(\mathrm H_d-\mathrm H_k)$ matches the order of~$\width(\B_p^d)$ in~\eqref{eq:unit-lp-width}.
If~$q\ge(\log d)/2$, taking~$k=1$ gives a lower bound of order~$G\log d$, and~$d^{1/q}\le\mathrm e^2$ holds. Otherwise, take~$k=\floor{d\mathrm e^{-q}}$. Then~$k\ge d\mathrm e^{-q}/2$ gives~$k^{1/q}\ge d^{1/q}/\prn{2\mathrm e}$, and integration gives~$\mathrm H_d-\mathrm H_k\ge\log\prn{\prn{d+1}/\prn{k+1}}\ge q-\log 2\ge\prn{1-\log 2}q$. Both choices give a lower bound of order~$G\width(\B_p^d)$. On the simplex, taking~$k=a=L=1$ gives successive faces of~$\triangle_d$ and the lower bound~$G(\mathrm H_d-1)\asymp G\log d$.

For the ellipsoid and box in~\eqref{eq:ellipsoid-def}, write~$x_i$ for coordinate~$i$ of~$U^\top(x-w_\circ)$ for each~$x\in\R^d$. For the ellipsoid, draw independent uniform signs~$\sigma_1,\ldots,\sigma_r$. Set~$u_i\coloneqq\sigma_i a_i^2/\norm{a}_2$, and use the loss~$G\abs{w_i-u_i}$ in round~$i$. All sign choices are feasible because~$\sum_{i=1}^r(u_i/ a_i)^2=1$. Given the prediction, the triangle inequality gives a lower bound of~$G a_i^2/\norm{a}_2$ on the expected loss over~$\sigma_i$. Since~$u$ incurs zero loss, the expected regret is at least~$G\norm{a}_2$. For a box, replace~$u_i$ by~$\sigma_i a_i$ to obtain~$G\sum_{i=1}^r a_i$. These lower bounds have order~$G\width(\mathcal K)$ by~\eqref{eq:norm-ball-width}, including in dimension one.

\subsection{Budget-Dependent Lower Bound for Ellipsoids and Boxes}\label{app:coordinate-budget}
We prove~$R_\infty^\star(\mathcal K,G,b)\gtrsim\sqrt{S_{\mathcal K}b}$ for ellipsoids and boxes.
We choose the half-widths $a'=\prn{a'_1,\ldots,a'_r}\in(0,\infty)^r$ so that the box~$w_\circ+U\diag(a'_i)\B_\infty^r$ lies in~$\mathcal K$, with~$w_\circ,U$ as in~\eqref{eq:ellipsoid-def}. Write~$x_i$ for coordinate~$i$ of~$U^\top(x-w_\circ)$.
Fix an integer~$n\ge1$. For each~$i\in\set{1,\ldots,r}$ and~$s\in\set{1,\ldots,n}$, set~$f_t\coloneqq f_{i,s}$ in round~$t=\prn{i-1}n+s$, where
\begin{equation}\label{eq:random-coordinate-loss}
f_{i,s}(w)\coloneqq G\abs{w_i-a'_i\xi_{i,s}}\quad\text{and}\quad \Pr(\xi_{i,s}=1)=\Pr(\xi_{i,s}=-1)=\tfrac12.
\end{equation}
Draw all signs $\xi_{i,s}$ independently at the start. Set~$S_i\coloneqq\sum_{s=1}^n\xi_{i,s}$ and choose the vertex~$u$ of this box with coordinates~$u_i\coloneqq a'_i\sgn(S_i)$, choosing either sign when~$S_i=0$. In coordinate~$i$, $u$ incurs loss only on the $\prn{n-\abs{S_i}}/2$ rounds with the opposite sign, at~$2Ga'_i$ per such round. Its total loss is therefore
\begin{equation}\label{eq:coordinate-budget}
\sum_{t=1}^{rn} f_t(u)=G\sum_{i=1}^r a'_i\prn{n-\abs{S_i}}\le G\norm{a'}_1 n.
\end{equation}
Each loss vanishes at a vertex with the corresponding sign, so~\eqref{eq:coordinate-budget} also bounds~$\Delta_{rn}(u)$.

Given the past and~$w_t$, the two signs are equally likely. By the triangle inequality, the conditional expected loss is~$G\prn{\abs{w_{t,i}-a'_i}+\abs{w_{t,i}+a'_i}}/2\ge Ga'_i$. Summing and subtracting the expected comparator loss in~\eqref{eq:coordinate-budget} gives~$\E R_{rn}(u)\ge G\sum_{i=1}^r a'_i\E\abs{S_i}$. To bound this sum, use~$\E S_i^2=n$ and~$\E S_i^4\le3n^2$. The Paley--Zygmund inequality applied to~$S_i^2$ then gives~$\Pr(S_i^2\ge n/2)\ge1/12$. Hence we have~$\E\abs{S_i}\ge\sqrt{n/2}/12$, which gives~$\E R_{rn}(u)\gtrsim G\norm{a'}_1\sqrt n$.

For an ellipsoid take~$a'_i=a_i^2/\norm{a}_2$ as in \cref{app:clean-lower}, and for a box take~$a'_i=a_i$. In either case (including~$r=1$), \eqref{eq:norm-ball-width} gives~$G\norm{a'}_1\asymp S_{\mathcal K}$. If~$b\ge2G\norm{a'}_1$, taking~$n=\floor{b/\prn{G\norm{a'}_1}}$ gives~$\Delta_{rn}(u)\le b$ by~\eqref{eq:coordinate-budget}. Since~$n\ge b/(2G\norm{a'}_1)$, the regret bound becomes~$\E R_{rn}(u)\gtrsim\sqrt{S_{\mathcal K}b}$. For smaller~$b$, we have~$\sqrt{S_{\mathcal K}b}\lesssim S_{\mathcal K}$. The best fixed comparator incurs at most the loss of~$u$, and \cref{app:clean-lower} gives a lower bound of order~$S_{\mathcal K}$ for every budget. Together these bounds give
\begin{equation}
R_\infty^\star(\mathcal K,G,b)\gtrsim S_{\mathcal K}+\sqrt{S_{\mathcal K}b}.
\end{equation}

\subsection{Small Comparator Loss and Large Regret on the Simplex}\label{app:sparse}
For the simplex and~$\ell_p$ balls, we need losses that force large expected regret while keeping the best comparator's loss small on every realization. To this end, we use 1-sparse losses \citep{kwon2016gains}: over~$n$ rounds, the best of~$m$ experts incurs loss at most~$n/m$.

Let~$m\ge2$ and~$n\ge m\log m$ be integers, and set $\delta\coloneqq\sqrt{m\log m/\prn{64n}}$. Draw a hidden index~$J$ uniformly at random from~$\set{1,\ldots,m}$.
Conditional on~$J$, draw~$I_1,\ldots,I_n$ independently from the distribution~$P_J$ on~$\set{1,\ldots,m}$ defined by
\begin{equation}\label{eq:sparse-distribution}
P_J(J)\coloneqq\frac{1-\delta}{m}\quad\text{and}\quad
P_J(i)\coloneqq\frac{1+\delta/\prn{m-1}}m\quad(i\in\set{1,\ldots,m}\setminus\set{J}).
\end{equation}
A learner plays~$p_t\in\triangle_m$, incurs~$\prn{p_t}_{I_t}$, and then sees~$I_t$.

\paragraph{Probability of Identifying the Hidden Expert.}
Before observing~$I_t$, draw an index~$\widehat J_t$ from the learner's prediction~$p_t$. We show that~$\Pr(\widehat J_t=J)=\E[(p_t)_J]\le2/3$: the observations so far carry little information about~$J$.
Relative to the uniform distribution~$P_0$ on~$\set{1,\ldots,m}$, we have
\begin{equation}\label{eq:sparse-kl}
\KL(P_J\Vert P_0)\le\chi^2(P_J,P_0)
=\frac{\delta^2}{m-1}\le\frac{2\delta^2}{m},
\end{equation}
where~$\KL(\cdot\Vert\cdot)$ denotes the Kullback--Leibler divergence.
Write~$I_{<t}\coloneqq(I_1,\ldots,I_{t-1})$. The learner and the draw from~$p_t$ use randomness independent of~$(J,I_1,\ldots,I_n)$, so~$\widehat J_t$ depends on~$J$ only through~$I_{<t}$. Applying the data-processing inequality and then the chain rule for relative entropy with reference law~$P_0^{\otimes(t-1)}$ for~$I_{<t}$ gives
\begin{equation}\label{eq:sparse-information}
\mathrm I(J;\widehat J_t)\le\mathrm I(J;I_{<t})
\le (t-1)\E_J\KL(P_J\Vert P_0)\le\frac{\log m}{32},
\end{equation}
where the last inequality uses~\eqref{eq:sparse-kl} and the choice of~$\delta$.

For~$m\ge4$, Fano's inequality for the uniform index~$J$ gives
$\Pr(\widehat J_t=J)\le(\mathrm I(J;\widehat J_t)+\log 2)/\log m\le17/32$ by~\eqref{eq:sparse-information}.
For~$m=2,3$, let~$P$ be the joint law of~$(J,\widehat J_t)$ and let~$Q$ be the product of its marginal laws.
Under~$Q$, $J$ is uniform and independent of~$\widehat J_t$, so the event~$\widehat J_t=J$ has probability~$1/m$.
By the definition of mutual information and~\eqref{eq:sparse-information}, we have~$\KL(P\Vert Q)=\mathrm I(J;\widehat J_t)\le\prn{\log m}/32$.
Pinsker's inequality then gives~$\TV(P,Q)\le\sqrt{\KL(P\Vert Q)/2}\le\sqrt{\prn{\log m}/64}$, where~$\TV(\cdot,\cdot)$ denotes the total variation distance.
Applying this bound to the event~$\widehat J_t=J$ and using its probability~$1/m$ under~$Q$ gives~$\Pr(\widehat J_t=J)\le1/m+\sqrt{\prn{\log m}/64}<2/3$.

\paragraph{Regret and Loss of the Best Expert.}\leavevmode
Conditional on~$J$, the expected loss difference between experts~$i\ne J$ and~$J$ is~$P_J(i)-P_J(J)=\delta/(m-1)$ by~\eqref{eq:sparse-distribution}. Averaging over~$J$ and~$p_t$ gives the learner's expected excess loss over~$J$ in round~$t$: $\delta(1-\E[(p_t)_J])/(m-1)\ge\delta/(3(m-1))$. Summing over~$t$ shows that the learner's expected regret against~$J$ is at least~$n\delta/(3(m-1))$.
Let~$L_n^\star$ be the smallest of the~$m$ realized counts of~$I_1,\ldots,I_n$. This is the loss of the best expert in hindsight: it is at most the loss of~$J$ and at most the average count~$n/m$. Substituting~$\delta$ thus gives the two bounds used in \cref{app:sparse-embedding}:
\begin{equation}\label{eq:sparse-lower}
L_n^\star\le\frac{n}{m}\quad\text{on every realization},\qquad
\text{and}\qquad
\E\brc*{\sum_{t=1}^n(p_t)_{I_t}-L_n^\star}\gtrsim\sqrt{\frac{n\log m}{m}}.
\end{equation}

\Needspace{4\baselineskip}
\subsection{\texorpdfstring{Budget-Dependent Lower Bound for the Simplex and~$\ell_p$ Balls}{Budget-Dependent Lower Bound for the Simplex and lp Balls}}\label{app:sparse-embedding}
For~$d\ge2$, we prove~$R_\infty^\star(\mathcal K,G,b)\gtrsim\sqrt{S_{\mathcal K}b}$ on~$\triangle_d$ and~$\B_p^d$ for all~$1\le p\le\infty$. We apply the bound from \cref{app:sparse} directly on the simplex and to a product of simplices inside the ball.

Choose positive integers~$k,m$ with~$km\le d$ and~$m\ge2$, and put~$a\coloneqq k^{-1/p}$. The product
\begin{equation}\label{eq:product-simplex}
\mathcal S\coloneqq\prn{a\triangle_m}^k\times\set{0}^{d-km}
\end{equation}
lies in~$\B_p^d$. For finite~$p$, the sum of the $p$th powers in each block is at most~$a^p=1/k$. For~$p=\infty$, every coordinate is at most~$a=1$ and nonnegative.
The following bounds apply whenever~$\mathcal S\subseteq\mathcal K$; for~$\mathcal K=\triangle_d$ we take~$k=1$ and~$m=d$, so that~$\mathcal S=\triangle_d$. Fix an integer~$n\ge m\log m$, the number of rounds per block. Run an independent copy of the construction in \cref{app:sparse} in each block~$j\in\set{1,\ldots,k}$. Let~$I_s^{(j)}\in\set{1,\ldots,m}$ be the coordinate selected at local round~$s\in\set{1,\ldots,n}$ in block~$j$.
At global round~$t=\prn{j-1}n+s$, set $i_t\coloneqq\prn{j-1}m+I_s^{(j)}$ and define on the whole space
\begin{equation}\label{eq:sparse-convex-loss}
f_t(w)\coloneqq\frac G2\prn{w_{i_t}+\dist_2(w,\mathcal S)}.
\end{equation}
This function is convex and~$G$-Lipschitz. For~$\bar w\coloneqq\Pi_{\mathcal S}(w)$, we have $w_{i_t}+\norm{w-\bar w}_2\ge \bar w_{i_t}\ge0$, and thus we obtain~$f_t(w)\ge f_t(\bar w)$. Its minimum on~$\mathcal K$ is zero: another coordinate in the relevant block can receive all the mass. Given~$i_t$ and the fixed set~$\mathcal S$, the learner can reconstruct the entire loss function from~\eqref{eq:sparse-convex-loss}. Thus full loss feedback gives no further information about the hidden index. For any prediction~$w_t$, the projection onto~$\mathcal S$ gives an expert distribution in the current block after division by~$a$. Multiplying its expert loss by~$Ga/2$ gives~$f_t(\Pi_{\mathcal S}(w_t))\le f_t(w_t)$.

\paragraph{Loss and Regret across Blocks.}
To satisfy the comparator budget, let~$u$ assign all the mass in each block to a coordinate with the smallest realized count. This choice puts~$u$ in~$\mathcal S$ and attains the best expert's loss in each block, scaled by~$Ga/2$. The smallest count is at most~$n/m$ in every block. Since the blocks are independent, the expected regret lower bound in~\eqref{eq:sparse-lower} applies to the projected predictions conditional on all earlier blocks. Summing over the~$k$ blocks gives
\begin{equation}\label{eq:sparse-block-transfer}
\Delta_{kn}(u)\le\frac{Gka}{2}\frac{n}{m}\quad\text{on every realization},\qquad
\text{and}\qquad\E R_{kn}(u)\gtrsim Gka\sqrt{\frac{n\log m}{m}}.
\end{equation}
If~$b\ge Gka\log m$, take~$n=\floor{2bm/(Gka)}$. Then~$m\log m\le bm/(Gka)\le n\le2bm/(Gka)$ holds, so the first bound in~\eqref{eq:sparse-block-transfer} gives~$\Delta_{kn}(u)\le b$, and the second gives
\begin{equation}\label{eq:sparse-budget-lower}
\E R_{kn}(u)\gtrsim\sqrt{Gka\log(m)\,b}.
\end{equation}

\paragraph{Simplex.}
Take~$k=1$ and~$m=d$, so that~$a=1$ and~$\mathcal S=\triangle_d$. On the simplex the distance in~\eqref{eq:sparse-convex-loss} vanishes, and the loss is~$(G/2)w_{I_t^{(1)}}$. By~\eqref{eq:simplex-comparison} and~\eqref{eq:unit-lp-width}, we have~$G\log d\asymp S_{\mathcal K}$. For~$b\ge G\log d$, \eqref{eq:sparse-budget-lower} gives~$R_\infty^\star(\mathcal K,G,b)\gtrsim\sqrt{S_{\mathcal K}b}$. For~$b<G\log d$, this follows from \cref{app:clean-lower} because~$\sqrt{S_{\mathcal K}b}\lesssim S_{\mathcal K}$.

\paragraph{\texorpdfstring{$\ell_p$ Balls.}{lp Balls.}}
For~$\mathcal K=\B_p^d$, let~$1/p+1/q=1$. Since~$ka=k^{1/q}$, we choose the number~$k$ of blocks and their size~$m$ to maximize~$k^{1/q}\log m$ in~\eqref{eq:sparse-budget-lower}. We show that
\begin{equation}\label{eq:block-optimization}
\max\Set*{k^{1/q}\log m}{k,m\in\mathbb{Z},\ k\ge1,\ m\ge2,\ km\le d}
\asymp d^{1/q}\min\set{q,\log(d+1)}.
\end{equation}
For the upper bound, $k\le d/m$ gives~$k^{1/q}\log m\le d^{1/q}m^{-1/q}\log m$. The last factor~$m^{-1/q}\log m$ is at most~$\log d$; for finite~$q$ it is also at most~$q/\mathrm e$, the maximum of~$x^{-1/q}\log x$ over~$x\ge1$.
For the lower bound, if~$q\ge(\log d)/2$, take~$k=1$ and~$m=d$. Since~$d^{1/q}\le\mathrm e^2$, this choice gives~$k^{1/q}\log m=\log d\gtrsim d^{1/q}\min\set{q,\log(d+1)}$ in \eqref{eq:block-optimization}. If~$q<(\log d)/2$, take~$m=\ceil{\mathrm e^q}$ and~$k=\floor{d/m}$. We have~$\mathrm e^q\le m\le 2\mathrm e^q\le d$ and~$k\ge d/(2m)$. Thus we have~$\log m\ge q$ and~$k^{1/q}\ge d^{1/q}/(4^{1/q}\mathrm e)$, which give the lower bound in~\eqref{eq:block-optimization}.
For a maximizing pair~$(k,m)$, \eqref{eq:block-optimization} and the width formula~\eqref{eq:unit-lp-width} give~$Gka\log m\asymp S_{\mathcal K}$ uniformly in~$p$ and~$d$. If~$b\ge Gka\log m$, \eqref{eq:sparse-budget-lower} gives~$R_\infty^\star(\mathcal K,G,b)\gtrsim\sqrt{S_{\mathcal K}b}$. For~$b<Gka\log m$, \cref{app:clean-lower} gives~$R_\infty^\star(\mathcal K,G,b)\gtrsim S_{\mathcal K}\gtrsim\sqrt{S_{\mathcal K}b}$. This completes the proof of \cref{thm:minimax}.

\section{Contextual Search with Comparison Noise}\label{app:noise}
We prove \cref{cor:search-adaptive}.
\Cref{app:noise-adaptive} bounds the cumulative prediction error with high probability for all horizons simultaneously and derives the expected-error bound.
\Cref{app:noise-lower} proves the matching lower bound under independent flips.
Recall the search procedure in \cref{sec:search}: we run \cref{alg:adaptive} on~$\B_2^d$
with~$G=1$ and~$W\ge\width(\B_2^d)$ of order~$\sqrt d$, predict~%
$\widehat y_t=\inpr{\phi_t,w_t}$, and supply~$g_t=Y_t\phi_t$ after observing the comparison bit.
These inputs give~$r_t(u)=Y_t\prn{\widehat y_t-y_t^\star}$ and~$\abs{r_t(u)}=\abs{\widehat y_t-y_t^\star}$, so~$R_T^{\mathrm{abs}}(u)$ is the cumulative absolute prediction error.
\restate{\adaptivesearchbound}

\subsection{High-Probability Bound on Cumulative Prediction Error}\label{app:noise-adaptive}
We bound the cumulative prediction error for all horizons on a single event and then derive the expected-error bound in \cref{cor:search-adaptive}.
\begin{proposition}\label{prop:noise-hp}
Under the assumptions of \cref{cor:search-adaptive}, for each~$\delta\in(0,1)$,
\cref{alg:adaptive} satisfies, with probability at least~$1-\delta$,
\begin{equation}\label{eq:adaptive-noise-hp}
\sup_{T\ge1}\sum_{t=1}^T\abs{\widehat y_t-y_t^\star}
\lesssim\frac{\sqrt d+\log\prn{1/\delta}}{\kappa^2}.
\end{equation}
\end{proposition}
\begin{proof}
We first compare the cumulative absolute prediction error with the linearized regret.
As in \cref{sec:search}, let~$\mathcal F_t$ be the sigma-algebra generated by the history, current context and prediction, and any tie-breaking sign.
Conditional on~$\mathcal F_t$, the linearized excess loss~$r_t(u)=Y_t\prn{\widehat y_t-y_t^\star}$ takes values in~$\set{\pm\abs{\widehat y_t-y_t^\star}}$ and has mean at least~$\kappa\abs{\widehat y_t-y_t^\star}$ by~\eqref{eq:noise-model}.
For each fixed~$\eta>0$, Hoeffding's lemma for this range of length~$2\abs{\widehat y_t-y_t^\star}$ and the bound~$\abs{\widehat y_t-y_t^\star}\le2$ give
\begin{equation}\label{eq:noise-mgf}
\begin{aligned}
\E\brc*{\exp\prn*{-\eta r_t(u)}\mid\mathcal F_t}
&\le\exp\prn*{-\eta\kappa\abs{\widehat y_t-y_t^\star}+\frac{\eta^2}{2}\abs{\widehat y_t-y_t^\star}^2}\\
&\le\exp\prn*{-\eta\prn{\kappa-\eta}\abs{\widehat y_t-y_t^\star}}.
\end{aligned}
\end{equation}
Since~$\abs{\widehat y_t-y_t^\star}$ is~$\mathcal F_t$-measurable, \eqref{eq:noise-mgf} shows that the products~$\prod_{t=1}^T\exp\prn{\eta\prn{\kappa-\eta}\abs{\widehat y_t-y_t^\star}-\eta r_t(u)}$ form a nonnegative supermartingale with initial value one for the filtration that includes the comparison bit at the end of each round.
By Ville's inequality, the probability that this supermartingale ever reaches~$1/\delta$ is at most~$\delta$. Hence, with probability at least~$1-\delta$, we have, for all~$T\ge1$,
\begin{equation}\label{eq:noise-martingale}
\prn{\kappa-\eta}R_T^{\mathrm{abs}}(u)
\le R_T^{\mathrm{lin}}(u)+\frac{\log\prn{1/\delta}}{\eta}.
\end{equation}

Set~$\eta=\kappa/4$. \Cref{thm:adaptive-known-width} with~$G=1$ gives~$R_T^{\mathrm{lin}}(u)\le c\prn{W+\sqrt{WN_T(u)}}$ for all~$T\ge1$, where~$c$ is a universal constant, and~$\brc{-r_t(u)}_+\le\abs{r_t(u)}$ gives~$N_T(u)\le R_T^{\mathrm{abs}}(u)$.
Substituting these bounds into~\eqref{eq:noise-martingale} gives, on the same event,
\begin{equation}\label{eq:noise-abs-recursion}
\frac{3\kappa}{4}R_T^{\mathrm{abs}}(u)
\le c\prn*{W+\sqrt{W R_T^{\mathrm{abs}}(u)}}+\frac{4\log\prn{1/\delta}}{\kappa}.
\end{equation}
Young's inequality gives~$c\sqrt{W R_T^{\mathrm{abs}}(u)}\le\kappa R_T^{\mathrm{abs}}(u)/4+c^2W/\kappa$.
Substituting this upper bound into~\eqref{eq:noise-abs-recursion} and rearranging with~$\kappa\le1$, we obtain, for all~$T\ge1$ on this event,
\begin{equation}\label{eq:noise-hp-explicit}
R_T^{\mathrm{abs}}(u)\le\frac{2\prn{c+c^2}W+8\log\prn{1/\delta}}{\kappa^2}.
\end{equation}
Since~$W\asymp\sqrt d$, this proves~\eqref{eq:adaptive-noise-hp}.
\end{proof}
The learner uses neither~$\kappa$ nor~$\delta$, and~$\eta$ appears only in the analysis.
Hence, for one run of the learner and every~$s>0$, \eqref{eq:noise-hp-explicit} with~$\delta=\mathrm e^{-s}$ holds for all~$T\ge1$ with probability at least~$1-\mathrm e^{-s}$, which gives
$\Pr\prn{\kappa^2\sup_{T\ge1}R_T^{\mathrm{abs}}(u)>2\prn{c+c^2}W+8s}\le\mathrm e^{-s}$.
Integrating this tail bound over~$s>0$ gives~$\E\brc{\sup_{T\ge1}R_T^{\mathrm{abs}}(u)}\le\prn{2\prn{c+c^2}W+8}/\kappa^2\lesssim\sqrt d/\kappa^2$.
Since~$\sup_{T\ge1}R_T^{\mathrm{abs}}(u)=\sum_{t=1}^\infty\abs{\widehat y_t-y_t^\star}$, this proves the upper bound in \cref{cor:search-adaptive}, even for the infinite sum.

\subsection{Lower Bound under Independent Flips}\label{app:noise-lower}
We prove the lower bound in \cref{cor:search-adaptive} by hiding an independent sign in each coordinate.
In each round, we independently flip the sign~$\sgn(\widehat y_t-y_t^\star)$ with probability~$\prn{1-\kappa}/2$ to generate the comparison bit. We take~$\sgn(0)\coloneqq1$.

\paragraph{Construction.}
Fix~$a>0$ and~$\kappa\in(0,1]$. Set~$n\coloneqq\floor{1/(16\kappa^2)}$ if~$\kappa\le1/4$ and~$n\coloneqq1$ otherwise.
Draw independent uniform signs~$\sigma_1,\ldots,\sigma_d$ before play and set~$u_i\coloneqq a\sigma_i$.
For each~$i\in\set{1,\ldots,d}$ in turn, present the context~$e_i$ in~$n$ consecutive rounds.
The contexts are fixed before play. Given the history and prediction, the bit law depends on~$\sigma_i$ only in a round with context~$e_i$, where~$y_t^\star=a\sigma_i$.
For each learner, there is a fixed sign vector for which the expected cumulative error is at least its average over~$\sigma$, as for the distributions in \cref{app:lower}.

\paragraph{Information Bound.}
Fix~$i$ and the signs~$\sigma_j$ for~$j\ne i$. Let~$P_i^+$ and~$P_i^-$ be the laws of the history through the last round with context~$e_i$, including the learner's internal randomness, under~$\sigma_i=1$ and~$\sigma_i=-1$, respectively.
In a round with context~$e_i$ and prediction~$\widehat y_t$, the bit has the same law under both signs unless~$-a\le\widehat y_t<a$.
In that case, conditional on the history and prediction, the laws of~$\prn{Y_t+1}/2$ under~$P_i^+$ and~$P_i^-$ are~$\operatorname{Bern}\prn{\prn{1-\kappa}/2}$ and~$\operatorname{Bern}\prn{\prn{1+\kappa}/2}$, respectively. Their relative entropy satisfies, for~$0<\kappa\le1/4$,
\begin{equation}\label{eq:noise-kl}
\KL\prn[\big]{\operatorname{Bern}\prn{\prn{1-\kappa}/2}\mathrel{\Vert}\operatorname{Bern}\prn{\prn{1+\kappa}/2}}
=\kappa\log\frac{1+\kappa}{1-\kappa}\le\frac{2\kappa^2}{1-\kappa}\le4\kappa^2.
\end{equation}
For~$0<\kappa\le1/4$, write~$P_i^\pm\prn{Y_t\mid\mathcal F_t}$ for the conditional laws of~$Y_t$ under~$P_i^\pm$.
The learner's internal randomness has the same law under both signs, and the conditional bit laws agree for~$t\le\prn{i-1}n$.
Thus the chain rule for relative entropy and~\eqref{eq:noise-kl} give
\begin{equation}\label{eq:noise-history-kl}
\begin{aligned}
\KL(P_i^+\Vert P_i^-)
&=\sum_{t=1}^{in}\E_{P_i^+}\brc*{\KL\prn*{P_i^+\prn{Y_t\mid\mathcal F_t}\mathrel{\Vert}P_i^-\prn{Y_t\mid\mathcal F_t}}}\\
&\le\sum_{t=\prn{i-1}n+1}^{in}4\kappa^2=4\kappa^2n\le\frac14.
\end{aligned}
\end{equation}
Pinsker's inequality and~\eqref{eq:noise-history-kl} then bound their total variation distance by~$\TV(P_i^+,P_i^-)\le\sqrt{\KL(P_i^+\Vert P_i^-)/2}\le\sqrt{1/8}<1/2$.

\paragraph{Error Bound.}
Each prediction in a round with context~$e_i$ is a function of the history before that round, which is part of the history with law~$P_i^\pm$.
The measure~$\min\set{\mathrm dP_i^+,\mathrm dP_i^-}$ has mass~$1-\TV(P_i^+,P_i^-)$, and on it the same history gives the same prediction under both signs.
Since~$\abs{\widehat y-a}+\abs{\widehat y+a}\ge2a$ for every~$\widehat y\in\R$, the expected error in such a round, averaged over~$\sigma_i$, is at least~$a\prn{1-\TV(P_i^+,P_i^-)}$.
For~$\kappa\le1/4$, this is at least~$a/2$. Averaging over the remaining signs, summing over the~$dn$ rounds and using~$n\ge1/(32\kappa^2)$, we find that the expected cumulative error is at least~$dna/2\ge da/(64\kappa^2)$.
For~$\kappa>1/4$, the law of the history before the first round with context~$e_i$ does not depend on~$\sigma_i$, so the expected error in that round is at least~$a$. The total is at least~$da\ge da/(16\kappa^2)$.
In both cases, the expected cumulative error is at least~$da/(64\kappa^2)$ for some fixed sign vector.

With~$a=1/\sqrt d$, every target lies on the Euclidean unit sphere and every context lies in~$\B_2^d$. The lower bound~$\sqrt d/(64\kappa^2)$ then matches the upper bound in \cref{cor:search-adaptive} up to a universal factor, which proves the minimax optimality in \cref{cor:search-adaptive}.

\paragraph{Cube Normalization.}
Suppose instead that~$u\in[-1,1]^d$ and~$\norm{\phi_t}_1\le1$.
For~$w\in[-1,1]^d$, we have~$\abs{\inpr{\phi_t,w}}\le1$, and~$g_t=Y_t\phi_t$ satisfies~$\norm{g_t}_2\le\norm{\phi_t}_1\le1$.
Run \cref{alg:adaptive} on~$[-1,1]^d$ with~$G=1$ and~$W=\width([-1,1]^d)\asymp d$ by~\eqref{eq:norm-ball-width}.
By the argument in \cref{app:noise-adaptive}, the algorithm bounds the expected cumulative error by~$O\prn{d/\kappa^2}$ without knowing~$\kappa$.
The construction above with~$a=1$ uses contexts~$e_i$ with~$\norm{e_i}_1=1$ and gives the matching lower bound~$d/(64\kappa^2)$.
\paragraph{Relation to Existing Lower Bounds.}
For exact comparison feedback, the known lower bounds are~$\Omega\prn{\sqrt d}$ on the Euclidean unit ball \citep[Proposition~E.1]{cosson2026contextual} and~$\Omega(d)$ under the cube normalization \citep[Section~2.4]{liu2021optimal}.
Under independent flips, the proof above combines independent coordinate signs with an upper bound on relative entropy to obtain the~$\Omega\prn{\sqrt d/\kappa^2}$ and~$\Omega\prn{d/\kappa^2}$ lower bounds for the Euclidean unit ball and the cube normalization, respectively.

\section{Online Inverse Linear Optimization}\label{app:oilo-proof}
We prove \cref{cor:oilo} and show that the additive~$D_T(u)$ in the bound on the regret relative to optimal actions in \cref{sec:oilo} cannot be removed.
\restate{\oilobound}
\begin{proof}
Recall the suboptimality loss \citep[Definition~2.3]{mohajerinesfahani2018data} used in \mbox{\cref{sec:oilo}},
\begin{equation}\label{eq:oilo-surrogate}
f_t(w)\coloneqq\max_{a\in\mathcal X_t}\inpr{w,a-y_t}\qquad(w\in\mathcal K).
\end{equation}
The loss is convex as a maximum of linear functions of~$w$. Each slope~$a-y_t$ has norm at most~$G$ by the diameter bound on~$\mathcal X_t$, so~$f_t$ is~$G$-Lipschitz on~$\mathcal K$.
Since~$\widehat a_t$ maximizes~$\inpr{w_t,a}$ over~$\mathcal X_t$, we have~$g_t=\widehat a_t-y_t\in\partial f_t(w_t)$, with~$\norm{g_t}_2\le G$.
Thus the losses~$f_t$ and the vectors~$g_t$ follow the protocol in \cref{sec:protocol}, and \cref{thm:adaptive-known-width} applies.
Equation~\eqref{eq:negative-gap} gives the first inequality below. Since~$y_t\in\mathcal X_t$, we have~$f_t\ge0$ on~$\mathcal K$. Subtracting the nonnegative minima from the comparator loss gives the second inequality, so we have
\begin{equation}\label{eq:oilo-gap}
N_T(u)\le\Delta_T(u)\le\sum_{t=1}^Tf_t(u)=D_T(u).
\end{equation}
By the choice of~$\widehat a_t$, we also have~$f_t(w_t)=\inpr{w_t,g_t}\ge0$. Summing the identity~$\inpr{u,y_t-\widehat a_t}=\inpr{g_t,w_t-u}-f_t(w_t)$ over rounds yields
\begin{equation}\label{eq:oilo-transfer}
R_T^{\mathrm{obs}}(u)=\sum_{t=1}^T\inpr{u,y_t-\widehat a_t}=R_T^{\mathrm{lin}}(u)-\sum_{t=1}^Tf_t(w_t)
\le R_T^{\mathrm{lin}}(u).
\end{equation}
Combining~\eqref{eq:oilo-transfer}, the bound of \cref{thm:adaptive-known-width} on~$R_T^{\mathrm{lin}}(u)$ in terms of~$N_T(u)$, and~\eqref{eq:oilo-gap}, we obtain
\begin{equation}\label{eq:oilo-adaptive-transfer}
R_T^{\mathrm{obs}}(u)\le R_T^{\mathrm{lin}}(u)
\le c\prn*{GW+\sqrt{GW N_T(u)}}
\le c\prn*{GW+\sqrt{GW D_T(u)}},
\end{equation}
thus completing the proof.
\end{proof}

\paragraph{Regret Relative to Optimal Actions.}
\Cref{sec:oilo} defines~$R_T^{\mathrm{opt}}(u)=D_T(u)+R_T^{\mathrm{obs}}(u)$ and derives~$R_T^{\mathrm{opt}}(u)\lesssim GW+D_T(u)$ from \cref{cor:oilo}.
The additive~$D_T(u)$ cannot be removed, even by a randomized learner.
Take~$\mathcal K=[-1,1]$, $\mathcal X_t=\set{-1,0,1}$, and~$y_t=0$ for every~$t$, so that~$G=2$.
For~$u=\pm1$, we have~$\max_{a\in\mathcal X_t}ua=1$, so~$D_T(u)=T$ and~$R_T^{\mathrm{opt}}(u)=\sum_{t=1}^T\prn{1-u\widehat a_t}$.
The observations~$\mathcal X_t$ and~$y_t$ are the same for~$u=1$ and~$u=-1$, so the recommendations~$\widehat a_t$ have the same law under both utilities.
Averaging over~$u=\pm1$ therefore gives~$\prn{\E\brc{R_T^{\mathrm{opt}}(1)}+\E\brc{R_T^{\mathrm{opt}}(-1)}}/2=T$.
Hence some~$u\in\set{-1,1}$ has~$\E\brc{R_T^{\mathrm{opt}}(u)}\ge T=D_T(u)$, and no bound of order~$1+\sqrt{D_T(u)}$ on the regret relative to optimal actions holds uniformly in this model; see also \citet[Section~4]{sakaue2025online}.

\section{Additional Related Work}\label{app:additional-related}
We compare our guarantees and methods with prior work on OCO, contextual search, and OILO.
\Crefrange{app:smooth-comparison}{app:random-order-comparison} discuss small-loss bounds, followed by discussions of mixtures in \cref{app:multiple-rates} and of the two applications in \cref{app:search-comparison,app:oilo-comparison}.

\subsection{Small-Loss Bounds via Smoothness}\label{app:smooth-comparison}
We first review how smoothness yields small-loss bounds and how their assumptions and comparator losses differ from those in our guarantee.
\subsubsection{Smooth Nonnegative Losses}
For convex losses that are $\beta_{\mathrm{sm}}$-smooth and nonnegative on~$\R^d$, with~$\beta_{\mathrm{sm}}>0$, the self-bounding inequality~$\norm{\nabla f_t(w_t)}_2^2\le4\beta_{\mathrm{sm}} f_t(w_t)$ yields small-loss bounds \citep[Lemma~3.1 and Theorem~2]{srebro2010smoothness}.
Let~$L_T(u)\coloneqq\sum_{t=1}^Tf_t(u)$ for~$u\in\mathcal K$.
Substituting this inequality into the online-gradient-descent regret bound and using~$\sum_{t=1}^Tf_t(w_t)=R_T(u)+L_T(u)$ gives a bound on~$R_T(u)$ in terms of~$L_T(u)$.
Scale-free online gradient descent (SOGD) sets its step size using the observed gradients.
\citet[Algorithm~1 and Theorem~2]{zhang2019adaptive} prove that SOGD satisfies
\begin{equation}\label{eq:sogd-comparison}
R_T(u)\le8\beta_{\mathrm{sm}}\prn{\operatorname{diam}_2\mathcal K}^2+\prn{\operatorname{diam}_2\mathcal K}\sqrt{2\delta+8\beta_{\mathrm{sm}} L_T(u)},
\end{equation}
where~$\delta>0$ is the regularization constant in the learning rate. Neither~$L_T(u)$ nor~$T$
is supplied to that update, and each round uses one projected gradient step.
For smooth nonnegative losses, this bound can have better dimension dependence than the bound in \cref{thm:adaptive-known-width}.
Further work develops small-loss guarantees that adapt to unknown strong convexity or exp-concavity \citep{yang2024projection,zhang2025universal,zhao2025adaptivity}, as well as guarantees on every time interval \citep{zhang2019adaptive,yang2024adaptive}.

We next explain why the SOGD bound in terms of~$L_T(u)$ does not directly extend to~$\Delta_T(u)$.
\citet[Assumptions~2 and~3]{zhang2019adaptive} assume Lipschitz gradients on~$\mathcal K$ and require nonnegativity outside~$\mathcal K$ for the self-bounding argument.
Subtracting~$\min_{u\in\mathcal K}f_t(u)$ need not preserve this nonnegativity: for~%
$\mathcal K=[0,1]$ and~$f(w)\coloneqq\prn{w+1}^2$, the shifted function~$f-\min_{\mathcal K}f$ is negative at~$w=-1$.
Consequently, replacing~$L_T(u)$ by~$\Delta_T(u)$ in the SOGD bound can violate the assumptions behind that bound.
Our analysis instead uses~$N_T(u)\le\Delta_T(u)$ from~\eqref{eq:negative-gap} and needs no inequality between gradient norms and loss values.

\subsubsection{Small-Loss Bounds for Shifted Losses}
Small-loss bounds in terms of the shifted comparator loss~$\sum_{t=1}^T\brc{f_t(u)-\min_{\mathcal K} f_t}$ have also been studied.
On a Euclidean-ball domain $\mathcal K$, \citet{orabona2012beyond} obtain a regret bound with logarithmic dependence on this quantity for exp-concave losses whose gradients are uniformly bounded and satisfy the self-bounding inequality~$\norm{\nabla f_t(w)}_2^2\le4H\prn{f_t(w)-\min_{\mathcal K}f_t}$ for all~$w\in\mathcal K$ and~$t$, where~$H>0$.
Their bound is achieved by Online Newton Step and is invariant under additive shifts of the losses.

\citet{zhao2025adaptivity} study small-loss bounds using per-round minima over an enlarged domain.
For convex losses, the analysis assumes~$\norm{\nabla f_t(w)}_2\le G$ on~$\mathcal K$ and~$\beta_{\mathrm{sm}}$-Lipschitz gradients on~$\mathcal K_+\coloneqq\mathcal K+\prn{G/\beta_{\mathrm{sm}}}\B_2^d$, where~$\beta_{\mathrm{sm}}>0$ \citep[Assumption~5]{zhao2025adaptivity}.
Under these assumptions, the regret bound has square-root dependence on~$\Delta_T^+\coloneqq L_T^\star-\sum_{t=1}^T\min_{w\in\mathcal K_+}f_t(w)$ \citep[Corollary~2]{zhao2025adaptivity}, where iterated logarithmic factors in~$T$ are suppressed \citep[Section~2 and Appendices~B.3 and~D.2.1]{zhao2025adaptivity}.
The same algorithm also adapts to gradient variation and to unknown strong convexity or exp-concavity; each round uses one gradient at the played point and updates an ensemble of~$O\prn{\log\prn{1+T}}$ learners \citep[Algorithm~5]{zhao2025adaptivity}.
Our guarantee in \cref{thm:adaptive-known-width} depends on~$\Delta_T(u)$, with minima taken over the decision set~$\mathcal K$, and requires no smoothness.

\subsection{Regret Bounds in Terms of Gradient Norms}\label{app:gradient-norm-comparison}
\citet{gao2026gradient} bound the regret for smooth losses in terms of the
squared gradient norms at the comparator, rather than at the learner's
predictions. In our notation, their Corollary~3.1 shows that AdaGrad-Norm
satisfies
\begin{equation}
R_T(u)\le \beta_{\mathrm{sm}}\prn{\operatorname{diam}_2\mathcal K}^2+
\sqrt2\prn{\operatorname{diam}_2\mathcal K}\sqrt{\sum_{t=1}^T\norm{\nabla f_t(u)}_2^2}
\end{equation}
for convex losses that are~$\beta_{\mathrm{sm}}$-smooth on the whole space. The algorithm does not need to observe the comparator's
gradients. Smoothness relates them to the observed gradients in the proof.
This bound can improve on bounds that depend on unconstrained loss minima, and it does
not require the losses to be bounded below.

We illustrate that the bound of \citet[Corollary~3.1]{gao2026gradient} can grow with $T$ even when our guarantee is independent of~$T$.
To compare them, we take~$\mathcal K=[0,1]$, $f_t(w)\coloneqq w$, and~$u=0$. Then we have~%
$\Delta_T(u)=0$ and $\sum_{t=1}^T\norm{\nabla f_t(u)}_2^2=T$.
Thus their upper bound grows as~$\sqrt T$, whereas \cref{thm:adaptive-known-width} gives a bound independent of~$T$.
For a different case, suppose that the losses have a common minimizer~$u\in\mathcal K$ over~$\R^d$.
Then we have~$\nabla f_t(u)=0$ for every~$t$, so their bound reduces to~$\beta_{\mathrm{sm}}\prn{\operatorname{diam}_2\mathcal K}^2$.
For fixed smoothness and diameter, this bound is independent of~$d$, while our guarantee at~$\Delta_T(u)=0$ retains the width-dependent term~$O\prn{GW}$.
Either bound can therefore be smaller, depending on the losses and the domain.

\citet[Section~4.3 and Appendix~C.9]{gao2026gradient} also explicitly consider
$\sum_{t=1}^T(f_t(u)-\min_{\mathcal K} f_t)$, the same quantity as~$\Delta_T(u)$.
Their Proposition~4.1 rules out replacing the usual $L_T(u)$ by
this shifted comparator loss while retaining a bound of the form
$\beta_{\mathrm{sm}}\prn{\operatorname{diam}_2\mathcal K}^2+\prn{\operatorname{diam}_2\mathcal K}\sqrt{\beta_{\mathrm{sm}}\Delta_T(u)}$.

\subsection{Small-Loss Bounds for Online Linear Optimization}\label{app:cone-small-loss}
For online linear optimization on the unit $\ell_p$ ball with fixed~$2\le p<\infty$ and loss vectors of dual norm at most one, \citet{mayo2026minimax} attains the minimax small-loss rate by tuning the FTRL regularizer to the learner's cumulative shifted loss.
On the Euclidean unit-ball domain, this gives a regret bound of~$O\prn{1+\sqrt{\Delta_T^\star}}$, as shown in \cref{tab:small-loss}.
\citet[Proposition~F.2]{mayo2026minimax} also obtains this Euclidean bound by restarting the multiplicative-weights algorithm of \citet{canyakmaz2023multiplicative} according to the learner's cumulative shifted loss.

These regret bounds concern linear losses. We now examine their use for general convex losses by linearization. For a fixed comparator~$u$, linearizing~$f_t$ as~$w\mapsto\inpr{g_t,w}$ replaces~$\Delta_T(u)$ by the shifted linear loss~$\sum_{t=1}^T\prn{\inpr{g_t,u}-\min_{v\in\mathcal K}\inpr{g_t,v}}$, which can be much larger.
Take~$f_t(w)\coloneqq\abs{w}$ on~$\mathcal K=[-1,1]$ and~$u=0$. For any played points and any subgradients~$g_t\in\partial f_t(w_t)$ with~$\abs{g_t}=1$, we have $N_T(0)=\Delta_T(0)=0$. In every round, we have~$\inpr{g_t,0}-\min_{v\in[-1,1]}g_tv=1$, so the shifted linear loss of~$u$ equals~$T$. For even~$T$, take~$w_t=0$ and~$g_t=(-1)^t$ in every round. Since the subgradients sum to zero, every comparator has shifted linear loss~$T$, while the original losses have~$\Delta_T^\star=0$. Therefore, our guarantees do not follow from the small-loss bounds for linear losses.

\subsection{Small-Loss Bounds via Follow the Perturbed Leader}\label{app:ftpl-comparison}
Another approach to small-loss bounds uses offline optimization with random perturbations.
Note that these methods require additional structural assumptions and access to offline optimization oracles. By contrast, our algorithm observes one subgradient per round and projects onto the decision set.
\subsubsection{Objective Perturbation Approach for Smooth Losses}
\citet[Theorem~3.2 and Corollary~3.3]{abernethy2019privacy} derive
first-order bounds using differential stability of objective perturbations.
Their OCO guarantee assumes an oblivious sequence of bounded nonnegative convex losses.
Each loss's Hessian has a bounded largest eigenvalue and rank at most one. Their algorithm
minimizes the regularized sum of past loss functions plus a random linear perturbation. For~$\mathcal K\subset R\B_2^d$ with~$R>0$, $\beta_{\mathrm{sm}}\ge0$,
$\norm{\nabla f_t(w)}_2\le G$, and~$\nabla^2 f_t(w)\preceq\beta_{\mathrm{sm}} I_d$
for every~$t$ and~$w\in\mathcal K$, their Equation~(3) gives, with Gamma
perturbations and~$0<\eps\le1$,
\begin{equation}\label{eq:dp-first-order}
\E R_T(u^\star)\le 2\eps L_T^\star+
\frac{3\beta_{\mathrm{sm}} R^2+12GRd}{\eps},
\end{equation}
where~$u^\star$ minimizes~$\sum_{t=1}^Tf_t$ over~$\mathcal K$, so that~$\sum_{t=1}^Tf_t(u^\star)=L_T^\star$, and the
expectation is over the perturbations.
The algorithm requires the past loss functions for its optimization step. %

\subsubsection{Oracle-Efficient Methods for Finite Action Sets}
For a finite action set and losses in~$[0,1]$, \citet[Theorem~1]{wang2022adaptive} bound the expected regret by~$O\prn{\sqrt{1+L_T^\star}}$, where the implicit constant depends on the number of actions, the number of columns in the perturbation matrix, and its approximability parameter. Their approximability condition (Definition~2) controls how adding the current loss changes the probability of selecting an action. The implementability condition (Definition~3) requires each column of the perturbation matrix to admit a representation, up to an additive constant, by weighted outcomes that can be passed to the offline oracle. Their Algorithm~1 minimizes the perturbed cumulative loss on past rounds to predict. After observing the outcome, it sets the learning rate for the next round by using the offline oracle to compute the minimum, over actions, of the cumulative loss on the outcomes observed so far.

\citet{block2025separated} give another method using Gaussian perturbations and an offline optimization oracle for a finite function class~$\mathcal F$.
Their separation assumption requires a supplied finite set of input points on which the average squared difference between any two distinct functions is bounded below by a common positive constant (Definition~5).
Their regret bound in Theorem~1 contains an explicit~$\log\prn{2T\abs{\mathcal F}}$ factor.
Our regret bound requires neither a separating set of inputs nor an oracle for minimizing past losses.

\subsection{Small-Loss Bounds under the Random-Order Model}\label{app:random-order-comparison}
Small-loss bounds are also studied under the random-order model.
\citet{sakaue2026randomorder} obtain expected small-loss regret bounds when
an adversarially chosen multiset of losses is revealed in a uniformly random
order. Their batch-to-online transformation applies offline approximation
algorithms to the observations so far and varies their approximation accuracy.
The bounds require control of average sensitivity and stability as the accuracy
changes. For online submodular function minimization and online~$\ell_1$ regression studied in their paper, applying our result improves the dependence on the comparator loss even under adversarial ordering.

Fix any sequence of submodular losses with values in~$[0,1]$ on an~$n$-element ground set. For each Lov\'asz extension, submodularity bounds the sums of the positive and negative parts of a greedy subgradient by one each, so we may take~$G=2$. The cube~$[0,1]^n$ has the width of $O\prn{n}$. Applying \cref{thm:adaptive-known-width} and rounding each prediction with a fresh uniform threshold bounds the expected regret by~$O\prn{n+\sqrt{nL_T^\star}}$.
In comparison, \citet[Theorem~5.2]{sakaue2026randomorder} obtain~$\widetilde O\prn{n^3+n^{3/4}(L_T^\star)^{3/4}}$ under the random-order model, where $\widetilde O$ suppresses logarithmic factors in~$T$ and the dimension. As they note, treating all~$2^n$ subsets as experts already gives~$O\prn{n+\sqrt{nL_T^\star}}$ in adversarial order, but this is prohibitively expensive.

For online~$\ell_1$ regression with absolute losses bounded by one on a compact convex decision set in~$\R^d$, \cref{thm:adaptive-known-width} bounds the regret by~$O\prn{d+\sqrt{dL_T^\star}}$ after an affine change of coordinates. This is obtained by choosing coordinates in the affine hull so that the maximum-volume ellipsoid in~$\mathcal K-\mathcal K$ is the unit ball; the loss bound then allows~$G=2$, and John's ellipsoid theorem gives a width bound of~$W=O\prn{d}$.
This improves on their random-order guarantee of $\widetilde O\prn{d+d^{1/4}(L_T^\star)^{3/4}}$ \citep[Theorem~6.1]{sakaue2026randomorder}.

\subsection{Excess-Loss Bounds and Multiple Learning Rates}\label{app:multiple-rates}
This section compares our method for achieving regret guarantees without knowing~$N_T(u)$ with existing approaches to bounding excess losses by mixing multiple learning rates.
\subsubsection{Second-Order and Worst-Case Regret Bounds}\label{app:second-order-comparison}
For prediction with expert advice, \citet[Corollary~4 and Theorem~9]{gaillard2014second} derive second-order regret bounds for Adapt-ML-Prod and obtain improved bounds when the sum of the negative parts of the excess losses against an expert is small.
\citet{luo2015normalhedge} propose AdaNormalHedge, which gives a small-loss bound for prediction with expert advice.
Their Theorem~2 also gives a bound in terms of the negative excess losses against each expert. The regret bounds for Adapt-ML-Prod and AdaNormalHedge contain factors that can grow double-logarithmically with~$T$. In a different direction,
\citet{chen2021impossible} give regret bounds that depend on each expert's sum of squared losses and hold for all experts simultaneously.

For OCO, MetaGrad Full \citep[Corollary~8]{erven2021metagrad} and Maler \citep[Theorem~1]{wang2020optimality} give second-order regret bounds of order $d\log\prn{1+T}+\sqrt{d\log\prn{1+T}\sum_{t=1}^T r_t(u)^2}$ against a best fixed decision~$u$ on the unit Euclidean ball when each loss has gradient norm at most one throughout the ball. Maler also guarantees~$O\prn{\sqrt T}$ regret for general convex losses. For linear losses, \citet[Proposition~F.5]{mayo2026minimax} obtains $O\prn{d\log\prn{1+T}+\sqrt{d\Delta_T^\star\log\prn{1+T}}}$ from MetaGrad on centrally symmetric convex bodies, with loss vectors bounded in the dual norm. \citet{sakaue2025online} apply MetaGrad to online inverse linear optimization to adapt to the unknown cumulative suboptimality of feedback actions.

Recent studies analyze variants of Squint, a familiar algorithm for prediction with expert advice that mixes over experts and learning rates to obtain second-order regret bounds \citep{koolen2015squint}. \citet{luo2026squint} replaces each expert's sum of squared excess losses with a single weighted sum shared by all experts. The resulting bound still includes~$\log\log T$ terms.
\citet{orabona2026squint} removes the~$\log\log T$ factor from Squint's worst-case bound by changing the prior over betting fractions.
This worst-case guarantee remains proportional to~$\sqrt T$ for fixed expert prior weights, so it does not give a bound in terms of~$N_T(u)$.

Our \cref{thm:adaptive-known-width} bounds the linearized regret by~$O\prn{GW+\sqrt{GW N_T(u)}}$ for every~$u\in\mathcal K$, without requiring the algorithm to know~$N_T(u)$ and with no explicit dependence on~$T$. It applies to general compact convex domains, in contrast to the negative-excess guarantees above that concern prediction with expert advice.

\subsubsection{Recursive Mixtures}
We next examine how the choice of mixing method affects the regret bound relative to each base learner.
In \cref{sec:adaptation}, the scalar correction separates the comparison of the mixed prediction with learner~$i$ from that of learner~$i$ with~$u$, and the recursive mixture handles the first comparison.
Let us discuss applying a standard mixture of expert predictions, such as ML-Prod, to our base learners.
ML-Prod assigns an initial weight and a learning rate to each expert. For expert~$i$ with initial weight~$\pi_i$ and rate~$\eta_i$, its regret bound includes~$\log\prn{1/\pi_i}/\eta_i$ \citep[Theorem~1]{gaillard2014second}. To compare this term with our bound, we specialize it to equal prior weights on~$\Theta(\log T)$ learners. The term then has order~$(\log\log T)/\eta_i$, involving an explicit dependence on $T$.
By contrast, in our recursive mixture, \eqref{eq:mix-path-bound} sums~$\prn{\log2}/\eta_k$ along the path to learner~$i$. With~$M=2GW/\log2$, this gives a comparison penalty of~$O\prn{GW\sum_{k=0}^i\eps_k^{-1}}=O\prn{GW/\eps_i}$ because~$\eps_k=2^{-k}/8$. Thus the recursive mixture removes the~$\log\log T$ factor from this comparison penalty relative to ML-Prod with equal prior weights.

Our recursive mixture is reminiscent of those used to bound switching regret, which combine predictions of learners run over different time intervals \citep{pasteris2024switching,yang2026logarithmic}. Those methods compete with a comparator that can change between intervals. Our mixture instead combines learners with different~$\eps_i$ to adapt to~$N_T(u)$ for a fixed comparator~$u$.

\subsubsection{Comparators in the Interior of the Simplex}
For prediction with expert advice, each expert corresponds to a vertex~$e_i$ of the simplex. For losses in~$[0,1]$, \citet[Theorem~2]{luo2015normalhedge} bound the regret of AdaNormalHedge against any distribution~$u$ over experts in terms of the weighted sum~$\sum_{i=1}^d u_iN_T(e_i)$.
We below clarify that this weighted sum can exceed~$N_T(u)$.
For $r_t(e_i)=\inpr{g_t,w_t-e_i}$ and weights~$u_i\ge0$ summing to one, we have
\begin{equation}\label{eq:vertex-negative-parts}
\brc{-r_t(u)}_+=\brc*{-\sum_{i=1}^d u_i r_t(e_i)}_+
\le\sum_{i=1}^d u_i\brc{-r_t(e_i)}_+.
\end{equation}
The right-hand side can be positive when the left-hand side is zero.
In fact, for~$d=2$, $w_t=u=(1/2,1/2)$, and~$g_t=(0,1)$, we have~$r_t(e_1)=1/2$ and~$r_t(e_2)=-1/2$, so the left-hand side is zero and the right-hand side is~$1/4$.
Our guarantee depends directly on~$N_T(u)$ and avoids this increase.

\subsection{Contextual Search, Pricing, and Comparison Feedback}\label{app:search-comparison}
Contextual search grew out of online pricing, where the seller observes only whether the buyer accepts the posted price.
\citet{kleinberg2003value} studied the case of buyers with the same unknown valuation. Later work allowed valuations to depend on observable features \citep{amin2014repeated,cohen2020feature}.
Contextual search has since been studied extensively with exact comparisons \citep{lobel2018multidimensional,mao2018contextual,leme2022intrinsic,cosson2026contextual} and under noise or adversarial corruption \citep{liu2021optimal,leme2022density,krishnamurthy2023contextual,zuo2024corruption,gupta2025robust}.

Two common losses in contextual search are pricing loss and absolute prediction error.
The pricing loss is the buyer's valuation minus the seller's revenue.
A price just above that valuation loses the sale even though the absolute prediction error is small.
Our guarantee in \cref{sec:search} concerns the cumulative absolute error, so we focus below on prior bounds for that quantity.

For a target~$u\in\R^d$ drawn from a known prior with a finite second moment, \citet[Theorem~2.1]{cosson2026contextual} use posterior moments to bound the expected cumulative absolute error by~$O\prn{\tr(\operatorname{Cov}(u)^{1/2})}$ under exact comparisons when each context lies in the unit Euclidean ball and is conditionally independent of~$u$ given the observed history. Their Theorem~2.4 bounds the expected cumulative absolute error by~$O\prn{\sqrt d}$ for every target in that ball, with~$O\prn{d^2}$ arithmetic operations per round. Our bound in \cref{cor:search-adaptive} has the same order under exact comparisons; our algorithm has a higher computational cost but also handles bit flips under the conditional bias assumption~\eqref{eq:noise-model}, without knowing the flip probabilities.

Noisy contextual search with independent bit flips is studied by \citet[Section~5.2]{liu2021optimal}. For each fixed flip probability~$p<1/2$ with bias~$\kappa=1-2p$, they obtain an~$O_\kappa\prn{d^{25}}$ bound on the expected cumulative absolute error and ask for a polynomial-time algorithm with a horizon-independent~$\poly\prn{d}$ bound. When both the target and the contexts lie in the unit Euclidean ball, our algorithm bounds the expected cumulative absolute error by~$O\prn{\sqrt d/\kappa^2}$ without knowing~$\kappa$. This resolves their question in the real-arithmetic model (\cref{sec:search}).

\citet{krishnamurthy2023contextual} consider a different model, in which an adversary corrupts the target responses on an unknown number of rounds. They give a high-probability bound on the cumulative absolute error relative to the uncorrupted target responses, with dependence on the number of corrupted rounds and~$T$.
\citet{leme2022density} treat adversarial target corruption and also apply
their density updates to convex optimization. Their Corollary~3.4 bounds the cumulative absolute error relative to the corrupted responses by~%
$O\prn{C_1+d\log T}$, where~%
$C_1$ is the sum of the absolute values of the corruptions to the target responses.
In the noisy comparison model of \cref{sec:search}, the target is fixed and the bit can flip.
Under the conditional bias assumption~\eqref{eq:noise-model}, \cref{cor:search-adaptive} bounds the expected cumulative error independently of~$T$, without assuming a bound on the total number of flipped bits. Our guarantee therefore uses a probabilistic condition on comparison errors and achieves independence of $T$, whereas the bounds of \citet{krishnamurthy2023contextual} and \citet{leme2022density} depend on $T$ and the number or total magnitude of adversarial corruptions.

\subsection{Online Inverse Optimization}\label{app:oilo-comparison}
OILO, a.k.a.\ contextual recommendation, and a related model called coactive learning have been studied extensively \citep{shivaswamy2012coactive,barmann2017emulating,barmann2020online,gollapudi2021contextual,sakaue2025revisiting,sakaue2025online,oki2026finite,sakaue2026corectron,cai2026efficient,kitaoka2026skipping,sakaue2026multiscale,gupta2026optimal}.
Below we compare our guarantees and methods for OILO with those of earlier algorithms that are particularly relevant to ours.

\subsubsection{Prior Work under Optimal Feedback Actions}
In OILO, a standard protocol requires the learner to choose its utility estimate before seeing the current action set \citep{barmann2017emulating,barmann2020online,sakaue2025online}.
Our algorithm uses the same timing.
When the utility vector and feasible actions lie in the Euclidean unit ball and the user always chooses an optimal action, \citet{cai2026efficient} bound the cumulative utility shortfall of the recommendations by~$O\prn{d}$ under this protocol. The deterministic polynomial-time algorithm of \citet{gupta2026optimal} achieves an $O\prn{\sqrt d}$ bound, matching the lower bound for~$T\ge d$ \citep{sakaue2025online,sakaue2026multiscale}. Optimal choices give~$D_T(u)=0$ and hence~$N_T(u)=\Delta_T(u)=0$ by~\eqref{eq:oilo-gap}. \Cref{cor:oilo} then gives the same~$O\prn{\sqrt d}$ bound on~$\mathcal K=\B_2^d$ by taking~$G=2$ and~$W\asymp\sqrt d$.

\subsubsection{Cutting-Plane Methods and Regret with Suboptimal Feedback Actions}\label{app:oilo-suboptimal}
\citet{gollapudi2021contextual} use cutting-plane algorithms for contextual recommendation when the user always chooses an optimal action. For a utility vector and feasible actions in the unit Euclidean ball, one algorithm bounds the recommendation regret by~$O\prn{d\log T}$ and the other by~$\exp\prn{O\prn{d\log d}}$. When the user's choice differs from the recommendation, the learner uses their difference to determine a halfspace containing the true utility vector~$u$.
\citet{besbes2021optimal,besbes2025contextual} also use optimal action feedback to cut the set of possible cost vectors. Their algorithm uses ellipsoidal cones and achieves regret logarithmic in~$T$.

If the user's choice is suboptimal, the halfspace inferred from that choice and the recommendation can exclude~$u$. \citet{cai2026efficient} and \citet{gupta2026optimal} study a cutting-plane game that allows such errors. At query~$w_t$, the oracle returns a unit direction~$-g_t$. The signed error is given by~$r_t(u)=\inpr{-g_t,u-w_t}$, which is negative if the indicated halfspace excludes the target~$u$.
For~$\norm{u}_2\le R$, \citet[Theorem~3.6]{cai2026efficient} bound the sum of the positive errors by~$O\prn{dR+C}$ given an upper bound~$C$ on the sum of the negative parts.
\citet[Section~4]{gupta2026optimal} improve this to~$O\prn{R\sqrt d+C}$, and their algorithm does not need to know~$C$. Both algorithms may query points outside the target domain.

Our algorithm predicts in~$\mathcal K$ without knowing~$N_T(u)$. To compare the bounds on the cumulative positive error, we add the sum of the negative parts to the signed regret bound in \cref{thm:adaptive-known-width}, obtaining
\begin{equation}
\sum_{t=1}^T\brc{r_t(u)}_+
=R_T^{\mathrm{lin}}(u)+N_T(u)
\le N_T(u)+c\prn*{GW+\sqrt{GW N_T(u)}},
\end{equation}
where~$c>0$ is a universal constant. On~$\mathcal K=R\B_2^d$, taking~$G=1$ and~$W\asymp R\sqrt d$ recovers the bound of $O\prn{R\sqrt d+C}$ on the cumulative positive error for~$N_T(u)\le C$. For the signed regret, \cref{thm:adaptive-known-width} gives the finer bound of~$O\prn{R\sqrt d+d^{1/4}\sqrt{R N_T(u)}}$.

We next discuss using the cutting-plane guarantee of \citet{gupta2026optimal} to bound the regret relative to optimal actions when the user may choose suboptimal actions. At each query~$w_t$, we recommend~$\widehat a_t\in\argmax_{a\in\mathcal X_t}\inpr{w_t,a}$ and return~$-g_t/G=\prn{y_t-\widehat a_t}/G$, skipping the update when~$g_t=0$.
The diameter bound gives $\norm{g_t/G}_2\le1$, which suffices for their proof \citep[Lemmas~2.3(c) and~2.7]{gupta2026optimal}. Each signed error is~$r_t(u)/G$, so for~$u\in\B_2^d$ their bound gives
\begin{equation}\label{eq:gupta-oilo-rescaling}
\sum_{t=1}^T\brc{r_t(u)}_+\lesssim G\sqrt d+\sum_{t=1}^T\brc{-r_t(u)}_+.
\end{equation}
To bound the sum on the right of~\eqref{eq:gupta-oilo-rescaling} by~$D_T(u)$, we use~$\brc{-r_t(u)}_+\le f_t(u)$, where~$f_t(u)$ is the utility gap between an optimal action and the user's choice~$y_t$ \citep[proof of Theorem~4.1]{sakaue2025online}.
Optimality of the recommendation under~$w_t$ gives~$\inpr{u,y_t-\widehat a_t}\le r_t(u)$.
Using~$\sum_{t=1}^Tf_t(u)=D_T(u)$ in~\eqref{eq:gupta-oilo-rescaling} therefore yields
\begin{equation}\label{eq:gupta-oilo-optimal}
R_T^{\mathrm{opt}}(u)
=\sum_{t=1}^T\max_{a\in\mathcal X_t}\inpr{u,a-\widehat a_t}
=D_T(u)+R_T^{\mathrm{obs}}(u)
\le D_T(u)+\sum_{t=1}^T\brc{r_t(u)}_+
\lesssim G\sqrt d+D_T(u).
\end{equation}
On the unit ball, this matches the bound on~$R_T^{\mathrm{opt}}(u)$ derived from \cref{cor:oilo} in \cref{sec:oilo}.

For the regret relative to the user's observed actions, $R_T^{\mathrm{obs}}(u)$, earlier methods already achieve square-root dependence on~$D_T(u)$. \citet[Theorem~4.1]{sakaue2025online} use MetaGrad, and \citet[Theorem~2]{sakaue2026corectron} develops CoRectron, an extension of the second-order perceptron.
For~$\mathcal K=\B_2^d$ and action sets of diameter at most one, both give
$R_T^{\mathrm{obs}}(u)\lesssim d\log\prn{T+2}+\sqrt{d\log\prn{T+2}D_T(u)}$.
With~$G=1$ and~$W\asymp\sqrt d$, our \cref{cor:oilo} replaces $d\log\prn{T+2}$ by~$GW$ in both terms, giving $R_T^{\mathrm{obs}}(u)\lesssim\sqrt d+d^{1/4}\sqrt{D_T(u)}$.
In terms of computation, CoRectron updates its utility estimate in~$O\prn{d^2}$ arithmetic operations without a projection \citep[Section~6]{sakaue2026corectron}. Our algorithm mixes base learners and uses Euclidean projections, so its updates are more costly.

\end{document}